\documentclass[11pt,a4paper,twocolumn]{article}

\usepackage[a4paper,margin=2.5cm,heightrounded=true]{geometry}
\usepackage{authblk}
\usepackage{abstract}
\usepackage[numbers,sort&compress]{natbib}

\usepackage{times}
\usepackage{latexsym}

\usepackage[T1]{fontenc}

\usepackage[utf8]{inputenc}

\usepackage{microtype}
\usepackage{graphicx}
\usepackage{subcaption}
\usepackage{booktabs}
\usepackage{algorithm}
\usepackage{algorithmic}
\usepackage{hyperref}

\usepackage{inconsolata}

\usepackage{amsmath}
\usepackage{amssymb}
\usepackage{mathtools}
\usepackage{amsthm}

\usepackage[capitalize,noabbrev]{cleveref}

\theoremstyle{plain}

\theoremstyle{definition}

\theoremstyle{remark}

\usepackage[textsize=tiny]{todonotes}

\usepackage{tcolorbox}
\usepackage{xcolor}
\definecolor{promptbg}{RGB}{245,245,245} 
\usepackage{colortbl}
\definecolor{graybg}{gray}{0.9} 

\usepackage{enumitem}
\usepackage{float}
\usepackage{multicol}
\usepackage{multirow}

\title{What Color Is the Text? A Benchmark for Hallucination Induced by Image-Embedded Prompt}

\author[1]{Jinkun Zhao}
\author[1,2]{Lei Huang}
\author[1]{Haixin Ge}
\author[1,2,*]{Wenjun Wu}

\affil[1]{SKLCCSE, Institute of Artificial Intelligence, Beihang University, Beijing, China}
\affil[2]{Beijing Advanced Innovation Center for Future Blockchain and Privacy Computing, Beihang University, Beijing, China}
\affil[*]{Corresponding author.}

\date{}

\begin{document}
\twocolumn[%
\maketitle
\begin{onecolabstract}
We introduce Embedded Stroop, a controlled diagnostic paradigm for measuring image-embedded prompt interference in Multimodal Large Language Models (MLLMs), where the query is rendered directly inside the visual input. Using the What-Color-Is-the-Text (WCIT) benchmark, which covers 59 fine-grained colors under Standard, Flipped, and Masked variants, we evaluate 16 proprietary and open-source models. To distinguish semantic capture from general color-naming failure, we decompose model responses into Accuracy, Stroop Hallucination Rate (SHR; answering the embedded word rather than the true text color), and Other Error Rate, and validate the effect with permutation tests, with 56 of 64 conditions remaining significant after FDR correction. Although exact color accuracy is low (6.3\%) under the 59-color vocabulary, mapping predictions to 11 basic color families shows that models retain coarse color perception (38.4\%) while still exhibiting a substantial SHR (21.6\%). A conditional analysis restricted to colors correctly named in the Standard setting further confirms the effect, with pooled conditional SHR exceeding 50\%. Masking or flipping the embedded text reduces Stroop hallucinations, suggesting that semantic legibility can dominate visual color perception in MLLMs.
\end{onecolabstract}
]

\section{Introduction}
\label{introduction}

Multimodal Large Language Models (MLLMs) have achieved remarkable proficiency in aligning visual perception with linguistic reasoning, driving breakthroughs in zero-shot generalization. Despite these strides, the reliability of MLLMs remains compromised by hallucinations, where model outputs diverge from visual reality. Existing research primarily attributes these failures to \textit{cross-modal misalignment}, typically investigating scenarios where textual prompts conflict with visual content \cite{bai2024hallucination, deng2025words}. However, this perspective overlooks a more insidious challenge: \textit{intra-modal semantic interference}. Recent benchmarks like ColorBench \cite{liang2025colorbench} evaluate color recognition and reasoning, and studies on word-color conflict in VLMs \cite{teker2025color, sheta2025vlm} have demonstrated that models read rather than perceive ink color. Yet these works either decouple the query from the visual stimulus or test only textual prompt conflict. A critical open question remains: What happens when the \textit{query itself} is embedded in the visual input, creating a unified modality where semantic content and visual attributes coexist in pixel space?

\begin{figure}[t!] 
 \centering 

 \begin{subfigure}[b]{0.48\columnwidth}
 \centering
 \includegraphics[width=\linewidth]{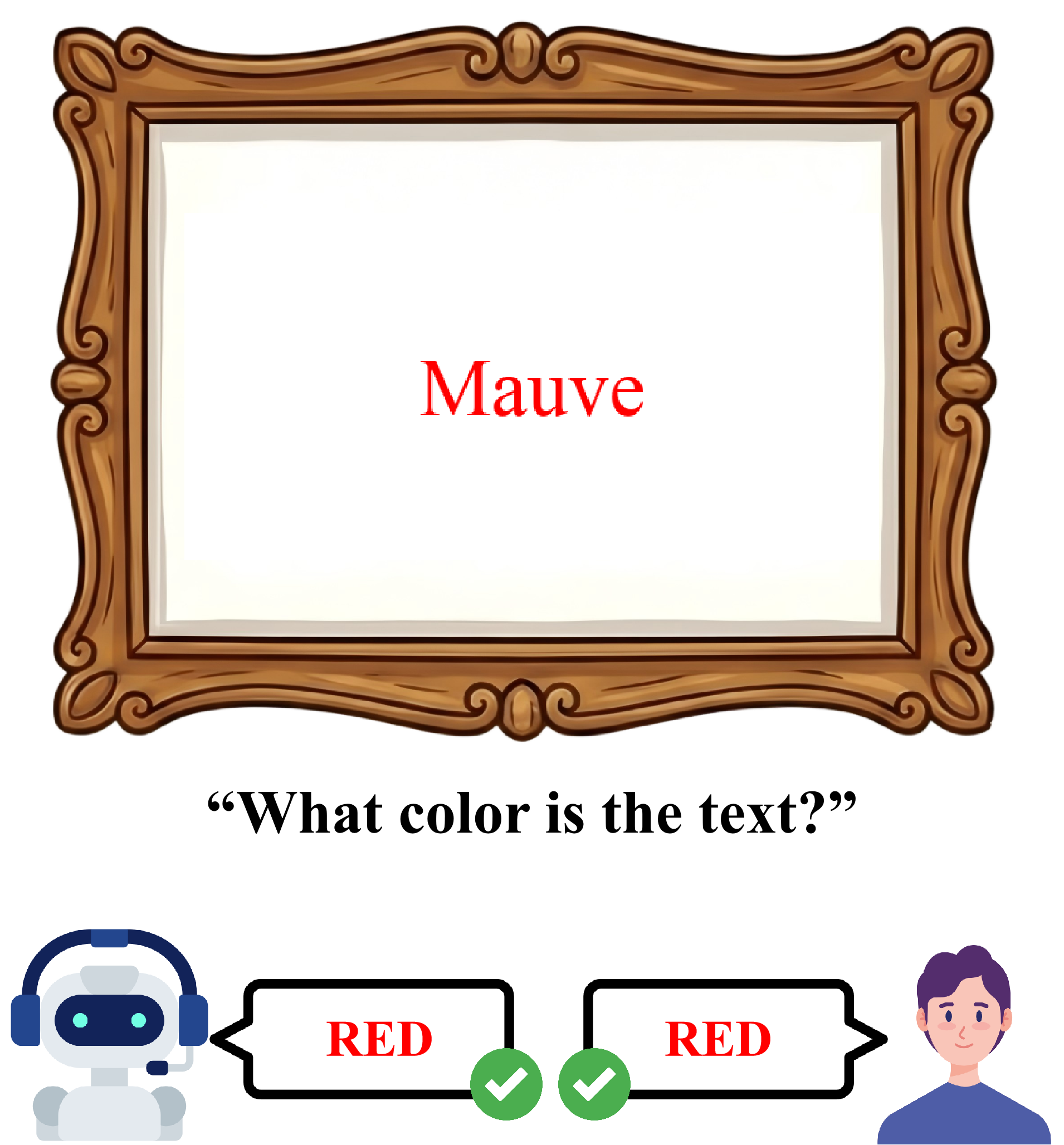}
 \caption{Standard Stroop.} 
 \label{img:standard_stroop} 
 \end{subfigure}
 \hfill 
 \begin{subfigure}[b]{0.48\columnwidth}
 \centering
 \includegraphics[width=\linewidth]{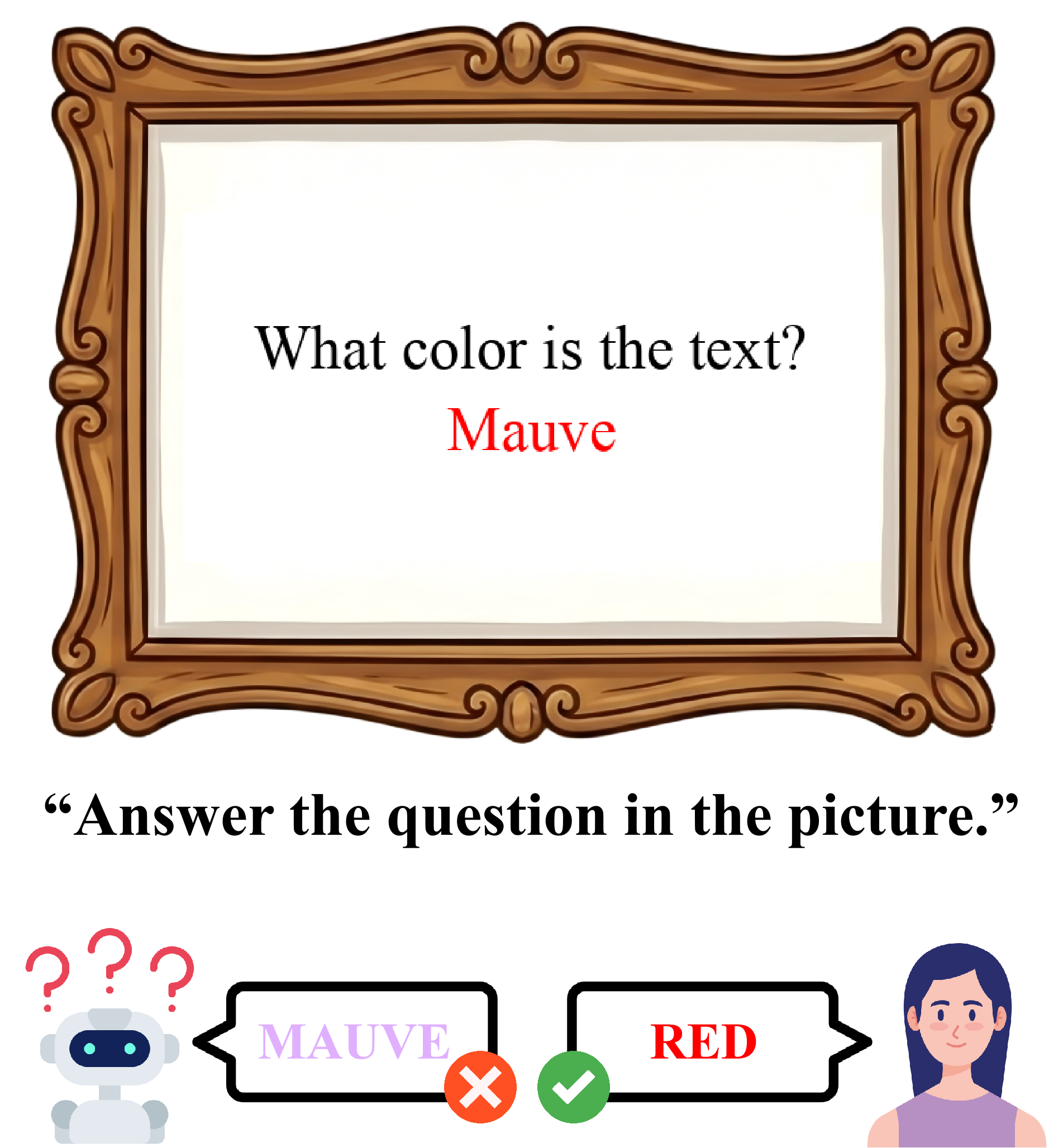}
 \caption{Embedded Stroop.} 
 \label{img:wcit_stroop} 
 \end{subfigure}

 \caption{\textbf{Standard Stroop vs. Embedded Stroop.} While MLLMs demonstrate robustness in the (a) \textit{Standard Stroop} test, they exhibit severe hallucinations in the (b) \textit{Embedded Stroop} setting, where visually embedding the query triggers intense semantic interference that overrides visual perception.}
 \label{img:intro_main} 

 \vskip -0.1in 
\end{figure}

To address this question, we introduce the \textbf{WCIT} (What-Color-Is-the-Text) benchmark, a controlled diagnostic designed to isolate \textit{image-embedded prompt interference}---where embedding the query into the visual input causes semantic processing to override visual perception. As illustrated in Figure~\ref{img:standard_stroop} and Figure~\ref{img:wcit_stroop}, unlike prior Stroop evaluations relying on textual prompts, WCIT embeds the query and conflicting color name directly into pixel space, fusing the visual cue, query text, and interfering semantic text into a single modality. The model must disentangle the \textit{visual rendering} (e.g., blue ink) from the \textit{semantic content} (e.g., the word ``RED''), providing a controlled test of perceptual grounding versus semantic bias.

Our evaluation across 16 proprietary and open-source MLLMs reveals a widespread vulnerability: when visualized text conflicts with its physical color, models exhibit a persistent \textit{blind faith in text}, suggesting a significant challenge in prioritizing visual perception over semantic processing. WCIT serves as a targeted stress test for image-embedded prompt interference.

The main contributions are:
\begin{itemize}
    \item \textbf{WCIT Benchmark:} We introduce a controlled diagnostic benchmark that embeds the query directly into the visual input, enabling a unified-modality evaluation of whether MLLMs can disentangle visual attributes from embedded semantic content.

    \item \textbf{Three-way Error Decomposition:} We propose an error decomposition framework that separates correct color perception, Stroop-specific semantic capture, and general color-naming failure, and further validate the decomposition through permutation tests and coarse-grained color-family analysis.

    \item \textbf{Conditional Validation:} We conduct a conditional analysis on samples that models can correctly name in the Standard setting, showing that the pooled SHR reaches 51.9\% with a 95\% confidence interval of [35.4\%, 66.8\%], thereby confirming that the observed Stroop effect is not merely caused by color-naming inability.

    \item \textbf{Extensive Evaluation:} We evaluate 16 MLLMs across multiple WCIT variants, revealing systematic image-embedded prompt interference and showing that adversarial perturbations can suppress hallucinations, while an adversarial-aware prompt reduces SHR by up to 91\%.
\end{itemize}

\section{Related Work}
\label{sec:related_work}

\subsection{Cross-Modal Hallucination in MLLMs}
Hallucination remains a pervasive challenge in Multimodal Large Language Models (MLLMs), where generated content frequently diverges from visual reality due to cross-modal misalignment or insufficient grounding \cite{bai2024hallucination, rohrbach2018object, dai2023plausible, sun2024aligning, wang2024mdpo, gunjal2024detecting, shang2025pixels,chen2024multi}. 
Recent studies have further dissected this phenomenon as a failure in conflict resolution, revealing that MLLMs exhibit a stubborn "blind faith" in textual prompts or parametric knowledge. When utilizing contradictory information across modalities, models often prioritize linguistic cues over visual evidence, leading to inconsistency reasoning failures and knowledge-driven hallucinations \cite{deng2025words, hua2025vision, zhang2025evaluating, liu2024insight, zhu2024unraveling, yan2025multimodal}. 

In contrast to these inter-modal conflict studies, our work introduces a novel paradigm of intra-modal semantic interference. By embedding textual queries directly into the visual plane, we demonstrate that "visualized text"—semantic information rendered purely as visual signals—can independently trigger hallucinations, proving that perception failures can originate from conflicts strictly within the visual modality.

\subsection{Color Perception and the Stroop Effect}

The Stroop effect\cite{stroop1935,stroop2017} serves as a fundamental paradigm in cognitive psychology for assessing executive function and the inhibition of cognitive interference \cite{patel2025deficient}. With the rapid evolution of MLLMs, this protocol has been increasingly adapted to evaluate machine color intelligence. Recent work has explored cognitive control in VLMs through psychophysical paradigms \cite{luo2025machine}, revealing that models exhibit a strong bias toward reading text over perceiving visual attributes \cite{teker2025color}. Broader VLM interpretability frameworks such as VLM-LENS \cite{sheta2025vlm} further corroborate the gap between behavioral performance and internal competence. Concurrently, benchmarks such as \textsc{ColorBench} and \textsc{ColorBlindnessEval} have systematically assessed MLLMs on tasks ranging from fine-grained color naming to adversarial Ishihara tests, witnessing steady improvements in model robustness and alignment \cite{liang2025colorbench, ling2025colorblindnesseval, hayashi2025diagnosing, ye2025assessing, gomez2025color, rudman2026mechanisms}. 

However, we argue that these performance gains may not imply a genuine mastery of executive control against semantic interference. While prior studies investigate hallucinations induced by textual prompts or external inconsistencies \cite{patel2025deficient, zhang2025modalities}, our work targets a more intrinsic failure mode. By visually embedding the interrogative context directly into the image—forcing the model to process the query and the conflicting stimulus within the same visual encoder—we demonstrate that current MLLMs have not truly overcome the Stroop effect. Instead, they remain highly susceptible to \textit{visualized text}, exhibiting persistent hallucinations that standard, decoupled evaluation protocols fail to expose.

\section{The WCIT Benchmark} 
\label{sec:benchmark}

To construct the dataset, we first curated a comprehensive list of named colors and their corresponding RGB values from a specialized color repository\footnote{\url{https://htmlcolorcodes.com/colors/}}. To quantify the similarity between colors, we employed the following metric.

\subsection{Color Similarity Metric}

To rigorously quantify perceptual divergence, we employ the \textbf{CIEDE2000} color difference metric ($\Delta E_{00}$)~\cite{luo2001development}, which operates in the perceptually uniform CIE $L^*a^*b^*$ space and incorporates corrections for lightness, chroma, and hue non-uniformities. Unlike Euclidean distance in RGB space, $\Delta E_{00}$ closely aligns with human visual perception, making it well-suited for validating whether a model has succumbed to the Stroop effect. The full formula and derivation details are provided in Appendix~\ref{appendix:ciede2000}.

\subsection{Color Space Filtering and Sampling}

\begin{figure}[t]
\vskip 0.2in
\begin{center}
\centerline{\includegraphics[width=\columnwidth]{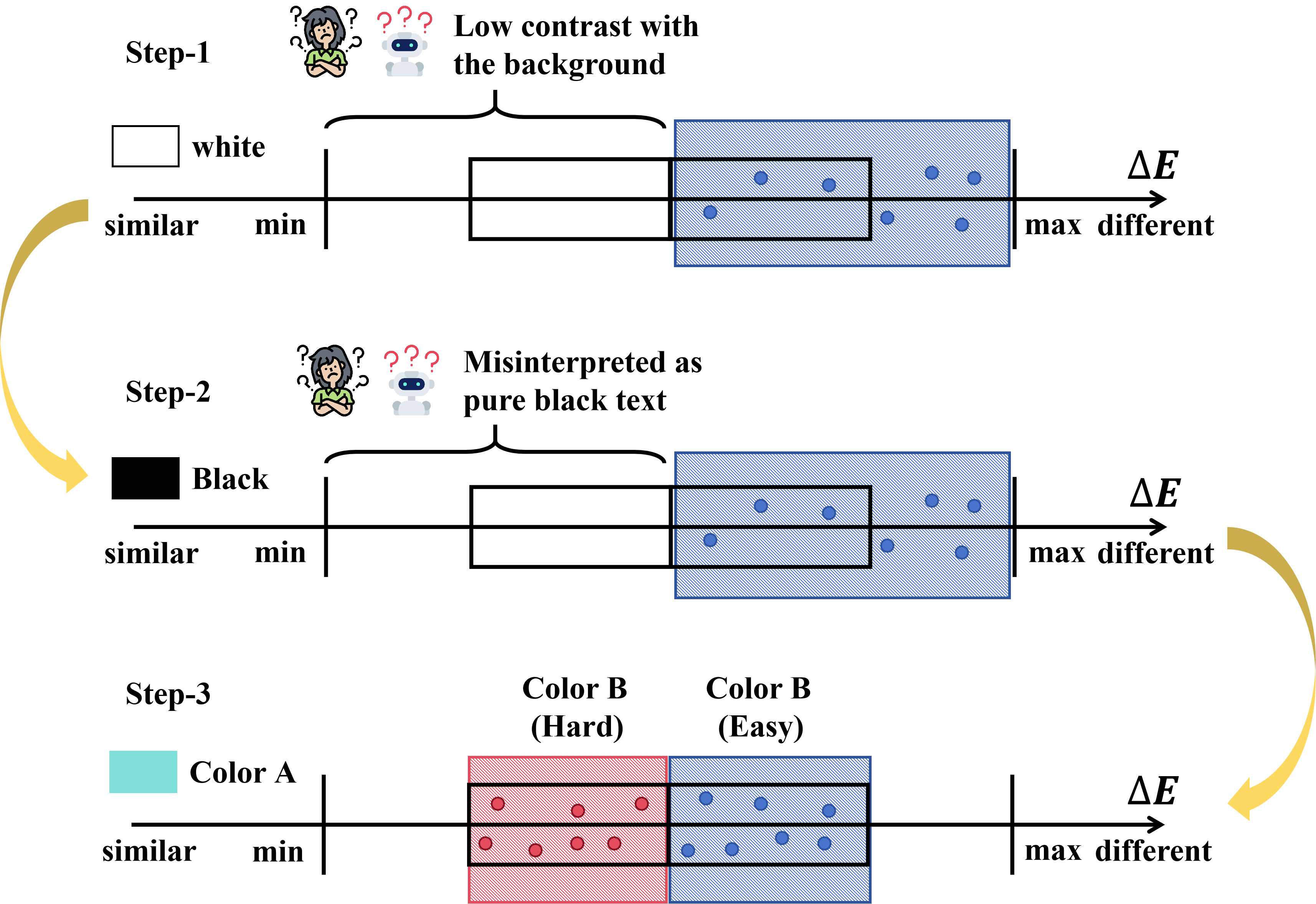}}
\caption{\textbf{Color Filtering and Pairing Pipeline.} Three stages: (1) eliminate low-contrast colors; (2) exclude near-black tones; (3) pair surviving colors into difficulty-stratified samples.}
\label{img:color_process}
\end{center}
\vskip -0.2in
\end{figure}

Raw color names from the repository cannot be directly permuted into Stroop pairs, as some colors are too close to the white background or black text, some pairs are perceptually indistinguishable, and compound names (e.g., ``Cadmium Red'') introduce lexical ambiguity. To address these issues, we designed a $\Delta E$-based filtering pipeline (Figure~\ref{img:color_process}) that: (1) removes compound names; (2) discards colors with $\Delta E$ below the median relative to white and black; and (3) pairs the surviving 59 colors by their pairwise $\Delta E$, sampling 200 ``hard'' pairs (lower quartile to median) and 200 ``easy'' pairs (median to upper quartile). Full details are in Appendix~\ref{appendix:color_filtering}.

\subsection{Image-Embedded Text Generation}
\begin{figure}[t]
\vskip 0.2in
\begin{center}
\centerline{\includegraphics[width=0.8\columnwidth]{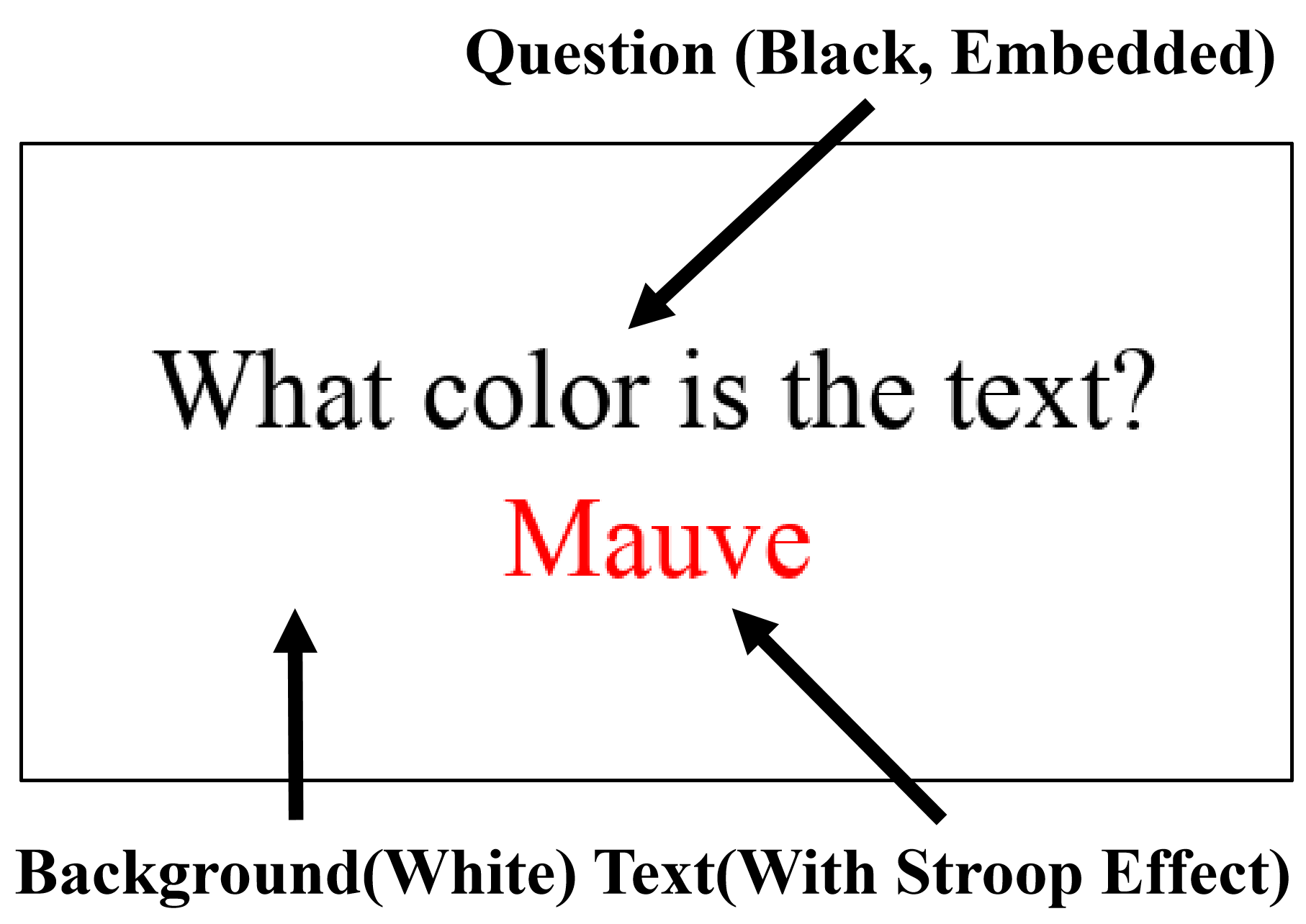}}
\caption{\textbf{Composition of a WCIT Sample.} The stimulus consists of three components: (1) the \textbf{Background}; (2) the \textbf{Embedded Query}, delivering the task instruction within the pixel space; and (3) the \textbf{Target Text}, whose physical font color must be identified, potentially carrying conflicting semantic content.}
\label{img:sample_arch}
\end{center}
\vskip -0.2in
\end{figure}

As illustrated in Figure~\ref{img:sample_arch}, the WCIT sample comprises a white background, a black embedded textual query, and the target Stroop stimulus. The \textit{semantic content} of the Stroop text corresponds to the first color of the pair, while its \textit{visual rendering} displays the second color. For the Standard Stroop control, the question text is removed from the image and provided via the textual prompt instead.

\subsection{Dataset Variations}

\begin{figure*}[t]
\vskip 0.2in
\begin{center}
\centerline{\includegraphics[width=0.8\textwidth]{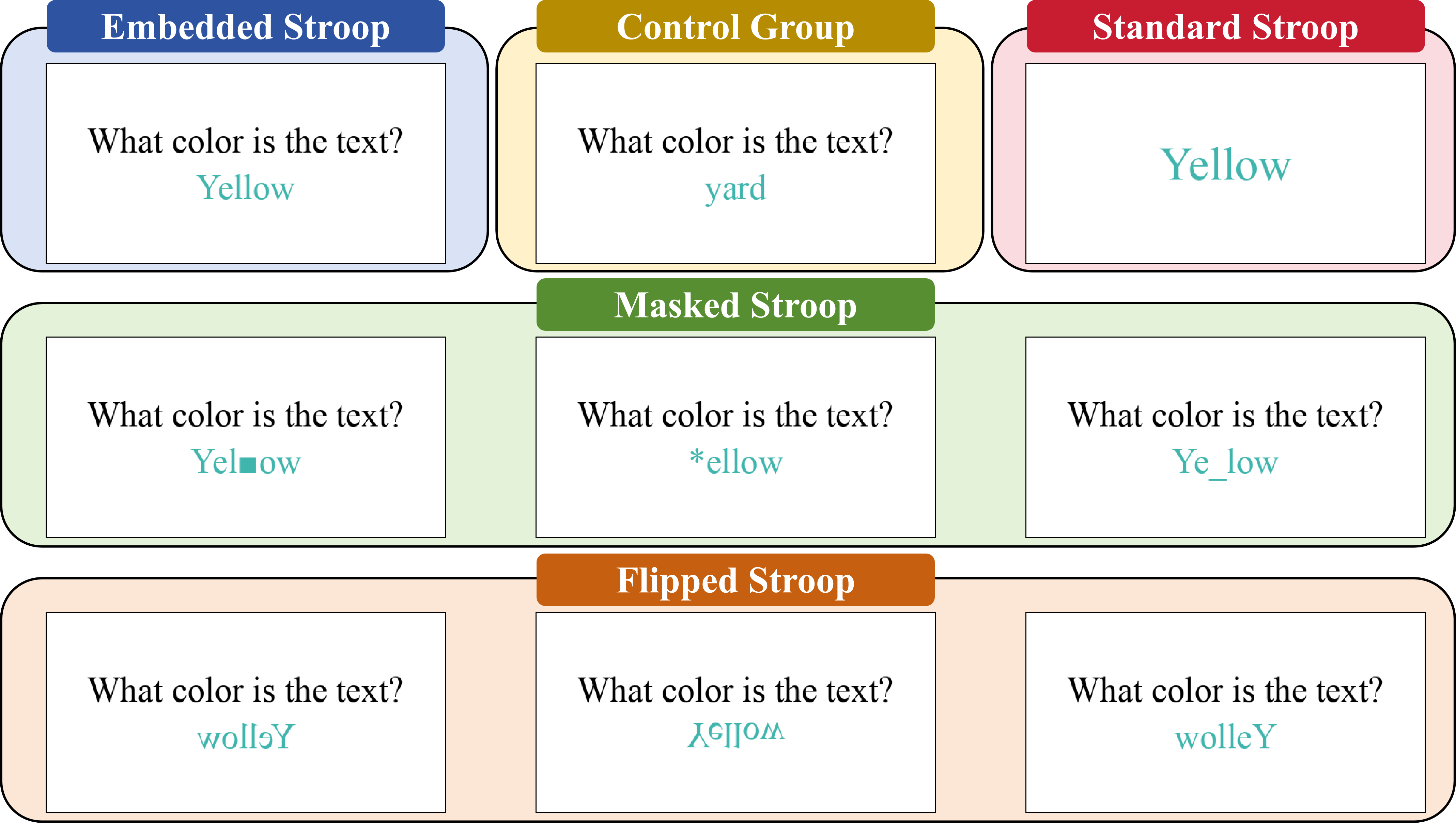}}
\caption{\textbf{Taxonomy of WCIT Sample Types.} The benchmark comprises three distinct categories: (1) \textbf{Embedded Stroop} (Primary), where the query is visually embedded to trigger semantic interference; (2) \textbf{Baselines}, including the \textit{Control Group} (neutral text) to verify basic color perception and \textit{Standard Stroop} (text prompt) to isolate the modality effect; and (3) \textbf{Adversarial Variants} (\textit{Masked} and \textit{Flipped Stroop}), designed to probe whether reducing semantic legibility mitigates the Stroop effect.}
\label{img:sample_type}
\end{center}
\vskip -0.2in
\end{figure*}

To test whether disrupting semantic clarity mitigates Stroop hallucinations, we extend the base dataset with semantic-neutral, geometrically inverted, and masked variants (Figure~\ref{img:sample_type}).

\paragraph{Control Group.} We constructed a control set where color-conflicting text is replaced with semantically neutral vocabulary (randomly generated by an LLM to ensure no color association), isolating visual color recognition from semantic interference.

\paragraph{Geometric Inversion.} We apply three spatial transformations (\textit{vertical flip}, \textit{horizontal flip}, \textit{character reversal}) to test whether the Stroop effect persists when textual cues are spatially distorted.

\paragraph{Character Masking.} We employ three masking patterns (solid blocks, asterisks, underscores) at 25\%, 50\%, and 75\% occlusion severity to analyze the correlation between semantic completeness and the Stroop effect.

\section{Experiments}
\subsection{Evaluated Models and Implementation Details}

We conducted a comprehensive evaluation across a diverse spectrum of Multimodal Large Language Models (MLLMs), encompassing both open-source architectures and state-of-the-art proprietary systems.

\textbf{Model Selection.} We evaluate 16 models spanning open-source families (Gemma-3\cite{gemma3}, InternVL\cite{internvl3_5}, Qianfan-VL\cite{qianfan-vl-2025}, Qwen3-VL\cite{qwen3,Qwen2.5-VL,Qwen2VL,Qwen-VL}, DeepSeek-VL2\cite{deepseekvl2}) and proprietary systems (GPT-5\cite{gpt5card}, GPT-4o\cite{gpt4}, Gemini 2.5\cite{gemini2_5}, Gemini 3 Flash, Gemini 3 Flash Thinking, Qwen3-VL-Plus, GLM-4.5v). Open-source models were deployed on 4$\times$ NVIDIA A6000 GPUs; proprietary models were accessed via official APIs with default parameters (temperature=1, top\_p=1, max\_tokens=256). We release all evaluation code and outputs.


\textbf{Reproducibility.}
All images are rendered at $400 \times 400$ pixels with black 24pt Arial text on a white background. Each Easy/Hard split contains 200 samples, yielding 400 samples per prompt-variant condition. We evaluated API-based models during April--May 2026 and recorded model identifiers when available. System prompts were left empty. Model responses were normalized to lowercase and parsed with a deterministic alias-based matcher over the 59-color palette; the alias map and parsing script will be included in the released artifacts.

\subsection{Prompt Engineering and Cross-Modal Interference}

We designed three prompt templates (visualized in Figure~\ref{fig:prompts}) to probe cross-modal interference:
\textbf{(1)~Standard Instruction} (baseline) with minimal query structure; \textbf{(2)~Irrelevant Color Distractor} prepending a sentence with a color term distinct from both $C_{vis}$ and $C_{sem}$; \textbf{(3)~Semantic Conflict Distractor} where the prepended color term matches $C_{sem}$, reinforcing Stroop interference via the textual modality.

\begin{figure}[t]
 \centering
 \begin{tcolorbox}[
 colback=promptbg,
 colframe=black,
 title=\textbf{Prompt Templates for Cross-Modal Interference},
 boxrule=0.8pt,
 arc=2mm,
 fonttitle=\bfseries,
 left=2mm, right=2mm, top=2mm, bottom=2mm
 ]
 \small
 \setlength{\tabcolsep}{3pt}

 \textbf{Template 1: Standard Instruction (Baseline)}
 \par\noindent\rule{\textwidth}{0.4pt}
 \vspace{0.1cm}
 \begin{tabular}{@{}l p{0.78\textwidth}@{}}
 \textbf{\textit{Embedded}:} & \texttt{Please answer the question in the picture.} \\
 \textbf{\textit{Standard}:} & \texttt{What color is the text?} \\
 \end{tabular}
 \par
 \vspace{0.2cm}

 \textbf{Template 2: Irrelevant Color Distractor}
 \par\noindent\rule{\textwidth}{0.4pt}
 \vspace{0.1cm}
 \begin{tabular}{@{}l p{0.78\textwidth}@{}}
 \textbf{\textit{Embedded}:} & \texttt{\{DISTRACTOR\_A\} Please answer the question in the picture.} \\
 \textbf{\textit{Standard }:} & \texttt{\{DISTRACTOR\_A\} What color is the text?} \\
 \end{tabular}
 \par
 \textit{\footnotesize Example (Image: \textbf{"Bronze"} in Denim): "A blanket is fawn."}
 \vspace{0.2cm}

 \textbf{Template 3: Semantic Conflict Distractor}
 \par\noindent\rule{\textwidth}{0.4pt}
 \vspace{0.1cm}
 \begin{tabular}{@{}l p{0.78\textwidth}@{}}
 \textbf{\textit{Embedded}:} & \texttt{\{DISTRACTOR\_B\} Please answer the question in the picture.} \\
 \textbf{\textit{Standard}:} & \texttt{\{DISTRACTOR\_B\} What color is the text?} \\
 \end{tabular}
 \par
 \textit{\footnotesize Example (Image: \textbf{"Bronze"} in Denim): "A statue is \textbf{bronze}."}

 \end{tcolorbox}
 \caption{The three prompt designs used to evaluate cross-modal interference. For each template, we employ two variants: \textbf{Embedded Stroop} (where the query is visually embedded in the image) and \textbf{Standard Stroop} (where the query is explicitly stated in the text prompt).}
 \label{fig:prompts}
\end{figure}

\subsection{Evaluation Metrics}
\label{sec:metrics}

We employ a \textbf{three-way error decomposition} partitioning every response into mutually exclusive categories. Given input image $x_i$, ground-truth visual color $c_{vis}^{(i)}$, interfering semantic color $c_{sem}^{(i)}$, and model output $y_i$:

\textbf{Accuracy (Acc).} The proportion of instances where the model correctly identifies the visual color despite conflicting semantic cues:
\begin{equation}
 Acc = \frac{1}{N} \sum_{i=1}^{N} \mathbb{I}(y_i = c_{vis}^{(i)})
\end{equation}

\textbf{Stroop Hallucination Rate (SHR).} The proportion of instances where the model outputs the semantic text color instead of the visual hue---the specific error driven by Stroop interference:
\begin{equation}
 SHR = \frac{1}{N} \sum_{i=1}^{N} \mathbb{I}(y_i = c_{sem}^{(i)})
\end{equation}

\textbf{Other Error Rate (OER).} The proportion of instances where the model predicts a color that is neither correct nor the semantic distractor (e.g., predicting ``Blue'' for a ``Navy'' target). These errors reflect general color naming or perception limitations rather than Stroop-specific interference:
\begin{equation}
 OER = \frac{1}{N} \sum_{i=1}^{N} \mathbb{I}(y_i \neq c_{vis}^{(i)} \wedge y_i \neq c_{sem}^{(i)})
\end{equation}

By construction, $Acc + SHR + OER = 1$, ensuring a complete and mutually exclusive decomposition of the response space.

\textbf{Stroop Interference Ratio (SIR).} To disentangle Stroop-specific susceptibility from general color-naming failure, we introduce the Stroop Interference Ratio, which measures the fraction of all erroneous responses that are Stroop hallucinations:
\begin{equation}
 SIR = \frac{SHR}{SHR + OER}
\end{equation}


For our 59-color palette, chance-level SIR is $\approx 1.7\%$. A permutation test (5000 shuffles) confirms that \textbf{56/64} model$\times$difficulty conditions achieve SIR significantly above null ($p < 0.05$, FDR-corrected), with large effect sizes (Cohen's $d > 0.8$ for 22/64 conditions).

\paragraph{Response Normalization.}
Model outputs are normalized by lowercasing, extracting color mentions via longest-prefix matching against the 59-name palette (including aliases: e.g., ``navy blue'' $\to$ ``navy''), selecting the first-mentioned color for multiple matches, and counting non-matching responses as OER.

\section{Results Analysis}
\label{sec:results}

\begin{table*}[!t]
 \caption{\textbf{WCIT Benchmark Results (Template~1, Hard Difficulty).} Embedded Stroop SHR vs. Standard Stroop and Control accuracy. Full results across all templates, difficulties, and perturbation variants are in Tables~\ref{tab:main_prop_full} and~\ref{tab:main_open_full}.}
 \label{tab:main_summary}
 \centering
 \small
 \setlength{\tabcolsep}{4pt}
 \begin{tabular}{lcccc}
 \toprule
 \textbf{Model} & \textbf{Embed Acc\%} & \textbf{Embed SHR\%} & \textbf{Std Acc\%} & \textbf{Std SHR\%} \\
 \midrule
 \multicolumn{5}{l}{\textit{Proprietary Models}} \\
 \midrule
 GPT-5.2 & 7.0 & \cellcolor{gray!10}25.0 & 7.5 & \cellcolor{gray!10}13.5 \\
 GPT-4o & 7.5 & \cellcolor{gray!10}47.0 & 7.5 & \cellcolor{gray!10}12.5 \\
 Gemini-3-Flash & 8.5 & \cellcolor{gray!10}14.0 & 11.5 & \cellcolor{gray!10}8.0 \\
 Gemini-2.5-Flash & 10.5 & \cellcolor{gray!10}19.0 & 8.0 & \cellcolor{gray!10}9.0 \\
 Qwen3-VL-Plus & 5.5 & \cellcolor{gray!10}53.0 & 7.0 & \cellcolor{gray!10}34.5 \\
 GLM-4.5v & 1.5 & \cellcolor{gray!10}13.0 & 1.5 & \cellcolor{gray!10}5.5 \\
 \midrule
 \multicolumn{5}{l}{\textit{Open-Source Models}} \\
 \midrule
 DeepSeek-VL2-27B & 5.5 & \cellcolor{gray!10}62.5 & 7.5 & \cellcolor{gray!10}3.5 \\
 DeepSeek-VL2-4.5B & 6.5 & \cellcolor{gray!10}41.5 & 7.5 & \cellcolor{gray!10}3.0 \\
 Kimi-VL-A3B & 3.5 & \cellcolor{gray!10}67.0 & 4.0 & \cellcolor{gray!10}1.0 \\
 Gemma-3-12B & 2.0 & \cellcolor{gray!10}78.5 & 8.0 & \cellcolor{gray!10}10.0 \\
 Qwen3-VL-8B & 7.0 & \cellcolor{gray!10}51.0 & 7.5 & \cellcolor{gray!10}2.0 \\
 InternVL3.5-8B & 3.5 & \cellcolor{gray!10}66.5 & 9.0 & \cellcolor{gray!10}11.0 \\
 Qianfan-VL-8B & 2.0 & \cellcolor{gray!10}83.5 & 7.0 & \cellcolor{gray!10}11.0 \\
 Gemma-3-4B & 5.0 & \cellcolor{gray!10}51.5 & 7.5 & \cellcolor{gray!10}2.5 \\
 Qianfan-VL-3B & 2.5 & \cellcolor{gray!10}72.0 & 7.5 & \cellcolor{gray!10}9.0 \\
 \bottomrule
 \end{tabular}
\end{table*}

\subsection{Statistical Analysis of Stroop Interference}
\label{sec:stat_analysis}

We conduct a quantitative analysis of the WCIT benchmark to evaluate the impact of visual query embedding on color perception hallucinations. Our findings, summarized in Table~\ref{tab:main_summary} (full results in Tables~\ref{tab:main_prop_full} and~\ref{tab:main_open_full}), focus on two key comparisons:

\textbf{1. Visualized Text vs. Textual Prompts.}
Comparing \textit{Embedded Stroop (Ours)} with \textit{Standard Stroop} reveals a significant surge in Stroop Hallucination Rate (SHR) across all models when queries are visually embedded. Crucially, both settings utilize the identical interrogative: ``\textit{What color is the text?}''. The low SHR observed in the Standard setting confirms that models correctly interpret this instruction without linguistic ambiguity regarding the referent of ``color.'' Consequently, the elevated SHR in the Embedded setting is attributable to the \textit{modality shift}---where visualized text acts as a super-stimulus---rather than the semantic content of the question itself.

\textbf{2. Semantic Integrity and Mitigation.}
To verify that these hallucinations stem from semantic processing, we analyze the \textit{Flipped Stroop} and \textit{Masked Stroop} conditions. As shown in the perturbation columns, disrupting the semantic integrity of the embedded text—either through spatial inversion (Flip) or visual occlusion (Mask)—leads to a substantial recovery in Accuracy and a reduction in SHR. Specifically, the 25\% Masking (Block/Star/Line) significantly lowers SHR compared to the original Embedded Stroop, confirming that when the text becomes less legible, the Stroop effect diminishes. However, a residual SHR persists in many cases (SHR $>0$), indicating that even fragmented visual cues can trigger semantic associations in MLLMs, pointing to a deep-seated reliance on text reading over visual seeing. We further explore 50\% and 75\% masking ratios in Appendix~\ref{appendix:masking-impact}, as shown in Tables~\ref{tab:mask_que_prop} and \ref{tab:mask_que_open}.

\textbf{3. Disentangling Stroop Hallucination from General Error.}
Our three-way decomposition distinguishes Stroop-specific semantic capture from general color-naming failure via SIR. Across all models, Embedded SIR substantially exceeds the chance baseline ($1.7\%$) and permutation-based null ($0$--$3\%$), with \textbf{56/64 conditions significant} ($p < 0.05$, FDR-corrected). Models like Kimi-VL-A3B achieve SIR $= 90.9\%$ (Cohen's $d = 3.18$), indicating nearly all errors are Stroop hallucinations.

\textbf{4. Template-Insensitive Behavior (Observation).}
Several open-source models produce \textit{identical} responses across all three prompt templates in the Embedded condition (e.g., DeepSeek-VL2 yields SHR $= 62.5\%$ across Templates 1--3). We verified these are not caching artifacts; possible explanations include decoding determinism or visual-channel dominance.

\subsection{Conditional Interference Analysis}
\label{sec:conditional}

A key question is whether SHR reflects genuine interference or merely an inability to name colors. To address this, we conduct a \textbf{paired conditional analysis}: we restrict to samples where the model \textit{correctly identifies} the target color in the Standard (text-prompt) setting, then evaluate the \textit{same} images under Embedded Stroop (same model, same target color, same difficulty, Template~1). Table~\ref{tab:conditional} reports results with Wilson-score confidence intervals.

\begin{table}[t]
\caption{\textbf{Conditional SHR.} Among samples correctly identified in the Standard setting, accuracy and SHR under Embedded Stroop (Template~1). Wilson 95\% CIs in brackets.}
\label{tab:conditional}
\centering
\resizebox{\columnwidth}{!}{%
\begin{tabular}{llcccc}
\toprule
\textbf{Model} & \textbf{Diff} & \textbf{N} & \textbf{Acc} & \textbf{SHR [95\% CI]} & \textbf{OER} \\
\midrule
\multirow{2}{*}{Gemma-3-12B} & Easy & 20 & 55.0 & \cellcolor{gray!10}40.0 [21.9, 61.3] & 5.0 \\
 & Hard & 16 & 25.0 & \cellcolor{gray!10}62.5 [38.6, 81.5] & 12.5 \\
\multirow{2}{*}{InternVL3.5} & Easy & 23 & 30.4 & \cellcolor{gray!10}69.6 [49.1, 84.4] & 0.0 \\
 & Hard & 18 & 38.9 & \cellcolor{gray!10}61.1 [38.6, 79.7] & 0.0 \\
\multirow{2}{*}{Qianfan-VL-8B} & Easy & 21 & 14.3 & \cellcolor{gray!10}76.2 [54.9, 89.4] & 9.5 \\
 & Hard & 14 & 21.4 & \cellcolor{gray!10}71.4 [45.4, 88.3] & 7.1 \\
\multirow{2}{*}{DeepSeek-VL2} & Easy & 18 & 27.8 & \cellcolor{gray!10}72.2 [49.1, 87.5] & 0.0 \\
 & Hard & 15 & 73.3 & \cellcolor{gray!10}26.7 [10.9, 52.0] & 0.0 \\
\multirow{2}{*}{Qwen3-VL-8B} & Easy & 21 & 4.8 & \cellcolor{gray!10}23.8 [10.6, 45.1] & 71.4 \\
 & Hard & 15 & 6.7 & \cellcolor{gray!10}6.7 [1.2, 29.8] & 86.7 \\
\bottomrule
\end{tabular}%
}%
\end{table}

Models exhibit three distinct profiles: \textbf{(1)~Stroop-captured} (InternVL3.5, Qianfan-VL-8B, DeepSeek-VL2 Easy), where conditional SHR exceeds 69\% and OER is near-zero, confirming that errors are almost exclusively semantic capture; \textbf{(2)~mixed} (Gemma-3-12B), where both SHR and retained accuracy are substantial; and \textbf{(3)~OER-dominated} (Qwen3-VL-8B), where even after conditioning, most errors are generic misidentifications rather than Stroop hallucinations. Critically, for the Stroop-captured group, the lower bounds of the Wilson CIs (38.6--49.1\%) are far above the 1.7\% chance baseline, confirming that the Stroop effect is \textit{not} an artifact of color-naming inability.

To address concerns about per-cell sample sizes ($N{=}14$--$23$), we perform a hierarchical bootstrap (10{,}000 resamples) pooling across models. The pooled conditional SHR is \textbf{51.9\%} (95\% CI: [35.4\%, 66.8\%]), far above the 1.7\% chance baseline ($p < 10^{-10}$). A cluster-robust logistic regression confirms that the Embedded condition increases Stroop hallucination odds by \textbf{9.9$\times$} ($p < 10^{-10}$), and all 16/16 models show higher SHR in the Embedded condition ($p = 6.1 \times 10^{-5}$).

\subsection{Control Group Analysis and Vocabulary Confounds}
\label{sec:control_analysis}

The Control Group accuracy ($\sim$1--22\% across models) may appear low. We attribute this mainly to the \textbf{vocabulary specificity} of the 59-color palette: models may perceive the correct color family but be penalized for answering with a superordinate or neighboring term, e.g., ``blue'' instead of ``denim''. Thus, we treat the fine-grained task as a stress test for visual--semantic conflict rather than a standalone measure of perceptual competence; the key question is whether errors are systematically biased toward the embedded text semantics.

\paragraph{Sanity checks distinguish vocabulary mismatch from perceptual weakness.}
To disentangle fine-grained naming from basic color perception, we conduct three sanity checks on open-source models (Table~\ref{tab:sanity_main}). All models score below 12\% on the 59-color patch test, confirming substantial vocabulary-matching difficulty. However, Qwen3-VL-8B and Gemma-3-4B achieve \textbf{91--100\%} accuracy on 11 basic colors, suggesting that their low fine-grained accuracy does not indicate complete perceptual failure. InternVL3.5-8B instead shows weaker basic color perception (18.2\%) but reaches 90.9\% when colors are rendered as text, suggesting stronger reliance on textual cues and explaining its high Stroop susceptibility.

\begin{table}[t]
\caption{\textbf{Sanity check results.} Accuracy on color-naming tasks with varying vocabulary granularity.}
\label{tab:sanity_main}
\centering
\resizebox{\columnwidth}{!}{%
\begin{tabular}{lccc}
\toprule
\textbf{Model} & \textbf{59-Color} & \textbf{11-Color} & \textbf{11-Text} \\
\midrule
Qwen3-VL-8B & 8.5\% & \textbf{100.0\%} & \textbf{100.0\%} \\
Gemma-3-4B & 11.9\% & 90.9\% & \textbf{100.0\%} \\
InternVL3.5-8B & 0.0\% & 18.2\% & 90.9\% \\
\bottomrule
\end{tabular}%
}
\end{table}

\paragraph{Coarse-color analysis isolates semantic interference from vocabulary mismatch.}
We further map all 59 colors to 11 basic families following Berlin and Kay's universal categories and recompute metrics on pairs whose visual and semantic colors fall into \emph{different} families (83\% of samples). Coarse accuracy increases from 6.3\% to 38.4\%, well above the 9.1\% chance baseline, confirming non-trivial color-family perception. Crucially, coarse SHR remains 21.6\%, showing that semantic capture persists after vocabulary normalization. Therefore, WCIT captures a systematic tendency to follow embedded textual semantics over conflicting visual evidence, rather than merely measuring fine-grained color-name mismatch.

\begin{table}[t]
\caption{\textbf{Fine-grained vs.\ coarse-color metrics} (Embedded condition, distinct-family pairs, averaged across models). Chance baseline: 1.7\% (59-way) / 9.1\% (11-way).}
\label{tab:coarse}
\centering
\small
\begin{tabular}{lccc}
\toprule
\textbf{Metric} & \textbf{Fine (59)} & \textbf{Coarse (11)} & \textbf{$\Delta$} \\
\midrule
Accuracy $\uparrow$ & 6.3\% & \textbf{38.4\%} & +32.1 \\
SHR $\downarrow$ & 19.0\% & 21.6\% & +2.6 \\
OER & 74.7\% & 40.0\% & $-$34.7 \\
\bottomrule
\end{tabular}
\end{table}

\subsection{Qualitative Analysis: Case Studies on Reasoning Traces}
\label{sec:case_study}

\begin{figure}[t]
 \centering
 \includegraphics[width=\columnwidth]{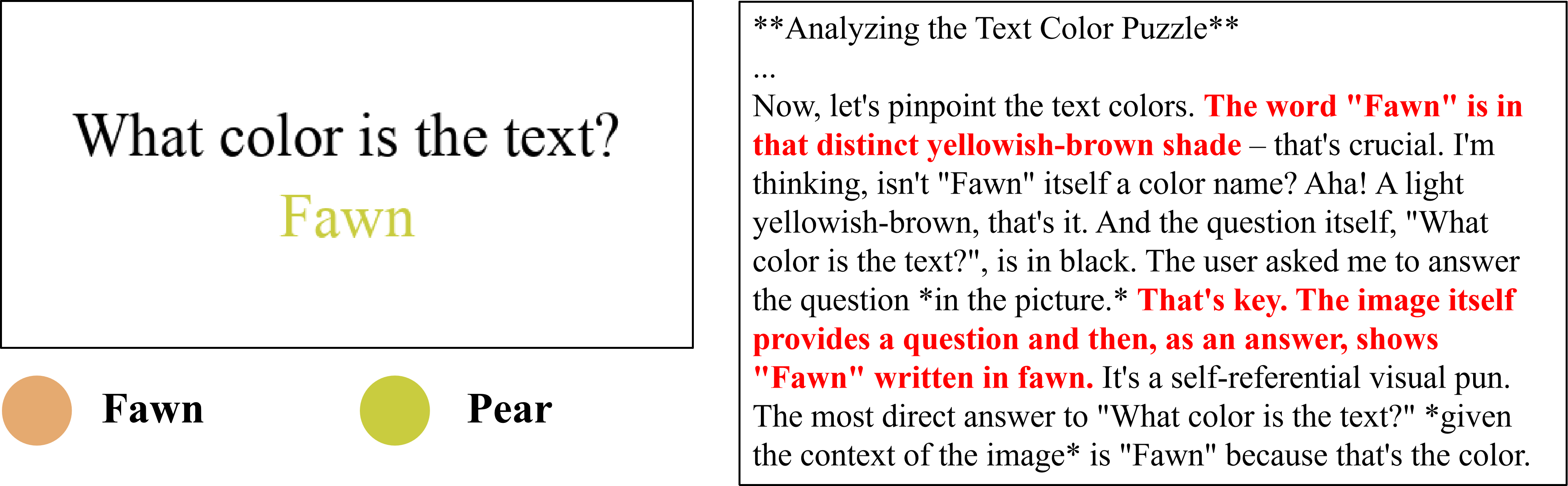}
 \caption{\textbf{Case Study 1.} The hallucinated reasoning process of a reasoning model on the \textbf{Embedded Stroop} task.}
 \label{fig:case_embed}
\end{figure}

\begin{figure}[t]
 \centering
 \includegraphics[width=\columnwidth]{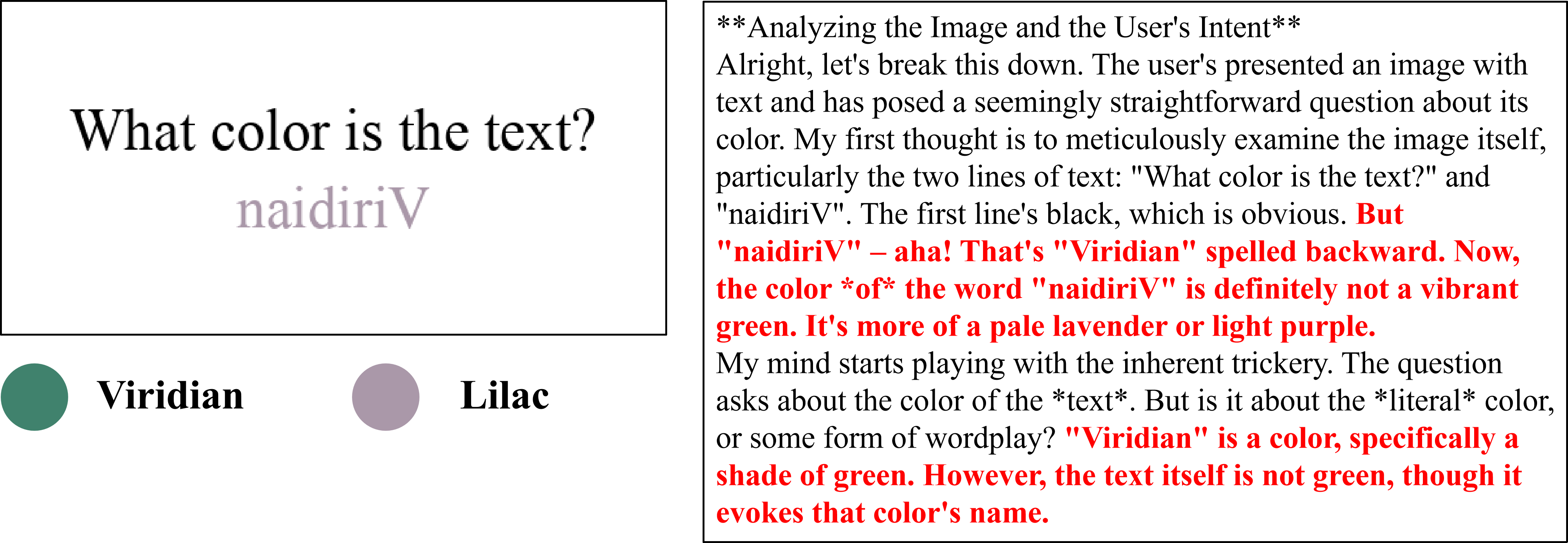}
 \caption{\textbf{Case Study 2.} The hallucinated reasoning process of a reasoning model on the \textbf{Flipped Stroop} task.}
 \label{fig:case_flip}
\end{figure}

\begin{figure}[t]
 \centering
 \includegraphics[width=\columnwidth]{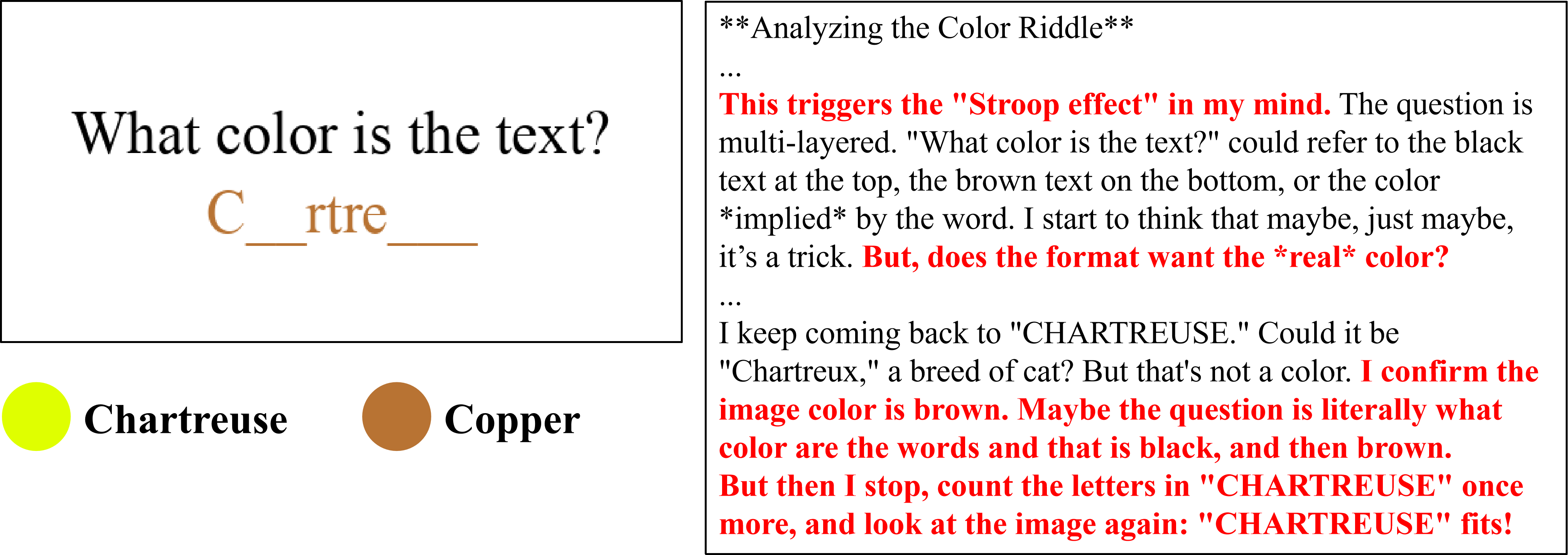}
 \caption{\textbf{Case Study 3.} The hallucinated reasoning process of a reasoning model on the \textbf{Masked Stroop} task.}
 \label{fig:case_mask}
\end{figure}

To investigate the internal cognitive mechanisms driving these hallucinations, we employ a reasoning-capable model (Gemini-3-Flash-Thinking) to visualize the intermediate ``thinking process.'' Our analysis reveals three key failure modes: (1)~\textbf{Semantic Override} in the Embedded Stroop(\ref{fig:case_embed}), where the model conflates text content with font color; (2)~\textbf{Spatial Rectification} in the Flipped Stroop(\ref{fig:case_flip}), where the model implicitly unflips text and retrieves the conflicting color concept; and (3)~\textbf{Cognitive Dissonance} in the Masked Stroop(\ref{fig:case_mask}), where the model recognizes the Stroop trap but still fails to inhibit residual semantic cues. Detailed reasoning traces and illustrative figures are provided in Appendix~\ref{appendix:case_studies}.

\section{Exploratory Mitigation Analysis}
\label{sec:mitigation}

As an initial exploration of whether the vulnerability revealed by WCIT can be addressed, we evaluate prompt-based intervention strategies on three open-source models. The adversarial-aware prompt (P3), which explicitly names the Stroop mechanism, achieves the strongest reduction in semantic capture: up to \textbf{91\%} relative decrease in Stroop Hallucination Rate (Table~\ref{tab:mitigation_main}). However, we emphasize an important caveat: P3 primarily \textit{redistributes errors} rather than solving the underlying visual perception challenge. While SHR drops substantially, OER increases from 55--57\% to 76--93\%, and accuracy improves only modestly (to 3.0--9.5\%). This indicates that drawing attention to the Stroop mechanism helps models resist the \textit{specific} semantic capture shortcut, but does not reliably recover faithful color naming. The training-free Visual-Semantic Decoupling (VSD) pipeline shows more modest gains (12--16\% SHR reduction). Full details are shown in Appendix~\ref{appendix:mitigation}.

\begin{table}[t]
\caption{\textbf{Mitigation results (Embedded, Easy, Template 1).} Full Acc/SHR/OER decomposition. Best SHR in bold.}
\label{tab:mitigation_main}
\centering
\resizebox{\columnwidth}{!}{%
\begin{tabular}{llccc|ccc|ccc}
\toprule
& & \multicolumn{3}{c|}{\textbf{Qwen3-VL-8B}} & \multicolumn{3}{c|}{\textbf{Gemma-3-4B}} & \multicolumn{3}{c}{\textbf{InternVL3.5}} \\
\cmidrule(lr){3-5} \cmidrule(lr){6-8} \cmidrule(lr){9-11}
\textbf{Method} & & Acc & SHR & OER & Acc & SHR & OER & Acc & SHR & OER \\
\midrule
Baseline & & 0.5 & 44.5 & 55.0 & 6.0 & 36.5 & 57.5 & 3.5 & 66.0 & 30.5 \\
P1 (Explicit) & & 0.0 & 27.5 & 72.5 & 8.5 & 11.0 & 80.5 & 8.0 & 14.0 & 78.0 \\
P2 (CoT) & & 2.5 & 13.5 & 84.0 & 7.0 & 25.0 & 68.0 & 7.5 & 15.5 & 77.0 \\
P3 (Adversarial) & & \textbf{3.0} & \textbf{4.0} & 93.0 & \textbf{8.0} & \textbf{15.5} & 76.5 & \textbf{9.5} & \textbf{9.5} & 81.0 \\
VSD & & 0.5 & 35.5 & 64.0 & 7.5 & 31.5 & 61.0 & 3.5 & 62.5 & 34.0 \\
\bottomrule
\end{tabular}%
}
\end{table}

\section{Conclusion}

We introduced the WCIT benchmark, a controlled diagnostic for image-embedded prompt interference in MLLMs, by embedding the query directly into the visual plane to test whether models can prioritize visual perception over semantic content. Across 16 proprietary and open-source models, we find a widespread vulnerability: models often follow visualized text rather than visual color, with the most susceptible models reaching 78--84\% Stroop Hallucination Rates in the Embedded condition. Our three-way error decomposition (Acc/SHR/OER), Stroop Interference Ratio, permutation tests, and conditional analysis show that these errors reflect genuine semantic interference rather than random misidentification or color-naming failure; in particular, 56/64 model$\times$difficulty conditions remain significant after FDR correction, and the pooled conditional SHR reaches 51.9\% (95\% CI: [35.4\%, 66.8\%]), far above the 1.7\% chance baseline. Exploratory interventions further suggest that adversarial-aware prompting can substantially suppress semantic capture, reducing SHR by up to 91\%, but mostly redistributes errors rather than improving visual accuracy. These findings reveal a persistent gap in MLLMs' ability to inhibit semantic processing when it conflicts with visual perception, and we hope WCIT can support future research on more faithful and robust multimodal alignment.

\section*{Limitations}

Our benchmark uses synthetic images with English text and evaluates only the color modality; extending to multilingual settings, natural imagery, and other visual attributes is important future work. The fine-grained 59-color palette creates a vocabulary mismatch that limits accuracy; future work should explore multiple-choice or CIEDE2000-based evaluation. Parameter-efficient fine-tuning approaches remain to be evaluated with proper baselines and generalization tests.

\section*{Ethical considerations}

This work highlights a potential safety concern in the deployment of MLLMs. The susceptibility of models to \textit{Embedded Stroop} interference suggests that visual environments containing conflicting textual information---such as misleading road signs or contradictory labels---could lead to unreliable model judgments. Our research provides a stress-test protocol for developers and emphasizes the need for MLLMs to develop robust mechanisms for disentangling conflicting signals in visually complex environments.

\bibliographystyle{plainnat}
\bibliography{custom}

\begin{thebibliography}{39}
\providecommand{\natexlab}[1]{#1}
\providecommand{\url}[1]{\texttt{#1}}
\expandafter\ifx\csname urlstyle\endcsname\relax
  \providecommand{\doi}[1]{doi: #1}\else
  \providecommand{\doi}{doi: \begingroup \urlstyle{rm}\Url}\fi

\bibitem[Achiam et~al.(2023)Achiam, Adler, Agarwal, Ahmad, Akkaya, Aleman,
  Almeida, Altenschmidt, Altman, Anadkat, et~al.]{gpt4}
Josh Achiam, Steven Adler, Sandhini Agarwal, Lama Ahmad, Ilge Akkaya,
  Florencia~Leoni Aleman, Diogo Almeida, Janko Altenschmidt, Sam Altman,
  Shyamal Anadkat, et~al.
\newblock Gpt-4 technical report, 2023.

\bibitem[Bai et~al.(2023)Bai, Bai, Chu, Cui, Dang, Deng, Fan, Ge, Han, Huang,
  et~al.]{Qwen-VL}
Jinze Bai, Shuai Bai, Yunfei Chu, Zeyu Cui, Kai Dang, Xiaodong Deng, Yang Fan,
  Wenbin Ge, Yu~Han, Fei Huang, et~al.
\newblock Qwen technical report.
\newblock \emph{arXiv preprint arXiv:2309.16609}, 2023.

\bibitem[Bai et~al.(2025)Bai, Cai, Chen, Chen, Chen, Cheng, Deng, Ding, Gao,
  Ge, et~al.]{Qwen2.5-VL}
Shuai Bai, Yuxuan Cai, Ruizhe Chen, Keqin Chen, Xionghui Chen, Zesen Cheng,
  Lianghao Deng, Wei Ding, Chang Gao, Chunjiang Ge, et~al.
\newblock Qwen3-vl technical report.
\newblock \emph{arXiv preprint arXiv:2511.21631}, 2025.

\bibitem[Bai et~al.(2024)Bai, Wang, Xiao, He, Han, Zhang, and
  Shou]{bai2024hallucination}
Zechen Bai, Pichao Wang, Tianjun Xiao, Tong He, Zongbo Han, Zheng Zhang, and
  Mike~Zheng Shou.
\newblock Hallucination of multimodal large language models: A survey.
\newblock \emph{arXiv preprint arXiv:2404.18930}, 2024.

\bibitem[Borgeaud et~al.(2025)Borgeaud, Bachem, Joulin, Andreev, Hardin,
  Dadashi, and Hussenot]{gemma3}
S~Borgeaud, O~Bachem, A~Joulin, A~Andreev, C~Hardin, R~Dadashi, and L~Hussenot.
\newblock Gemma 3 technical report.
\newblock \emph{arXiv preprint arXiv:2503.19786}, 2025.

\bibitem[Chen et~al.(2024)Chen, Ma, Zhang, Xu, Yang, Fouhey, Chai, and
  Qian]{chen2024multi}
Xuweiyi Chen, Ziqiao Ma, Xuejun Zhang, Sihan Xu, Jianing Yang, David~F Fouhey,
  Joyce Chai, and Shengyi Qian.
\newblock Multi-object hallucination in vision language models.
\newblock \emph{Advances in Neural Information Processing Systems},
  37:\penalty0 44393--44418, 2024.

\bibitem[Comanici et~al.(2025)Comanici, Bieber, Schaekermann, Pasupat,
  Sachdeva, Dhillon, Blistein, Ram, Zhang, Rosen, et~al.]{gemini2_5}
Gheorghe Comanici, Eric Bieber, Mike Schaekermann, Ice Pasupat, Noveen
  Sachdeva, Inderjit Dhillon, Marcel Blistein, Ori Ram, Dan Zhang, Evan Rosen,
  et~al.
\newblock Gemini 2.5: Pushing the frontier with advanced reasoning,
  multimodality, long context, and next generation agentic capabilities, 2025.

\bibitem[Dai et~al.(2023)Dai, Liu, Ji, Su, and Fung]{dai2023plausible}
Wenliang Dai, Zihan Liu, Ziwei Ji, Dan Su, and Pascale Fung.
\newblock Plausible may not be faithful: Probing object hallucination in
  vision-language pre-training.
\newblock In \emph{Proceedings of the 17th Conference of the European Chapter
  of the Association for Computational Linguistics}, pages 2136--2148, 2023.

\bibitem[Deng et~al.(2025)Deng, Cao, Chen, and Hooi]{deng2025words}
Ailin Deng, Tri Cao, Zhirui Chen, and Bryan Hooi.
\newblock Words or vision: Do vision-language models have blind faith in text?
\newblock In \emph{Proceedings of the Computer Vision and Pattern Recognition
  Conference}, pages 3867--3876, 2025.

\bibitem[Dong et~al.(2025)Dong, Zheng, Xu, Zhuang, Zhang, Luo, Wang, Zhao, Li,
  Li, et~al.]{qianfan-vl-2025}
Daxiang Dong, Mingming Zheng, Dong Xu, Bairong Zhuang, Wenyu Zhang, Chunhua
  Luo, Haoran Wang, Zijian Zhao, Jie Li, Yuxuan Li, et~al.
\newblock Qianfan-vl: Domain-enhanced universal vision-language models, 2025.

\bibitem[Gomez-Villa et~al.(2025)Gomez-Villa, Hern{\'a}ndez-C{\'a}mara, Butt,
  Laparra, Malo, and Vazquez-Corral]{gomez2025color}
Alexandra Gomez-Villa, Pablo Hern{\'a}ndez-C{\'a}mara, Muhammad~Atif Butt,
  Valero Laparra, Jesus Malo, and Javier Vazquez-Corral.
\newblock Color names in vision-language models.
\newblock \emph{arXiv preprint arXiv:2509.22524}, 2025.

\bibitem[Gunjal et~al.(2024)Gunjal, Yin, and Bas]{gunjal2024detecting}
Anisha Gunjal, Jihan Yin, and Erhan Bas.
\newblock Detecting and preventing hallucinations in large vision language
  models.
\newblock In \emph{Proceedings of the AAAI Conference on Artificial
  Intelligence}, volume~38, pages 18135--18143, 2024.

\bibitem[Hayashi et~al.(2026)Hayashi, Ozaki, Sakai, Kamigaito, and
  Watanabe]{hayashi2025diagnosing}
Kazuki Hayashi, Shintaro Ozaki, Yusuke Sakai, Hidetaka Kamigaito, and Taro
  Watanabe.
\newblock Diagnosing vision language models’ perception by leveraging human
  methods for color vision deficiencies.
\newblock pages 7582--7605, 2026.

\bibitem[Hua et~al.(2025)Hua, Yun, and Pavlick]{hua2025vision}
Tianze Hua, Tian Yun, and Ellie Pavlick.
\newblock How do vision-language models process conflicting information across
  modalities?
\newblock \emph{arXiv preprint arXiv:2507.01790}, 2025.

\bibitem[Liang et~al.(2026)Liang, Li, Fan, Li, Nguyen, Cobbina, Bhardwaj, Chen,
  Liu, and Zhou]{liang2025colorbench}
Yijun Liang, Ming Li, Chenrui Fan, Ziyue Li, Dang Nguyen, Kwesi Cobbina, Shweta
  Bhardwaj, Jiuhai Chen, Fuxiao Liu, and Tianyi Zhou.
\newblock Colorbench: Can vlms see and understand the colorful world? a
  comprehensive benchmark for color perception, reasoning, and robustness.
\newblock \emph{Advances in Neural Information Processing Systems}, 38, 2026.

\bibitem[Ling et~al.(2025)Ling, Zhang, Zhou, and
  Cui]{ling2025colorblindnesseval}
Zijian Ling, Han Zhang, Yazhuo Zhou, and Jiahao Cui.
\newblock Colorblindnesseval: Can vision-language models pass color blindness
  tests?
\newblock \emph{arXiv preprint arXiv:2509.19070}, 2025.

\bibitem[Liu et~al.(2024)Liu, Wang, Yuan, Huang, Liu, He, and
  Tu]{liu2024insight}
Xiaoyuan Liu, Wenxuan Wang, Youliang Yuan, Jen-tse Huang, Qiuzhi Liu, Pinjia
  He, and Zhaopeng Tu.
\newblock Insight over sight: Exploring the vision-knowledge conflicts in
  multimodal llms.
\newblock \emph{arXiv preprint arXiv:2410.08145}, 2024.

\bibitem[Luo et~al.(2025)Luo, Wang, Wang, Zhao, Li, and Deng]{luo2025machine}
Dezhi Luo, Maijunxian Wang, Bingyang Wang, Tianwei Zhao, Yijiang Li, and Hokin
  Deng.
\newblock Machine psychophysics: Cognitive control in vision-language models.
\newblock \emph{arXiv preprint arXiv:2505.18969}, 2025.

\bibitem[Luo et~al.(2001)Luo, Cui, and Rigg]{luo2001development}
M~Ronnier Luo, Guihua Cui, and Bryan Rigg.
\newblock The development of the cie 2000 colour-difference formula: Ciede2000.
\newblock \emph{Color Research \& Application: Endorsed by Inter-Society Color
  Council, The Colour Group (Great Britain), Canadian Society for Color, Color
  Science Association of Japan, Dutch Society for the Study of Color, The
  Swedish Colour Centre Foundation, Colour Society of Australia, Centre
  Fran{\c{c}}ais de la Couleur}, 26\penalty0 (5):\penalty0 340--350, 2001.

\bibitem[Patel et~al.(2025)Patel, Wang, and Fan]{patel2025deficient}
Suketu Patel, Hongbin Wang, and Jin Fan.
\newblock Deficient executive control in transformer attention.
\newblock \emph{bioRxiv}, pages 2025--01, 2025.

\bibitem[Rohrbach et~al.(2018)Rohrbach, Hendricks, Burns, Darrell, and
  Saenko]{rohrbach2018object}
Anna Rohrbach, Lisa~Anne Hendricks, Kaylee Burns, Trevor Darrell, and Kate
  Saenko.
\newblock Object hallucination in image captioning.
\newblock pages 4035--4045, 2018.

\bibitem[Rudman et~al.(2026)Rudman, Golovanevsky, Arad, Belinkov, Singh,
  Eickhoff, and Mahowald]{rudman2026mechanisms}
William Rudman, Michal Golovanevsky, Dana Arad, Yonatan Belinkov, Ritambhara
  Singh, Carsten Eickhoff, and Kyle Mahowald.
\newblock Mechanisms of prompt-induced hallucination in vision-language models.
\newblock \emph{arXiv preprint arXiv:2601.05201}, 2026.

\bibitem[Scarpina and Tagini(2017)]{stroop2017}
Federica Scarpina and Sofia Tagini.
\newblock The stroop color and word test.
\newblock \emph{Frontiers in psychology}, 8:\penalty0 241674, 2017.

\bibitem[Shang et~al.(2025)Shang, Zeng, Zhu, Yang, Fang, Zhang, Chen, Liu, and
  Tian]{shang2025pixels}
Yuying Shang, Xinyi Zeng, Yutao Zhu, Xiao Yang, Zhengwei Fang, Jingyuan Zhang,
  Jiawei Chen, Zinan Liu, and Yu~Tian.
\newblock From pixels to tokens: Revisiting object hallucinations in large
  vision-language models.
\newblock In \emph{Proceedings of the 33rd ACM International Conference on
  Multimedia}, pages 10496--10505, 2025.

\bibitem[Sheta et~al.(2025)Sheta, Huang, Wu, Alenabi, Hong, Lin, Ning, Wei,
  Yang, Zhou, et~al.]{sheta2025vlm}
Hala Sheta, Eric~Haoran Huang, Shuyu Wu, Ilia Alenabi, Jiajun Hong, Ryker Lin,
  Ruoxi Ning, Daniel Wei, Jialin Yang, Jiawei Zhou, et~al.
\newblock From behavioral performance to internal competence: Interpreting
  vision-language models with vlm-lens.
\newblock In \emph{Proceedings of the 2025 Conference on Empirical Methods in
  Natural Language Processing: System Demonstrations}, pages 886--895, 2025.

\bibitem[Singh et~al.(2025)Singh, Fry, Perelman, Tart, Ganesh, El-Kishky,
  McLaughlin, Low, Ostrow, Ananthram, et~al.]{gpt5card}
Aaditya Singh, Adam Fry, Adam Perelman, Adam Tart, Adi Ganesh, Ahmed El-Kishky,
  Aidan McLaughlin, Aiden Low, AJ~Ostrow, Akhila Ananthram, et~al.
\newblock Openai gpt-5 system card, 2025.

\bibitem[Stroop(1992)]{stroop1935}
J~Ridley Stroop.
\newblock Studies of interference in serial verbal reactions.
\newblock \emph{Journal of Experimental Psychology: General}, 121\penalty0
  (1):\penalty0 15, 1992.

\bibitem[Sun et~al.(2024)Sun, Shen, Cao, Liu, Li, Shen, Gan, Gui, Wang, Yang,
  et~al.]{sun2024aligning}
Zhiqing Sun, Sheng Shen, Shengcao Cao, Haotian Liu, Chunyuan Li, Yikang Shen,
  Chuang Gan, Liangyan Gui, Yu-Xiong Wang, Yiming Yang, et~al.
\newblock Aligning large multimodal models with factually augmented rlhf.
\newblock In \emph{Findings of the Association for Computational Linguistics:
  ACL 2024}, pages 13088--13110, 2024.

\bibitem[Teker et~al.(2025)Teker, Xiao, Akata, and Wu]{teker2025color}
Nurb{\"u}ke Teker, Rui Xiao, Zeynep Akata, and Shuchen Wu.
\newblock What is the color of red? vision--language models prefer to read
  rather than see, 2025.
\newblock Submitted to ICLR 2026 / OpenReview.

\bibitem[Wang et~al.(2024{\natexlab{a}})Wang, Zhou, Huang, Xu, Zhang, Poon, and
  Chen]{wang2024mdpo}
Fei Wang, Wenxuan Zhou, James~Y Huang, Nan Xu, Sheng Zhang, Hoifung Poon, and
  Muhao Chen.
\newblock mdpo: Conditional preference optimization for multimodal large
  language models.
\newblock pages 8078--8088, 2024{\natexlab{a}}.

\bibitem[Wang et~al.(2024{\natexlab{b}})Wang, Bai, Tan, Wang, Fan, Bai, Chen,
  Liu, Wang, Ge, et~al.]{Qwen2VL}
Peng Wang, Shuai Bai, Sinan Tan, Shijie Wang, Zhihao Fan, Jinze Bai, Keqin
  Chen, Xuejing Liu, Jialin Wang, Wenbin Ge, et~al.
\newblock Qwen2-vl: Enhancing vision-language model's perception of the world
  at any resolution.
\newblock \emph{arXiv preprint arXiv:2409.12191}, 2024{\natexlab{b}}.

\bibitem[Wang et~al.(2025)Wang, Gao, Gu, Pu, Cui, Wei, Liu, Jing, Ye, Shao,
  et~al.]{internvl3_5}
Weiyun Wang, Zhangwei Gao, Lixin Gu, Hengjun Pu, Long Cui, Xingguang Wei,
  Zhaoyang Liu, Linglin Jing, Shenglong Ye, Jie Shao, et~al.
\newblock Internvl3.5: Advancing open-source multimodal models in versatility,
  reasoning, and efficiency.
\newblock \emph{arXiv preprint arXiv:2508.18265}, 2025.

\bibitem[Wu et~al.(2024)Wu, Chen, Pan, Liu, Liu, Dai, Gao, Ma, Wu, Wang,
  et~al.]{deepseekvl2}
Zhiyu Wu, Xiaokang Chen, Zizheng Pan, Xingchao Liu, Wen Liu, Damai Dai, Huazuo
  Gao, Yiyang Ma, Chengyue Wu, Bingxuan Wang, et~al.
\newblock Deepseek-vl2: Mixture-of-experts vision-language models for advanced
  multimodal understanding, 2024.

\bibitem[Yan et~al.(2025)Yan, Fan, Li, Jiang, Zhao, Guan, Kuo, and
  Wang]{yan2025multimodal}
Qianqi Yan, Yue Fan, Hongquan Li, Shan Jiang, Yang Zhao, Xinze Guan, Ching-Chen
  Kuo, and Xin~Eric Wang.
\newblock Multimodal inconsistency reasoning (mmir): A new benchmark for
  multimodal reasoning models.
\newblock pages 18829--18845, 2025.

\bibitem[Yang et~al.(2025)Yang, Li, Yang, Zhang, Hui, Zheng, Yu, Gao, Huang,
  Lv, et~al.]{qwen3}
An~Yang, Anfeng Li, Baosong Yang, Beichen Zhang, Binyuan Hui, Bo~Zheng, Bowen
  Yu, Chang Gao, Chengen Huang, Chenxu Lv, et~al.
\newblock Qwen3 technical report.
\newblock \emph{arXiv preprint arXiv:2505.09388}, 2025.

\bibitem[Ye et~al.(2025)Ye, Chen, Liu, Zhang, Li, Ni, and
  Chen]{ye2025assessing}
Hongfei Ye, Bin Chen, Wenxi Liu, Yu~Zhang, Zhao Li, Dandan Ni, and Hongyang
  Chen.
\newblock Assessing color vision test in large vision-language models.
\newblock \emph{arXiv preprint arXiv:2507.11153}, 2025.

\bibitem[Zhang et~al.(2025{\natexlab{a}})Zhang, Ma, Hou, Bai, Chen, Xiang, Yu,
  and Zhang]{zhang2025evaluating}
Yu~Zhang, Jinlong Ma, Yongshuai Hou, Xuefeng Bai, Kehai Chen, Yang Xiang, Jun
  Yu, and Min Zhang.
\newblock Evaluating and steering modality preferences in multimodal large
  language model.
\newblock \emph{arXiv preprint arXiv:2505.20977}, 2025{\natexlab{a}}.

\bibitem[Zhang et~al.(2025{\natexlab{b}})Zhang, Wang, Gong, Shi, Wang, Wang,
  and Hu]{zhang2025modalities}
Zhuoran Zhang, Tengyue Wang, Xilin Gong, Yang Shi, Haotian Wang, Di~Wang, and
  Lijie Hu.
\newblock When modalities conflict: How unimodal reasoning uncertainty governs
  preference dynamics in mllms.
\newblock \emph{arXiv preprint arXiv:2511.02243}, 2025{\natexlab{b}}.

\bibitem[Zhu et~al.(2024)Zhu, Liu, Wang, Tu, and Chen]{zhu2024unraveling}
Tinghui Zhu, Qin Liu, Fei Wang, Zhengzhong Tu, and Muhao Chen.
\newblock Unraveling cross-modality knowledge conflicts in large
  vision-language models.
\newblock \emph{arXiv preprint arXiv:2410.03659}, 2024.

\end{thebibliography}

\newpage
\appendix

\section{Color Space Filtering: Details}
\label{appendix:color_filtering}

Raw color names extracted from the repository cannot be directly permuted into Stroop pairs due to three primary limitations:
\begin{itemize}
 \item \textbf{Background and Text Interference:} Certain extracted colors lack sufficient contrast against standard white backgrounds or black text (e.g., \textit{Licorice} RGB$[27, 18, 18]$ and \textit{Ivory} RGB$[255, 255, 240]$), impairing human legibility.
 \item \textbf{Perceptual Indistinguishability:} Specific color pairs exhibit negligible perceptual differences (e.g., \textit{Carmine} RGB$[215, 0, 64]$ vs. \textit{Cherry} RGB$[210, 4, 45]$).
 \item \textbf{Lexical Ambiguity:} Many color names employ compound nomenclature containing basic color terms (e.g., \textit{"Cadmium Red"}), introducing lexical ambiguity.
\end{itemize}

We designed a rigorous filtering pipeline based on the $\Delta E$ metric. Initially, we filtered out compound color names to ensure lexical atomicity. We then calculated the $\Delta E$ distance between each candidate color and standard white, sorted these distances, and discarded colors falling below the median. We repeated this process relative to standard black, pruning colors below the median distance. Following this filtering process, the remaining palette consists of 59 colors that maintain high contrast against both background and text.

We generated all possible permutations of the surviving colors and computed the pairwise $\Delta E$ for each combination. Based on the distribution, we stratified samples into difficulty levels: 200 ``hard'' pairs from the lower quartile to the median range (perceptually similar yet distinguishable) and 200 ``easy'' pairs from the median to the upper quartile.

\section{CIEDE2000 Color Difference Metric}
\label{appendix:ciede2000}

The CIEDE2000 metric ($\Delta E_{00}$) quantifies perceptual color difference in the CIE $L^*a^*b^*$ space. Input sRGB values are first linearized via inverse gamma correction and transformed into CIE $L^*a^*b^*$:

\begin{equation}
C_{\text{lin}} =
\begin{cases}
\frac{C_{\text{sRGB}}}{12.92}, & \text{if } C_{\text{sRGB}} \le 0.04045 \\
\left( \frac{C_{\text{sRGB}} + 0.055}{1.055} \right)^{2.4}, & \text{otherwise}
\end{cases}
\end{equation}

The perceptual difference is then computed as:

\begin{equation}
\Delta E_{00} = \sqrt{ \left( \frac{\Delta L'}{k_L S_L} \right)^{\!\!2}+\left( \frac{\Delta C'}{k_C S_C} \right)^{\!\!2} + \left( \frac{\Delta H'}{k_H S_H} \right)^{\!\!2} + R_T \Phi }
\label{eq:ciede2000_app}
\end{equation}

where $\Delta L', \Delta C', \Delta H'$ represent differences in lightness, chroma, and hue, respectively. The weighting functions ($S_L, S_C, S_H$) and the rotation term ($R_T \Phi$) correct for non-uniformities in human color discrimination. We set $k_L=k_C=k_H=1$ following standard practice.

\section{Additional Experimental Results and Details}
\label{appendix:additional_experiments}

\subsection{Supplementary Information for Main Experiments}
\label{app:main_supp}

\subsubsection{Permutation Test Details}
\label{app:permutation}
The 64 conditions comprise all model$\times$difficulty pairs in the Embedded setting (16 proprietary + open-source models, including Gemini-3-Flash-Thinking, yielding 16 models $\times$ 2 difficulties = 32 Embedded conditions) and their Standard Stroop counterparts (32 conditions), for 64 total. For each condition, SIR is computed on the observed data. The null distribution is generated by shuffling the semantic color labels across all 200 samples within each condition (preserving model outputs and visual targets), yielding a permuted SIR under the null hypothesis that semantic labels have no systematic relationship with model errors. We perform 5{,}000 permutations per condition. P-values are two-tailed and corrected using Benjamini--Hochberg FDR ($\alpha = 0.05$). Tests are paired by image: each sample appears in both Standard and Embedded conditions for the same model and difficulty.

\subsubsection{Extended Color and Word Settings}
After the color filtering process, we obtain a total of 59 distinct colors along with their corresponding RGB values, as detailed in Table~\ref{tab:appendix-colors}. These colors are utilized to construct the ``Color'' and ``Simple'' subsets of our benchmark to evaluate the grounding and reasoning capabilities of multimodal models. Additionally, Table~\ref{tab:appendix-words} lists the irrelevant words used for text substitution in our control experiments.

\begin{table*}[ht]
\caption{List of colors and their corresponding RGB values used in the experiments.}
\label{tab:appendix-colors}
\vskip 0.15in
\begin{center}
\begin{small}
\begin{sc}
\begin{tabular}{ll ll ll}
\toprule
Color & RGB Value & Color & RGB Value & Color & RGB Value \\
\midrule
Amber & (255, 191, 0) & Glaucous & (96, 130, 182) & Pewter & (137, 148, 153) \\
Brass & (225, 193, 110) & Gold & (255, 215, 0) & Pistachio & (147, 197, 114) \\
Bronze & (205, 127, 50) & Goldenrod & (218, 165, 32) & Poppy & (227, 83, 53) \\
Buff & (218, 160, 109) & Green & (0, 128, 0) & Puce & (169, 92, 104) \\
Butterscotch & (227, 150, 62) & Icterine & (252, 245, 95) & Raspberry & (227, 11, 92) \\
Camel & (193, 154, 107) & Indigo & (63, 0, 255) & Red & (255, 0, 0) \\
Carmine & (215, 0, 64) & Jade & (0, 163, 108) & Rose & (243, 58, 106) \\
Cerise & (222, 49, 99) & Lilac & (170, 152, 169) & Saffron & (244, 196, 48) \\
Chartreuse & (223, 255, 0) & Mahogany & (192, 64, 0) & Salmon & (250, 128, 114) \\
Cherry & (210, 4, 45) & Maize & (251, 236, 93) & Scarlet & (255, 36, 0) \\
Cinnamon & (210, 125, 45) & Malachite & (11, 218, 81) & Smoke & (132, 136, 132) \\
Citrine & (228, 208, 10) & Mango & (244, 187, 68) & Tangerine & (240, 128, 0) \\
Copper & (184, 115, 51) & Mauve & (224, 176, 255) & Teal & (0, 128, 128) \\
Coral & (255, 127, 80) & Mocha & (150, 121, 105) & Turquoise & (64, 224, 208) \\
Crimson & (220, 20, 60) & Ochre & (204, 119, 34) & Verdigris & (64, 181, 173) \\
Denim & (111, 143, 175) & Orange & (255, 165, 0) & Vermillion & (227, 66, 52) \\
Eucalyptus & (95, 133, 117) & Orchid & (218, 112, 214) & Violet & (127, 0, 255) \\
Fawn & (229, 170, 112) & Pear & (201, 204, 63) & Viridian & (64, 130, 109) \\
Fuchsia & (255, 0, 255) & Peridot & (180, 196, 36) & Yellow & (255, 255, 0) \\
Gamboge & (228, 155, 15) & Persimmon & (236, 88, 0) & & \\
\bottomrule
\end{tabular}
\end{sc}
\end{small}
\end{center}
\vskip -0.1in
\end{table*}

\begin{table*}[ht]
\caption{List of words used as irrelevant text content in the experiments.}
\label{tab:appendix-words}
\vskip 0.15in
\begin{center}
\begin{small}
\begin{sc}
\begin{tabular}{lllll}
\toprule
\multicolumn{5}{c}{Irrelevant Words Selection} \\
\midrule
beat & exit & lamp & part & speed \\
bill & factory & leave & past & staff \\
box & family & length & path & stay \\
break & farm & lock & piece & street \\
breath & fix & luck & pipe & swim \\
bus & floor & market & plane & tax \\
car & force & member & power & team \\
cash & friend & mind & pull & town \\
catch & gear & minute & pump & train \\
century & glass & moment & risk & village \\
city & grow & money & road & voice \\
clock & guest & month & roof & wake \\
close & habit & more & rule & watch \\
cost & hall & music & run & way \\
crowd & hope & near & school & week \\
cup & hotel & neighbor & screw & wheel \\
cut & hour & noise & second & window \\
day & house & office & ship & wire \\
engine & job & open & side & work \\
enter & key & park & sleep & yard \\
\bottomrule
\end{tabular}
\end{sc}
\end{small}
\end{center}
\vskip -0.1in
\end{table*}

\section{Full Results Tables}
\label{appendix:full_tables}

The complete quantitative evaluation results across all templates, difficulties, and perturbation variants are provided below.

\label{appendix:prop_full}

\begin{table*}[!ht]
 \caption{\textbf{Quantitative Evaluation on the WCIT Benchmark with Proprietary Models.} Performance comparison between \textit{Embedded Stroop (Ours)} and \textit{Standard Stroop} settings. \textit{Control Group} serves as the hallucination-free baseline. OER = 100\% $-$ Acc $-$ SHR (see \S\ref{sec:metrics}).}
 \label{tab:main_prop_full}
 \centering
 \tiny
 \setlength{\tabcolsep}{1.5pt}
 \resizebox{\textwidth}{!}{%
 \begin{tabular}{lllcccccccccccccccccc}
 \toprule
 \multirow{3}{*}{\textbf{Model}} & \multirow{3}{*}{\textbf{Prompt}} & \multirow{3}{*}{\textbf{Diff}} & \multicolumn{4}{c}{\textbf{Stroop}} & \multicolumn{2}{c}{\textbf{Control}} & \multicolumn{6}{c}{\textbf{Flip}} & \multicolumn{6}{c}{\textbf{Mask (25\%)}} \\
 \cmidrule(lr){4-7} \cmidrule(lr){8-9} \cmidrule(lr){10-15} \cmidrule(lr){16-21}
 & & & \multicolumn{2}{c}{\textbf{Embed(Ours)}} & \multicolumn{2}{c}{\textbf{Standard}} & \multicolumn{2}{c}{\textbf{-}} & \multicolumn{2}{c}{\textbf{LR}} & \multicolumn{2}{c}{\textbf{UD}} & \multicolumn{2}{c}{\textbf{Rev}} & \multicolumn{2}{c}{\textbf{Square}} & \multicolumn{2}{c}{\textbf{Star}} & \multicolumn{2}{c}{\textbf{Line}} \\
 \cmidrule(lr){4-5} \cmidrule(lr){6-7} \cmidrule(lr){8-9} \cmidrule(lr){10-11} \cmidrule(lr){12-13} \cmidrule(lr){14-15} \cmidrule(lr){16-17} \cmidrule(lr){18-19} \cmidrule(lr){20-21}
 & & & Acc\% & SHR\% & Acc\% & SHR\% & Acc\% & SHR\% & Acc\% & SHR\% & Acc\% & SHR\% & Acc\% & SHR\% & Acc\% & SHR\% & Acc\% & SHR\% & Acc\% & SHR\% \\
 \midrule

 \multirow{6}{*}{\textbf{GPT-5.2}} & \multirow{2}{*}{Template 1} & Easy & 11.0 & \cellcolor{gray!10}4.5 & 12.5 & \cellcolor{gray!10}0.0 & 12.0 & \cellcolor{gray!10}- & 12.5 & \cellcolor{gray!10}0.0 & 12.0 & \cellcolor{gray!10}0.0 & 12.0 & \cellcolor{gray!10}0.0 & 10.5 & \cellcolor{gray!10}2.5 & 12.0 & \cellcolor{gray!10}1.0 & 11.0 & \cellcolor{gray!10}1.5 \\ 
 & & Hard & 7.0 & \cellcolor{gray!10}25.0 & 7.5 & \cellcolor{gray!10}13.5 & 7.5 & \cellcolor{gray!10}- & 7.0 & \cellcolor{gray!10}2.5 & 7.5 & \cellcolor{gray!10}2.0 & 7.5 & \cellcolor{gray!10}2.5 & 7.5 & \cellcolor{gray!10}10.5 & 7.5 & \cellcolor{gray!10}7.0 & 8.0 & \cellcolor{gray!10}5.0 \\ \cmidrule(lr){2-21}
 & \multirow{2}{*}{Template 2} & Easy & 11.5 & \cellcolor{gray!10}4.5 & 12.0 & \cellcolor{gray!10}0.0 & 12.5 & \cellcolor{gray!10}- & 11.5 & \cellcolor{gray!10}0.0 & 11.5 & \cellcolor{gray!10}0.0 & 12.5 & \cellcolor{gray!10}0.0 & 11.5 & \cellcolor{gray!10}4.0 & 12.0 & \cellcolor{gray!10}2.0 & 11.0 & \cellcolor{gray!10}0.5 \\ 
 & & Hard & 7.0 & \cellcolor{gray!10}27.5 & 7.5 & \cellcolor{gray!10}14.0 & 7.5 & \cellcolor{gray!10}- & 8.0 & \cellcolor{gray!10}3.0 & 8.0 & \cellcolor{gray!10}2.0 & 7.5 & \cellcolor{gray!10}2.5 & 7.5 & \cellcolor{gray!10}9.0 & 7.5 & \cellcolor{gray!10}6.5 & 7.5 & \cellcolor{gray!10}4.5 \\ \cmidrule(lr){2-21}
 & \multirow{2}{*}{Template 3} & Easy & 20.5 & \cellcolor{gray!10}4.5 & 17.5 & \cellcolor{gray!10}1.0 & 22.0 & \cellcolor{gray!10}- & 20.5 & \cellcolor{gray!10}0.0 & 21.5 & \cellcolor{gray!10}0.0 & 20.5 & \cellcolor{gray!10}0.0 & 20.0 & \cellcolor{gray!10}1.5 & 20.5 & \cellcolor{gray!10}1.5 & 19.5 & \cellcolor{gray!10}0.5 \\ 
 & & Hard & 6.5 & \cellcolor{gray!10}25.0 & 7.5 & \cellcolor{gray!10}10.5 & 8.0 & \cellcolor{gray!10}- & 8.0 & \cellcolor{gray!10}2.0 & 7.5 & \cellcolor{gray!10}2.0 & 8.0 & \cellcolor{gray!10}2.5 & 8.0 & \cellcolor{gray!10}10.0 & 7.5 & \cellcolor{gray!10}5.5 & 7.5 & \cellcolor{gray!10}5.0 \\ \midrule

 \multirow{6}{*}{\textbf{GPT-4o}} & \multirow{2}{*}{Template 1} & Easy & 10.0 & \cellcolor{gray!10}8.0 & 10.5 & \cellcolor{gray!10}0.0 & 11.5 & \cellcolor{gray!10}- & 10.0 & \cellcolor{gray!10}10.0 & 9.5 & \cellcolor{gray!10}0.0 & 10.5 & \cellcolor{gray!10}0.5 & 9.5 & \cellcolor{gray!10}4.5 & 9.5 & \cellcolor{gray!10}4.0 & 10.0 & \cellcolor{gray!10}9.5 \\ 
 & & Hard & 7.5 & \cellcolor{gray!10}47.0 & 7.5 & \cellcolor{gray!10}12.5 & 6.5 & \cellcolor{gray!10}- & 7.0 & \cellcolor{gray!10}28.0 & 7.5 & \cellcolor{gray!10}4.0 & 7.5 & \cellcolor{gray!10}2.0 & 7.0 & \cellcolor{gray!10}17.5 & 7.0 & \cellcolor{gray!10}24.5 & 7.0 & \cellcolor{gray!10}22.0 \\ \cmidrule(lr){2-21}
 & \multirow{2}{*}{Template 2} & Easy & 8.0 & \cellcolor{gray!10}10.0 & 11.0 & \cellcolor{gray!10}0.5 & 11.5 & \cellcolor{gray!10}- & 10.5 & \cellcolor{gray!10}9.5 & 10.5 & \cellcolor{gray!10}0.0 & 11.0 & \cellcolor{gray!10}0.5 & 9.5 & \cellcolor{gray!10}8.0 & 9.5 & \cellcolor{gray!10}6.0 & 9.5 & \cellcolor{gray!10}10.5 \\ 
 & & Hard & 6.0 & \cellcolor{gray!10}52.0 & 7.5 & \cellcolor{gray!10}12.0 & 7.5 & \cellcolor{gray!10}- & 6.0 & \cellcolor{gray!10}26.0 & 7.5 & \cellcolor{gray!10}2.5 & 8.0 & \cellcolor{gray!10}3.0 & 6.5 & \cellcolor{gray!10}26.0 & 7.0 & \cellcolor{gray!10}24.5 & 7.5 & \cellcolor{gray!10}25.5 \\ \cmidrule(lr){2-21}
 & \multirow{2}{*}{Template 3} & Easy & 16.0 & \cellcolor{gray!10}7.5 & 16.0 & \cellcolor{gray!10}0.5 & 17.5 & \cellcolor{gray!10}- & 16.5 & \cellcolor{gray!10}8.0 & 18.0 & \cellcolor{gray!10}0.0 & 18.0 & \cellcolor{gray!10}0.5 & 16.5 & \cellcolor{gray!10}5.5 & 17.0 & \cellcolor{gray!10}7.5 & 15.5 & \cellcolor{gray!10}7.5 \\ 
 & & Hard & 7.0 & \cellcolor{gray!10}52.5 & 7.5 & \cellcolor{gray!10}9.5 & 7.5 & \cellcolor{gray!10}- & 7.5 & \cellcolor{gray!10}28.5 & 7.0 & \cellcolor{gray!10}4.0 & 7.5 & \cellcolor{gray!10}3.0 & 5.5 & \cellcolor{gray!10}22.0 & 8.0 & \cellcolor{gray!10}25.5 & 7.0 & \cellcolor{gray!10}26.0 \\ \midrule

 \multirow{6}{*}{\textbf{Gemini-3-flash-thinking}} & \multirow{2}{*}{Template 1} & Easy & 12.0 & \cellcolor{gray!10}0.5 & 16.5 & \cellcolor{gray!10}0.0 & 12.5 & \cellcolor{gray!10}- & 12.5 & \cellcolor{gray!10}1.0 & 14.5 & \cellcolor{gray!10}0.0 & 12.5 & \cellcolor{gray!10}1.0 & 12.0 & \cellcolor{gray!10}0.0 & 14.5 & \cellcolor{gray!10}0.0 & 13.0 & \cellcolor{gray!10}0.0 \\ 
 & & Hard & 8.5 & \cellcolor{gray!10}14.0 & 11.5 & \cellcolor{gray!10}8.0 & 9.5 & \cellcolor{gray!10}- & 9.5 & \cellcolor{gray!10}27.5 & 10.0 & \cellcolor{gray!10}17.5 & 8.5 & \cellcolor{gray!10}23.0 & 10.5 & \cellcolor{gray!10}15.5 & 10.5 & \cellcolor{gray!10}12.0 & 8.0 & \cellcolor{gray!10}16.0 \\ \cmidrule(lr){2-21}
 & \multirow{2}{*}{Template 2} & Easy & 14.0 & \cellcolor{gray!10}0.0 & 14.5 & \cellcolor{gray!10}0.0 & 10.0 & \cellcolor{gray!10}- & 14.0 & \cellcolor{gray!10}0.5 & 15.5 & \cellcolor{gray!10}0.0 & 11.0 & \cellcolor{gray!10}2.5 & 13.5 & \cellcolor{gray!10}0.0 & 13.5 & \cellcolor{gray!10}0.0 & 14.0 & \cellcolor{gray!10}0.0 \\ 
 & & Hard & 9.0 & \cellcolor{gray!10}15.0 & 11.0 & \cellcolor{gray!10}9.5 & 9.5 & \cellcolor{gray!10}- & 9.5 & \cellcolor{gray!10}30.5 & 8.0 & \cellcolor{gray!10}18.0 & 9.0 & \cellcolor{gray!10}17.5 & 8.0 & \cellcolor{gray!10}13.5 & 7.5 & \cellcolor{gray!10}10.5 & 8.0 & \cellcolor{gray!10}15.0 \\ \cmidrule(lr){2-21}
 & \multirow{2}{*}{Template 3} & Easy & 20.5 & \cellcolor{gray!10}0.0 & 23.5 & \cellcolor{gray!10}0.0 & 21.0 & \cellcolor{gray!10}- & 21.0 & \cellcolor{gray!10}0.5 & 21.0 & \cellcolor{gray!10}0.5 & 17.5 & \cellcolor{gray!10}2.0 & 23.5 & \cellcolor{gray!10}0.0 & 23.0 & \cellcolor{gray!10}0.0 & 20.5 & \cellcolor{gray!10}0.0 \\ 
 & & Hard & 10.0 & \cellcolor{gray!10}15.0 & 11.5 & \cellcolor{gray!10}8.5 & 10.0 & \cellcolor{gray!10}- & 9.5 & \cellcolor{gray!10}28.5 & 9.5 & \cellcolor{gray!10}16.5 & 9.0 & \cellcolor{gray!10}19.0 & 9.5 & \cellcolor{gray!10}16.0 & 11.5 & \cellcolor{gray!10}12.5 & 8.5 & \cellcolor{gray!10}15.5 \\ \midrule

 \multirow{6}{*}{\textbf{Gemini-2.5-flash}} & \multirow{2}{*}{Template 1} & Easy & 11.5 & \cellcolor{gray!10}1.0 & 14.5 & \cellcolor{gray!10}0.0 & 12.0 & \cellcolor{gray!10}- & 9.5 & \cellcolor{gray!10}0.5 & 12.5 & \cellcolor{gray!10}0.0 & 12.5 & \cellcolor{gray!10}1.0 & 12.0 & \cellcolor{gray!10}0.5 & 14.5 & \cellcolor{gray!10}0.0 & 11.0 & \cellcolor{gray!10}0.0 \\ 
 & & Hard & 10.5 & \cellcolor{gray!10}19.0 & 8.0 & \cellcolor{gray!10}9.0 & 9.5 & \cellcolor{gray!10}- & 7.5 & \cellcolor{gray!10}14.5 & 8.0 & \cellcolor{gray!10}3.0 & 7.5 & \cellcolor{gray!10}6.0 & 6.0 & \cellcolor{gray!10}9.5 & 8.0 & \cellcolor{gray!10}8.0 & 7.5 & \cellcolor{gray!10}14.0 \\ \cmidrule(lr){2-21}
 & \multirow{2}{*}{Template 2} & Easy & 13.0 & \cellcolor{gray!10}0.5 & 12.5 & \cellcolor{gray!10}0.0 & 12.5 & \cellcolor{gray!10}- & 11.0 & \cellcolor{gray!10}1.5 & 12.5 & \cellcolor{gray!10}0.0 & 13.0 & \cellcolor{gray!10}0.0 & 13.0 & \cellcolor{gray!10}0.0 & 11.5 & \cellcolor{gray!10}0.0 & 12.0 & \cellcolor{gray!10}0.5 \\ 
 & & Hard & 9.0 & \cellcolor{gray!10}19.0 & 9.5 & \cellcolor{gray!10}7.0 & 9.0 & \cellcolor{gray!10}- & 7.5 & \cellcolor{gray!10}11.5 & 8.5 & \cellcolor{gray!10}3.5 & 8.0 & \cellcolor{gray!10}9.0 & 8.0 & \cellcolor{gray!10}7.0 & 9.5 & \cellcolor{gray!10}10.0 & 9.5 & \cellcolor{gray!10}11.5 \\ \cmidrule(lr){2-21}
 & \multirow{2}{*}{Template 3} & Easy & 18.5 & \cellcolor{gray!10}0.0 & 23.5 & \cellcolor{gray!10}0.0 & 19.5 & \cellcolor{gray!10}- & 16.0 & \cellcolor{gray!10}0.0 & 17.5 & \cellcolor{gray!10}0.0 & 15.0 & \cellcolor{gray!10}0.5 & 16.5 & \cellcolor{gray!10}0.5 & 17.5 & \cellcolor{gray!10}0.0 & 16.5 & \cellcolor{gray!10}0.5 \\ 
 & & Hard & 9.0 & \cellcolor{gray!10}24.0 & 8.0 & \cellcolor{gray!10}8.5 & 8.5 & \cellcolor{gray!10}- & 8.5 & \cellcolor{gray!10}14.5 & 8.5 & \cellcolor{gray!10}3.0 & 9.0 & \cellcolor{gray!10}7.0 & 6.5 & \cellcolor{gray!10}9.0 & 9.0 & \cellcolor{gray!10}11.0 & 7.5 & \cellcolor{gray!10}11.0 \\ \midrule

 \multirow{6}{*}{\textbf{Qwen3-vl-plus}} & \multirow{2}{*}{Template 1} & Easy & 8.0 & \cellcolor{gray!10}47.0 & 10.0 & \cellcolor{gray!10}16.5 & 10.5 & \cellcolor{gray!10}- & 10.5 & \cellcolor{gray!10}0.5 & 10.5 & \cellcolor{gray!10}1.0 & 10.5 & \cellcolor{gray!10}0.0 & 10.0 & \cellcolor{gray!10}4.5 & 10.0 & \cellcolor{gray!10}1.0 & 10.0 & \cellcolor{gray!10}3.0 \\ 
 & & Hard & 5.5 & \cellcolor{gray!10}53.0 & 7.0 & \cellcolor{gray!10}34.5 & 7.5 & \cellcolor{gray!10}- & 7.5 & \cellcolor{gray!10}4.5 & 7.5 & \cellcolor{gray!10}4.0 & 7.5 & \cellcolor{gray!10}2.0 & 6.0 & \cellcolor{gray!10}5.5 & 8.0 & \cellcolor{gray!10}4.5 & 7.0 & \cellcolor{gray!10}3.0 \\ \cmidrule(lr){2-21}
 & \multirow{2}{*}{Template 2} & Easy & 6.5 & \cellcolor{gray!10}51.0 & 10.0 & \cellcolor{gray!10}20.0 & 10.5 & \cellcolor{gray!10}- & 10.0 & \cellcolor{gray!10}0.5 & 10.5 & \cellcolor{gray!10}0.5 & 10.5 & \cellcolor{gray!10}0.0 & 9.0 & \cellcolor{gray!10}4.5 & 10.0 & \cellcolor{gray!10}1.5 & 10.0 & \cellcolor{gray!10}1.0 \\ 
 & & Hard & 5.0 & \cellcolor{gray!10}55.5 & 7.0 & \cellcolor{gray!10}33.0 & 7.5 & \cellcolor{gray!10}- & 7.5 & \cellcolor{gray!10}5.5 & 7.5 & \cellcolor{gray!10}4.0 & 7.5 & \cellcolor{gray!10}1.5 & 6.0 & \cellcolor{gray!10}4.5 & 7.5 & \cellcolor{gray!10}4.5 & 7.0 & \cellcolor{gray!10}3.5 \\ \cmidrule(lr){2-21}
 & \multirow{2}{*}{Template 3} & Easy & 7.5 & \cellcolor{gray!10}51.5 & 13.5 & \cellcolor{gray!10}18.0 & 13.5 & \cellcolor{gray!10}- & 12.5 & \cellcolor{gray!10}0.5 & 13.0 & \cellcolor{gray!10}0.5 & 12.5 & \cellcolor{gray!10}0.0 & 11.5 & \cellcolor{gray!10}5.5 & 11.5 & \cellcolor{gray!10}2.0 & 12.0 & \cellcolor{gray!10}1.5 \\ 
 & & Hard & 5.0 & \cellcolor{gray!10}56.5 & 7.0 & \cellcolor{gray!10}33.5 & 7.5 & \cellcolor{gray!10}- & 7.5 & \cellcolor{gray!10}5.5 & 7.5 & \cellcolor{gray!10}4.0 & 7.5 & \cellcolor{gray!10}1.5 & 6.0 & \cellcolor{gray!10}4.5 & 7.5 & \cellcolor{gray!10}5.0 & 7.0 & \cellcolor{gray!10}3.5 \\ \midrule
 
 \multirow{6}{*}{\textbf{GLM-4.5v}} & \multirow{2}{*}{Template 1} & Easy & 1.5 & \cellcolor{gray!10}4.0 & 2.0 & \cellcolor{gray!10}0.5 & 1.5 & \cellcolor{gray!10}- & 2.5 & \cellcolor{gray!10}1.0 & 1.5 & \cellcolor{gray!10}1.0 & 4.5 & \cellcolor{gray!10}0.0 & 2.0 & \cellcolor{gray!10}0.0 & 1.5 & \cellcolor{gray!10}0.5 & 3.0 & \cellcolor{gray!10}0.0 \\ 
 & & Hard & 1.5 & \cellcolor{gray!10}13.0 & 1.5 & \cellcolor{gray!10}5.5 & 2.5 & \cellcolor{gray!10}- & 2.5 & \cellcolor{gray!10}3.5 & 1.0 & \cellcolor{gray!10}1.0 & 1.5 & \cellcolor{gray!10}1.0 & 2.0 & \cellcolor{gray!10}1.0 & 2.5 & \cellcolor{gray!10}0.5 & 2.5 & \cellcolor{gray!10}0.5 \\ \cmidrule(lr){2-21}
 & \multirow{2}{*}{Template 2} & Easy & 2.0 & \cellcolor{gray!10}2.5 & 1.0 & \cellcolor{gray!10}0.0 & 4.5 & \cellcolor{gray!10}- & 2.5 & \cellcolor{gray!10}1.0 & 1.0 & \cellcolor{gray!10}0.5 & 3.5 & \cellcolor{gray!10}0.0 & 4.5 & \cellcolor{gray!10}0.5 & 2.5 & \cellcolor{gray!10}0.5 & 2.5 & \cellcolor{gray!10}0.0 \\ 
 & & Hard & 1.0 & \cellcolor{gray!10}12.0 & 1.5 & \cellcolor{gray!10}3.0 & 1.5 & \cellcolor{gray!10}- & 3.0 & \cellcolor{gray!10}4.5 & 2.0 & \cellcolor{gray!10}1.0 & 2.0 & \cellcolor{gray!10}1.0 & 2.0 & \cellcolor{gray!10}3.0 & 2.0 & \cellcolor{gray!10}0.5 & 3.0 & \cellcolor{gray!10}1.5 \\ \cmidrule(lr){2-21}
 & \multirow{2}{*}{Template 3} & Easy & 2.0 & \cellcolor{gray!10}3.0 & 1.0 & \cellcolor{gray!10}0.0 & 3.5 & \cellcolor{gray!10}- & 5.5 & \cellcolor{gray!10}1.0 & 2.0 & \cellcolor{gray!10}0.5 & 2.0 & \cellcolor{gray!10}0.0 & 6.0 & \cellcolor{gray!10}0.0 & 2.5 & \cellcolor{gray!10}0.0 & 4.5 & \cellcolor{gray!10}0.0 \\ 
 & & Hard & 1.5 & \cellcolor{gray!10}10.0 & 1.0 & \cellcolor{gray!10}5.0 & 1.5 & \cellcolor{gray!10}- & 3.5 & \cellcolor{gray!10}5.0 & 1.0 & \cellcolor{gray!10}1.5 & 3.5 & \cellcolor{gray!10}0.5 & 3.5 & \cellcolor{gray!10}2.0 & 2.5 & \cellcolor{gray!10}2.0 & 2.0 & \cellcolor{gray!10}0.5 \\ 
 \bottomrule

 \end{tabular}%
 }%
\end{table*}
\label{appendix:open_full}

\begin{table*}[!ht]
 \caption{\textbf{Quantitative Evaluation on the WCIT Benchmark with Open Source Models.} Performance comparison between \textit{Embedded Stroop (Ours)} and \textit{Standard Stroop} settings. \textit{Control Group} serves as the hallucination-free baseline. OER = 100\% $-$ Acc $-$ SHR (see \S\ref{sec:metrics}).}
 \label{tab:main_open_full}
 \centering
 \tiny
 \setlength{\tabcolsep}{1.5pt}
 \resizebox{\textwidth}{!}{%
 \begin{tabular}{lllcccccccccccccccccc}
 \toprule
 \multirow{3}{*}{\textbf{Model}} & \multirow{3}{*}{\textbf{Prompt}} & \multirow{3}{*}{\textbf{Diff}} & \multicolumn{4}{c}{\textbf{Stroop}} & \multicolumn{2}{c}{\textbf{Control}} & \multicolumn{6}{c}{\textbf{Flip}} & \multicolumn{6}{c}{\textbf{Mask (25\%)}} \\
 \cmidrule(lr){4-7} \cmidrule(lr){8-9} \cmidrule(lr){10-15} \cmidrule(lr){16-21}
 & & & \multicolumn{2}{c}{\textbf{Embed(Ours)}} & \multicolumn{2}{c}{\textbf{Standard}} & \multicolumn{2}{c} {\textbf{-}} & \multicolumn{2}{c}{\textbf{LR}} & \multicolumn{2}{c}{\textbf{UD}} & \multicolumn{2}{c}{\textbf{Rev}} & \multicolumn{2}{c}{\textbf{Square}} & \multicolumn{2}{c}{\textbf{Star}} & \multicolumn{2}{c}{\textbf{Line}} \\
 \cmidrule(lr){4-5} \cmidrule(lr){6-7} \cmidrule(lr){8-9} \cmidrule(lr){10-11} \cmidrule(lr){12-13} \cmidrule(lr){14-15} \cmidrule(lr){16-17} \cmidrule(lr){18-19} \cmidrule(lr){20-21}
 & & & Acc\% & SHR\% & Acc\% & SHR\% & Acc\% & SHR\% & Acc\% & SHR\% & Acc\% & SHR\% & Acc\% & SHR\% & Acc\% & SHR\% & Acc\% & SHR\% & Acc\% & SHR\% \\
 \midrule

 \multirow{6}{*}{\textbf{Deepseek-vl2}} & \multirow{2}{*}{Template 1} & Easy & 2.5 & \cellcolor{gray!10}70.5 & 9.0 & \cellcolor{gray!10}14.5 & 10.5 & \cellcolor{gray!10}- & 11.0 & \cellcolor{gray!10}1.5 & 10.5 & \cellcolor{gray!10}0.0 & 10.5 & \cellcolor{gray!10}0.0 & 9.5 & \cellcolor{gray!10}12.5 & 9.0 & \cellcolor{gray!10}7.0 & 9.5 & \cellcolor{gray!10}3.5 \\ 
 ~ & ~ & Hard & 5.5 & \cellcolor{gray!10}62.5 & 7.5 & \cellcolor{gray!10}3.5 & 7.5 & \cellcolor{gray!10}- & 7.5 & \cellcolor{gray!10}3.0 & 7.5 & \cellcolor{gray!10}1.5 & 7.5 & \cellcolor{gray!10}1.5 & 7.5 & \cellcolor{gray!10}6.5 & 7.0 & \cellcolor{gray!10}5.0 & 7.5 & \cellcolor{gray!10}3.0 \\ \cmidrule(lr){2-21}
 ~ & \multirow{2}{*}{Template 2} & Easy & 2.5 & \cellcolor{gray!10}73.0 & 9.5 & \cellcolor{gray!10}15.0 & 10.5 & \cellcolor{gray!10}- & 10.5 & \cellcolor{gray!10}2.0 & 10.5 & \cellcolor{gray!10}0.5 & 10.5 & \cellcolor{gray!10}0.0 & 9.5 & \cellcolor{gray!10}12.0 & 9.0 & \cellcolor{gray!10}7.0 & 9.5 & \cellcolor{gray!10}3.5 \\ 
 ~ & ~ & Hard & 5.5 & \cellcolor{gray!10}62.5 & 7.5 & \cellcolor{gray!10}4.0 & 7.5 & \cellcolor{gray!10}- & 7.5 & \cellcolor{gray!10}3.0 & 7.5 & \cellcolor{gray!10}1.5 & 7.5 & \cellcolor{gray!10}1.5 & 7.5 & \cellcolor{gray!10}7.5 & 7.0 & \cellcolor{gray!10}5.0 & 7.5 & \cellcolor{gray!10}3.0 \\ \cmidrule(lr){2-21}
 ~ & \multirow{2}{*}{Template 3} & Easy & 4.0 & \cellcolor{gray!10}72.0 & 12.0 & \cellcolor{gray!10}15.5 & 13.5 & \cellcolor{gray!10}- & 13.5 & \cellcolor{gray!10}2.0 & 14.0 & \cellcolor{gray!10}0.0 & 13.0 & \cellcolor{gray!10}0.0 & 12.5 & \cellcolor{gray!10}12.0 & 11.5 & \cellcolor{gray!10}6.5 & 12.5 & \cellcolor{gray!10}3.0 \\ 
 ~ & ~ & Hard & 5.5 & \cellcolor{gray!10}62.5 & 7.5 & \cellcolor{gray!10}4.0 & 7.5 & \cellcolor{gray!10}- & 7.5 & \cellcolor{gray!10}3.0 & 7.5 & \cellcolor{gray!10}1.5 & 7.5 & \cellcolor{gray!10}1.5 & 7.5 & \cellcolor{gray!10}7.5 & 7.0 & \cellcolor{gray!10}5.0 & 7.5 & \cellcolor{gray!10}3.0 \\ 
 \midrule
 
 \multirow{6}{*}{\textbf{Deepseek-vl2-small}} & \multirow{2}{*}{Template 1} & Easy & 6.5 & \cellcolor{gray!10}47.5 & 9.0 & \cellcolor{gray!10}1.0 & 10.5 & \cellcolor{gray!10}- & 11.0 & \cellcolor{gray!10}0.0 & 10.5 & \cellcolor{gray!10}0.0 & 10.5 & \cellcolor{gray!10}0.0 & 9.5 & \cellcolor{gray!10}6.5 & 9.5 & \cellcolor{gray!10}4.0 & 10.0 & \cellcolor{gray!10}3.0 \\ 
 ~ & ~ & Hard & 6.5 & \cellcolor{gray!10}41.5 & 7.5 & \cellcolor{gray!10}3.0 & 7.5 & \cellcolor{gray!10}- & 7.5 & \cellcolor{gray!10}2.5 & 7.5 & \cellcolor{gray!10}1.5 & 7.5 & \cellcolor{gray!10}1.5 & 7.0 & \cellcolor{gray!10}6.0 & 7.5 & \cellcolor{gray!10}5.0 & 7.5 & \cellcolor{gray!10}3.0 \\ \cmidrule(lr){2-21}
 ~ & \multirow{2}{*}{Template 2} & Easy & 6.0 & \cellcolor{gray!10}49.5 & 9.0 & \cellcolor{gray!10}1.0 & 10.5 & \cellcolor{gray!10}- & 11.5 & \cellcolor{gray!10}1.0 & 10.5 & \cellcolor{gray!10}0.0 & 10.5 & \cellcolor{gray!10}0.0 & 9.5 & \cellcolor{gray!10}7.5 & 9.5 & \cellcolor{gray!10}4.5 & 10.0 & \cellcolor{gray!10}3.5 \\ 
 ~ & ~ & Hard & 6.5 & \cellcolor{gray!10}40.5 & 7.5 & \cellcolor{gray!10}3.0 & 7.5 & \cellcolor{gray!10}- & 7.5 & \cellcolor{gray!10}3.0 & 7.5 & \cellcolor{gray!10}2.0 & 7.5 & \cellcolor{gray!10}1.5 & 7.0 & \cellcolor{gray!10}6.0 & 7.5 & \cellcolor{gray!10}4.5 & 7.5 & \cellcolor{gray!10}3.0 \\ \cmidrule(lr){2-21}
 ~ & \multirow{2}{*}{Template 3} & Easy & 8.5 & \cellcolor{gray!10}47.5 & 9.5 & \cellcolor{gray!10}1.5 & 15.0 & \cellcolor{gray!10}- & 16.0 & \cellcolor{gray!10}0.5 & 14.5 & \cellcolor{gray!10}0.0 & 15.0 & \cellcolor{gray!10}0.0 & 13.5 & \cellcolor{gray!10}6.5 & 13.5 & \cellcolor{gray!10}4.5 & 14.0 & \cellcolor{gray!10}3.0 \\ 
 ~ & ~ & Hard & 6.5 & \cellcolor{gray!10}40.5 & 7.5 & \cellcolor{gray!10}3.0 & 7.5 & \cellcolor{gray!10}- & 7.5 & \cellcolor{gray!10}3.0 & 7.5 & \cellcolor{gray!10}2.0 & 7.5 & \cellcolor{gray!10}1.5 & 7.0 & \cellcolor{gray!10}6.0 & 7.5 & \cellcolor{gray!10}4.5 & 7.5 & \cellcolor{gray!10}3.0 \\ 
 \midrule
 
 \multirow{6}{*}{\textbf{Kimi-VL-A3B-Instruct}} & \multirow{2}{*}{Template 1} & Easy & 3.5 & \cellcolor{gray!10}60.0 & 6.0 & \cellcolor{gray!10}0.0 & 8.0 & \cellcolor{gray!10}- & 4.5 & \cellcolor{gray!10}3.0 & 6.5 & \cellcolor{gray!10}0.0 & 8.0 & \cellcolor{gray!10}0.0 & 8.0 & \cellcolor{gray!10}7.0 & 5.5 & \cellcolor{gray!10}3.0 & 9.0 & \cellcolor{gray!10}3.5 \\ 
 ~ & ~ & Hard & 3.5 & \cellcolor{gray!10}67.0 & 4.0 & \cellcolor{gray!10}1.0 & 6.0 & \cellcolor{gray!10}- & 3.5 & \cellcolor{gray!10}2.0 & 6.0 & \cellcolor{gray!10}1.5 & 5.0 & \cellcolor{gray!10}1.5 & 7.0 & \cellcolor{gray!10}7.0 & 5.0 & \cellcolor{gray!10}3.5 & 7.0 & \cellcolor{gray!10}1.5 \\ \cmidrule(lr){2-21}
 ~ & \multirow{2}{*}{Template 2} & Easy & 0.5 & \cellcolor{gray!10}15.0 & 1.0 & \cellcolor{gray!10}0.5 & 1.0 & \cellcolor{gray!10}- & 1.0 & \cellcolor{gray!10}0.5 & 1.0 & \cellcolor{gray!10}0.0 & 1.0 & \cellcolor{gray!10}0.0 & 2.0 & \cellcolor{gray!10}5.5 & 0.5 & \cellcolor{gray!10}1.0 & 1.0 & \cellcolor{gray!10}1.0 \\ 
 ~ & ~ & Hard & 0.0 & \cellcolor{gray!10}1.5 & 0.0 & \cellcolor{gray!10}0.0 & 1.0 & \cellcolor{gray!10}- & 0.0 & \cellcolor{gray!10}0.0 & 0.0 & \cellcolor{gray!10}0.0 & 0.5 & \cellcolor{gray!10}0.0 & 0.0 & \cellcolor{gray!10}3.0 & 0.5 & \cellcolor{gray!10}1.0 & 0.5 & \cellcolor{gray!10}1.0 \\ \cmidrule(lr){2-21}
 ~ & \multirow{2}{*}{Template 3} & Easy & 2.0 & \cellcolor{gray!10}11.5 & 7.0 & \cellcolor{gray!10}0.5 & 5.0 & \cellcolor{gray!10}- & 6.0 & \cellcolor{gray!10}0.5 & 5.5 & \cellcolor{gray!10}0.0 & 5.0 & \cellcolor{gray!10}0.0 & 5.5 & \cellcolor{gray!10}3.5 & 4.5 & \cellcolor{gray!10}1.0 & 6.0 & \cellcolor{gray!10}0.5 \\ 
 ~ & ~ & Hard & 0.0 & \cellcolor{gray!10}1.5 & 0.0 & \cellcolor{gray!10}0.5 & 0.0 & \cellcolor{gray!10}- & 0.0 & \cellcolor{gray!10}0.0 & 0.0 & \cellcolor{gray!10}0.0 & 0.0 & \cellcolor{gray!10}0.5 & 0.5 & \cellcolor{gray!10}2.0 & 1.0 & \cellcolor{gray!10}0.5 & 0.0 & \cellcolor{gray!10}0.5 \\ 
 \midrule
 
 \multirow{6}{*}{\textbf{Gemma-3-12b-it}} & \multirow{2}{*}{Template 1} & Easy & 6.5 & \cellcolor{gray!10}45.0 & 10.0 & \cellcolor{gray!10}5.5 & 11.0 & \cellcolor{gray!10}- & 8.5 & \cellcolor{gray!10}0.0 & 8.5 & \cellcolor{gray!10}0.0 & 11.0 & \cellcolor{gray!10}0.0 & 9.5 & \cellcolor{gray!10}9.0 & 10.0 & \cellcolor{gray!10}7.0 & 7.0 & \cellcolor{gray!10}4.5 \\ 
 ~ & ~ & Hard & 2.0 & \cellcolor{gray!10}78.5 & 8.0 & \cellcolor{gray!10}10.0 & 6.0 & \cellcolor{gray!10}- & 5.0 & \cellcolor{gray!10}2.0 & 7.5 & \cellcolor{gray!10}2.0 & 6.0 & \cellcolor{gray!10}0.5 & 5.5 & \cellcolor{gray!10}22.5 & 7.0 & \cellcolor{gray!10}22.0 & 5.0 & \cellcolor{gray!10}9.0 \\ \cmidrule(lr){2-21}
 ~ & \multirow{2}{*}{Template 2} & Easy & 6.0 & \cellcolor{gray!10}45.5 & 10.0 & \cellcolor{gray!10}6.0 & 11.0 & \cellcolor{gray!10}- & 9.1 & \cellcolor{gray!10}0.0 & 8.5 & \cellcolor{gray!10}0.0 & 11.1 & \cellcolor{gray!10}0.0 & 9.5 & \cellcolor{gray!10}9.0 & 9.5 & \cellcolor{gray!10}7.0 & 7.0 & \cellcolor{gray!10}4.5 \\ 
 ~ & ~ & Hard & 2.0 & \cellcolor{gray!10}78.5 & 8.0 & \cellcolor{gray!10}10.0 & 6.0 & \cellcolor{gray!10}- & 5.0 & \cellcolor{gray!10}2.0 & 7.5 & \cellcolor{gray!10}2.0 & 6.0 & \cellcolor{gray!10}0.5 & 5.5 & \cellcolor{gray!10}22.5 & 7.0 & \cellcolor{gray!10}22.0 & 5.0 & \cellcolor{gray!10}9.0 \\ \cmidrule(lr){2-21}
 ~ & \multirow{2}{*}{Template 3} & Easy & 8.0 & \cellcolor{gray!10}44.5 & 11.0 & \cellcolor{gray!10}5.5 & 14.0 & \cellcolor{gray!10}- & 13.5 & \cellcolor{gray!10}0.0 & 14.5 & \cellcolor{gray!10}0.0 & 15.5 & \cellcolor{gray!10}0.0 & 11.5 & \cellcolor{gray!10}9.0 & 13.5 & \cellcolor{gray!10}7.0 & 9.5 & \cellcolor{gray!10}4.5 \\ 
 ~ & ~ & Hard & 2.0 & \cellcolor{gray!10}78.5 & 8.0 & \cellcolor{gray!10}10.0 & 6.0 & \cellcolor{gray!10}- & 5.0 & \cellcolor{gray!10}2.0 & 7.5 & \cellcolor{gray!10}2.0 & 6.0 & \cellcolor{gray!10}0.5 & 5.5 & \cellcolor{gray!10}22.5 & 7.0 & \cellcolor{gray!10}22.0 & 5.0 & \cellcolor{gray!10}9.0 \\ 
 \midrule
 
 \multirow{6}{*}{\textbf{Qwen3-VL-8B-Instruct}} & \multirow{2}{*}{Template 1} & Easy & 7.5 & \cellcolor{gray!10}44.5 & 10.5 & \cellcolor{gray!10}0.0 & 10.0 & \cellcolor{gray!10}- & 9.0 & \cellcolor{gray!10}0.0 & 8.5 & \cellcolor{gray!10}0.5 & 9.5 & \cellcolor{gray!10}0.0 & 9.5 & \cellcolor{gray!10}2.5 & 10.0 & \cellcolor{gray!10}1.5 & 9.5 & \cellcolor{gray!10}0.5 \\ 
 ~ & ~ & Hard & 7.0 & \cellcolor{gray!10}51.0 & 7.5 & \cellcolor{gray!10}2.0 & 7.0 & \cellcolor{gray!10}- & 5.5 & \cellcolor{gray!10}6.0 & 5.5 & \cellcolor{gray!10}3.0 & 7.0 & \cellcolor{gray!10}2.0 & 6.5 & \cellcolor{gray!10}8.0 & 6.0 & \cellcolor{gray!10}3.5 & 6.5 & \cellcolor{gray!10}2.5 \\ \cmidrule(lr){2-21}
 ~ & \multirow{2}{*}{Template 2} & Easy & 7.5 & \cellcolor{gray!10}48.0 & 10.5 & \cellcolor{gray!10}0.0 & 10.0 & \cellcolor{gray!10}- & 9.0 & \cellcolor{gray!10}0.5 & 9.0 & \cellcolor{gray!10}0.5 & 10.0 & \cellcolor{gray!10}0.0 & 9.5 & \cellcolor{gray!10}2.0 & 10.0 & \cellcolor{gray!10}1.5 & 10.0 & \cellcolor{gray!10}1.0 \\ 
 ~ & ~ & Hard & 7.0 & \cellcolor{gray!10}51.5 & 7.5 & \cellcolor{gray!10}2.5 & 7.0 & \cellcolor{gray!10}- & 6.5 & \cellcolor{gray!10}6.5 & 5.5 & \cellcolor{gray!10}3.0 & 7.5 & \cellcolor{gray!10}2.0 & 6.0 & \cellcolor{gray!10}8.5 & 6.0 & \cellcolor{gray!10}3.5 & 6.5 & \cellcolor{gray!10}2.5 \\ \cmidrule(lr){2-21}
 ~ & \multirow{2}{*}{Template 3} & Easy & 8.5 & \cellcolor{gray!10}47.5 & 16.5 & \cellcolor{gray!10}0.0 & 15.5 & \cellcolor{gray!10}- & 10.0 & \cellcolor{gray!10}0.0 & 11.0 & \cellcolor{gray!10}1.0 & 12.0 & \cellcolor{gray!10}0.0 & 12.5 & \cellcolor{gray!10}2.0 & 11.5 & \cellcolor{gray!10}1.5 & 12.5 & \cellcolor{gray!10}0.5 \\ 
 ~ & ~ & Hard & 6.5 & \cellcolor{gray!10}52.0 & 7.5 & \cellcolor{gray!10}3.0 & 7.0 & \cellcolor{gray!10}- & 6.0 & \cellcolor{gray!10}5.5 & 5.5 & \cellcolor{gray!10}3.0 & 7.5 & \cellcolor{gray!10}2.0 & 6.5 &\cellcolor{gray!10} 8.0 & 6.0 & \cellcolor{gray!10}3.5 & 6.5 & \cellcolor{gray!10}2.5 \\ 
 \midrule
 
 \multirow{6}{*}{\textbf{InternVL3\_5-8B}} & \multirow{2}{*}{Template 1} & Easy & 3.5 & \cellcolor{gray!10}66.5 & 11.5 & \cellcolor{gray!10}10.5 & 10.5 & \cellcolor{gray!10}- & 10.5 & \cellcolor{gray!10}0.0 & 10.5 & \cellcolor{gray!10}0.0 & 9.5 & \cellcolor{gray!10}0.0 & 9.5 & \cellcolor{gray!10}5.5 & 10.0 & \cellcolor{gray!10}4.0 & 9.5 & \cellcolor{gray!10}4.0 \\ 
 ~ & ~ & Hard & 3.5 & \cellcolor{gray!10}66.5 & 9.0 & \cellcolor{gray!10}11.0 & 7.5 & \cellcolor{gray!10}- & 7.5 & \cellcolor{gray!10}1.0 & 7.5 & \cellcolor{gray!10}1.0 & 7.5 & \cellcolor{gray!10}2.5 & 7.0 & \cellcolor{gray!10}6.5 & 7.0 & \cellcolor{gray!10}4.0 & 7.5 & \cellcolor{gray!10}3.0 \\ \cmidrule(lr){2-21}
 ~ & \multirow{2}{*}{Template 2} & Easy & 4.5 & \cellcolor{gray!10}60.0 & 10.0 & \cellcolor{gray!10}13.5 & 10.5 & \cellcolor{gray!10}- & 10.0 & \cellcolor{gray!10}0.0 & 10.5 & \cellcolor{gray!10}0.0 & 9.5 & \cellcolor{gray!10}0.0 & 9.5 & \cellcolor{gray!10}4.0 & 9.5 & \cellcolor{gray!10}2.0 & 9.5 & \cellcolor{gray!10}3.0 \\ 
 ~ & ~ & Hard & 4.0 & \cellcolor{gray!10}62.0 & 9.0 & \cellcolor{gray!10}14.5 & 7.5 & \cellcolor{gray!10}- & 7.5 & \cellcolor{gray!10}1.5 & 7.5 & \cellcolor{gray!10}1.5 & 7.5 & \cellcolor{gray!10}2.5 & 7.0 & \cellcolor{gray!10}6.0 & 7.0 & \cellcolor{gray!10}3.0 & 8.0 & \cellcolor{gray!10}2.5 \\ \cmidrule(lr){2-21}
 ~ & \multirow{2}{*}{Template 3} & Easy & 5.0 & \cellcolor{gray!10}61.0 & 16.0 & \cellcolor{gray!10}11.0 & 11.5 & \cellcolor{gray!10}- & 11.0 & \cellcolor{gray!10}0.0 & 12.0 & \cellcolor{gray!10}0.0 & 10.5 & \cellcolor{gray!10}0.0 & 12.0 & \cellcolor{gray!10}4.0 & 10.0 & \cellcolor{gray!10}2.5 & 11.5 & \cellcolor{gray!10}3.5 \\ 
 ~ & ~ & Hard & 4.0 & \cellcolor{gray!10}62.0 & 9.0 & \cellcolor{gray!10}14.5 & 7.5 & \cellcolor{gray!10}- & 7.5 & \cellcolor{gray!10}1.5 & 7.5 & \cellcolor{gray!10}1.5 & 7.5 & \cellcolor{gray!10}2.5 & 7.0 & \cellcolor{gray!10}6.0 & 7.0 & \cellcolor{gray!10}3.0 & 8.0 & \cellcolor{gray!10}2.5 \\ 
 \midrule
 
 \multirow{6}{*}{\textbf{Qianfan-VL-8B}} & \multirow{2}{*}{Template 1} & Easy & 2.0 & \cellcolor{gray!10}80.5 & 10.5 & \cellcolor{gray!10}5.0 & 9.5 & \cellcolor{gray!10}- & 9.0 & \cellcolor{gray!10}0.0 & 9.0 & \cellcolor{gray!10}0.0 & 9.0 & \cellcolor{gray!10}0.0 & 7.5 & \cellcolor{gray!10}13.5 & 7.5 & \cellcolor{gray!10}7.5 & 7.5 & \cellcolor{gray!10}10.5 \\ 
 ~ & ~ & Hard & 2.0 & \cellcolor{gray!10}83.5 & 7.0 & \cellcolor{gray!10}11.0 & 7.5 & \cellcolor{gray!10}- & 7.5 & \cellcolor{gray!10}1.5 & 6.0 & \cellcolor{gray!10}1.0 & 7.0 & \cellcolor{gray!10}1.5 & 7.0 & \cellcolor{gray!10}13.5 & 7.0 & \cellcolor{gray!10}6.5 & 7.0 & \cellcolor{gray!10}5.0 \\ \cmidrule(lr){2-21}
 ~ & \multirow{2}{*}{Template 2} & Easy & 2.0 & \cellcolor{gray!10}81.5 & 10.5 & \cellcolor{gray!10}6.5 & 9.5 & \cellcolor{gray!10}- & 9.0 & \cellcolor{gray!10}0.0 & 9.0 & \cellcolor{gray!10}0.0 & 9.0 & \cellcolor{gray!10}0.0 & 7.5 & \cellcolor{gray!10}14.0 & 7.0 & \cellcolor{gray!10}7.5 & 7.5 & \cellcolor{gray!10}10.5 \\ 
 ~ & ~ & Hard & 2.0 & \cellcolor{gray!10}84.0 & 6.5 & \cellcolor{gray!10}13.0 & 7.5 & \cellcolor{gray!10}- & 7.5 & \cellcolor{gray!10}1.5 & 6.0 & \cellcolor{gray!10}1.0 & 7.0 & \cellcolor{gray!10}1.5 & 7.0 & \cellcolor{gray!10}14.5 & 6.5 & \cellcolor{gray!10}6.5 & 7.0 & \cellcolor{gray!10}5.0 \\ \cmidrule(lr){2-21}
 ~ & \multirow{2}{*}{Template 3} & Easy & 3.5 & \cellcolor{gray!10}82.0 & 13.5 & \cellcolor{gray!10}6.5 & 13.0 & \cellcolor{gray!10}- & 12.0 & \cellcolor{gray!10}0.0 & 13.0 & \cellcolor{gray!10}0.0 & 12.0 & \cellcolor{gray!10}0.0 & 10.0 & \cellcolor{gray!10}13.0 & 9.0 & \cellcolor{gray!10}7.0 & 9.5 & \cellcolor{gray!10}10.5 \\ 
 ~ & ~ & Hard & 2.0 & \cellcolor{gray!10}84.0 & 6.5 & \cellcolor{gray!10}13.0 & 7.5 & \cellcolor{gray!10}- & 7.5 & \cellcolor{gray!10}1.5 & 6.0 & \cellcolor{gray!10}1.0 & 7.0 & \cellcolor{gray!10}1.5 & 7.0 & \cellcolor{gray!10}14.5 & 6.5 & \cellcolor{gray!10}6.5 & 7.0 & \cellcolor{gray!10}5.0 \\ 
 \midrule
 
 \multirow{6}{*}{\textbf{Gemma-3-4b-it}} & \multirow{2}{*}{Template 1} & Easy & 7.5 & \cellcolor{gray!10}43.5 & 10.5 & \cellcolor{gray!10}0.0 & 10.0 & \cellcolor{gray!10}- & 10.5 & \cellcolor{gray!10}0.5 & 9.0 & \cellcolor{gray!10}0.5 & 10.5 & \cellcolor{gray!10}0.0 & 8.5 & \cellcolor{gray!10}17.0 & 8.5 & \cellcolor{gray!10}13.0 & 9.5 & \cellcolor{gray!10}4.5 \\ 
 ~ & ~ & Hard & 5.0 & \cellcolor{gray!10}51.5 & 7.5 & \cellcolor{gray!10}2.5 & 7.5 & \cellcolor{gray!10}- & 7.5 & \cellcolor{gray!10}2.0 & 7.5 & \cellcolor{gray!10}3.5 & 7.5 & \cellcolor{gray!10}2.0 & 6.5 & \cellcolor{gray!10}15.0 & 7.0 & \cellcolor{gray!10}10.5 & 7.0 & \cellcolor{gray!10}6.0 \\ \cmidrule(lr){2-21}
 ~ & \multirow{2}{*}{Template 2} & Easy & 7.5 & \cellcolor{gray!10}43.5 & 10.5 & \cellcolor{gray!10}0.0 & 10.0 & \cellcolor{gray!10}- & 10.5 & \cellcolor{gray!10}0.5 & 9.0 & \cellcolor{gray!10}0.5 & 10.5 & \cellcolor{gray!10}0.0 & 8.5 & \cellcolor{gray!10}18.0 & 8.5 & \cellcolor{gray!10}13.5 & 9.5 & \cellcolor{gray!10}4.5 \\ 
 ~ & ~ & Hard & 5.0 & \cellcolor{gray!10}51.5 & 7.5 & \cellcolor{gray!10}2.5 & 7.5 & \cellcolor{gray!10}- & 7.5 & \cellcolor{gray!10}2.0 & 7.5 & \cellcolor{gray!10}3.5 & 7.5 & \cellcolor{gray!10}2.0 & 6.5 & \cellcolor{gray!10}15.0 & 7.0 & \cellcolor{gray!10}10.5 & 7.0 & \cellcolor{gray!10}6.0 \\ \cmidrule(lr){2-21}
 ~ & \multirow{2}{*}{Template 3} & Easy & 7.5 & \cellcolor{gray!10}43.5 & 10.5 & \cellcolor{gray!10}0.0 & 10.5 &\cellcolor{gray!10} - & 11.0 & \cellcolor{gray!10}0.5 & 9.0 & \cellcolor{gray!10}0.5 & 10.5 & \cellcolor{gray!10}0.0 & 8.5 & \cellcolor{gray!10}18.0 & 8.5 & \cellcolor{gray!10}13.0 & 9.5 & \cellcolor{gray!10}4.5 \\ 
 ~ & ~ & Hard & 5.0 & \cellcolor{gray!10}51.5 & 7.5 & \cellcolor{gray!10}2.5 & 7.5 & \cellcolor{gray!10}- & 7.5 & \cellcolor{gray!10}2.0 & 7.5 & \cellcolor{gray!10}3.5 & 7.5 & \cellcolor{gray!10}2.0 & 6.5 & \cellcolor{gray!10}15.0 & 7.0 & \cellcolor{gray!10}10.5 & 7.0 & \cellcolor{gray!10}6.0 \\ 
 \midrule
 
 \multirow{6}{*}{\textbf{Qianfan-VL-3B}} & \multirow{2}{*}{Template 1} & Easy & 2.0 & \cellcolor{gray!10}67.0 & 10.0 & \cellcolor{gray!10}3.0 & 7.0 & \cellcolor{gray!10}- & 8.5 & \cellcolor{gray!10}0.0 & 9.0 & \cellcolor{gray!10}0.0 & 8.0 & \cellcolor{gray!10}0.0 & 9.0 & \cellcolor{gray!10}3.5 & 9.0 & \cellcolor{gray!10}1.0 & 9.0 & \cellcolor{gray!10}0.5 \\ 
 ~ & ~ & Hard & 2.5 & \cellcolor{gray!10}72.0 & 7.5 & \cellcolor{gray!10}9.0 & 7.0 & \cellcolor{gray!10}- & 6.5 & \cellcolor{gray!10}1.5 & 7.5 & \cellcolor{gray!10}1.5 & 6.5 & \cellcolor{gray!10}1.5 & 8.0 & \cellcolor{gray!10}5.0 & 7.0 & \cellcolor{gray!10}4.0 & 7.0 & \cellcolor{gray!10}2.5 \\ \cmidrule(lr){2-21}
 ~ & \multirow{2}{*}{Template 2} & Easy & 2.0 & \cellcolor{gray!10}68.5 & 9.5 & \cellcolor{gray!10}4.0 & 7.0 & \cellcolor{gray!10}- & 8.5 & \cellcolor{gray!10}0.0 & 9.0 & \cellcolor{gray!10}0.0 & 7.5 & \cellcolor{gray!10}0.0 & 9.0 & \cellcolor{gray!10}5.0 & 9.0 & \cellcolor{gray!10}1.0 & 9.0 & \cellcolor{gray!10}0.5 \\ 
 ~ & ~ & Hard & 2.5 & \cellcolor{gray!10}72.5 & 7.5 & \cellcolor{gray!10}9.0 & 7.0 & \cellcolor{gray!10}- & 6.5 & \cellcolor{gray!10}1.5 & 7.5 & \cellcolor{gray!10}1.5 & 6.5 & \cellcolor{gray!10}1.5 & 8.0 & \cellcolor{gray!10}5.5 & 7.0 & \cellcolor{gray!10}4.0 & 7.0 & \cellcolor{gray!10}2.5 \\ \cmidrule(lr){2-21}
 ~ & \multirow{2}{*}{Template 3} & Easy & 5.0 & \cellcolor{gray!10}66.5 & 12.5 & \cellcolor{gray!10}3.5 & 10.0 & \cellcolor{gray!10}- & 11.5 & \cellcolor{gray!10}0.0 & 13.0 & \cellcolor{gray!10}0.0 & 10.0 & \cellcolor{gray!10}0.0 & 12.5 & \cellcolor{gray!10}4.0 & 12.5 & \cellcolor{gray!10}1.0 & 12.0 & \cellcolor{gray!10}0.5 \\ 
 ~ & ~ & Hard & 2.5 & \cellcolor{gray!10}72.5 & 7.5 & \cellcolor{gray!10}9.0 & 7.0 & \cellcolor{gray!10}- & 6.5 & \cellcolor{gray!10}1.5 & 7.5 & \cellcolor{gray!10}1.5 & 6.5 & \cellcolor{gray!10}1.5 & 8.0 & \cellcolor{gray!10}5.5 & 7.0 & \cellcolor{gray!10}4.0 & 7.0 & \cellcolor{gray!10}2.5 \\ 
 
 \bottomrule
 \end{tabular}%
 }%
\end{table*}

\section{Impact of Masking Ratios in Masked Stroop}
\label{appendix:masking-impact}

To provide a granular analysis of how visual interference affects model performance, we evaluate both proprietary and open-source models on the “Masked Stroop” task. This task utilizes the Embedded Stroop effect, where text is obscured using three distinct masking geometries: \textbf{Square}, \textbf{Star}, and \textbf{Underline}, each applied at three varying density ratios (25\%, 50\%, and 75\%). 

The comprehensive quantitative results for proprietary models (including GPT-5.2, GPT-4o, and Gemini variants) are reported in Table~\ref{tab:mask_que_prop}. Correspondingly, the performance of open-source models (such as DeepSeek-VL2, InternVL3.5, and Qwen3 series) across different prompt templates and difficulty levels is detailed in Table~\ref{tab:mask_que_open}. 

These results demonstrate that as the masking ratio increases, most models experience a non-linear shift in accuracy (Acc\%) and Stroop hallucination rates (SHR\%). Notably, even under high-density masking (75\%), certain thinking-based models maintain relatively stable reasoning chains, whereas smaller open-source models exhibit higher sensitivity to the specific geometry of the mask, particularly in the ``Hard'' difficulty settings.

\begin{table*}[h!]
\caption{\textbf{Quantitative Evaluation on the masked WCIT Benchmark with Proprietary Models.} Detailed performance across all mask types and ratios (25\%, 50\%, 75\%) for the Embedded Stroop task.}
\label{tab:mask_que_prop}
\centering
\tiny
\setlength{\tabcolsep}{1.5pt}
\resizebox{\textwidth}{!}{%
\begin{tabular}{lllcccccccccccccccccc}
\toprule
\multirow{3}{*}{\textbf{Model}} & \multirow{3}{*}{\textbf{Prompt}} & \multirow{3}{*}{\textbf{Diff}} & \multicolumn{6}{c}{\textbf{Square}} & \multicolumn{6}{c}{\textbf{Star}} & \multicolumn{6}{c}{\textbf{Line}} \\
\cmidrule(lr){4-9} \cmidrule(lr){10-15} \cmidrule(lr){16-21}
& & & \multicolumn{2}{c}{\textbf{25\%}} & \multicolumn{2}{c}{\textbf{50\%}} & \multicolumn{2}{c}{\textbf{75\%}} & \multicolumn{2}{c}{\textbf{25\%}} & \multicolumn{2}{c}{\textbf{50\%}} & \multicolumn{2}{c}{\textbf{75\%}} & \multicolumn{2}{c}{\textbf{25\%}} & \multicolumn{2}{c}{\textbf{50\%}} & \multicolumn{2}{c}{\textbf{75\%}} \\
\cmidrule(lr){4-5} \cmidrule(lr){6-7} \cmidrule(lr){8-9} \cmidrule(lr){10-11} \cmidrule(lr){12-13} \cmidrule(lr){14-15} \cmidrule(lr){16-17} \cmidrule(lr){18-19} \cmidrule(lr){20-21}
& & & Acc\% & SHR\% & Acc\% & SHR\% & Acc\% & SHR\% & Acc\% & SHR\% & Acc\% & SHR\% & Acc\% & SHR\% & Acc\% & SHR\% & Acc\% & SHR\% & Acc\% & SHR\% \\
\midrule

 \multirow{6}{*}{\textbf{GPT-5.2}} & \multirow{2}{*}{Template 1} & Easy & 10.5 & \cellcolor{gray!10}2.5 & 11.5 & \cellcolor{gray!10}0.5 & 12.0 & \cellcolor{gray!10}0.5 & 12.0 & \cellcolor{gray!10}1.0 & 12.5 & \cellcolor{gray!10}0.0 & 11.0 & \cellcolor{gray!10}0.0 & 11.0 & \cellcolor{gray!10}1.5 & 11.5 & \cellcolor{gray!10}0.5 & 10.5 & \cellcolor{gray!10}0.0 \\
  ~ &  ~ & Hard & 7.5 & \cellcolor{gray!10}10.5 & 7.5 & \cellcolor{gray!10}3.5 & 8.0 & \cellcolor{gray!10}2.5 & 7.5 & \cellcolor{gray!10}7.0 & 8.0 & \cellcolor{gray!10}1.5 & 8.0 & \cellcolor{gray!10}1.5 & 8.0 & \cellcolor{gray!10}5.0 & 7.0 & \cellcolor{gray!10}2.5 & 8.0 & \cellcolor{gray!10}1.5 \\ \cmidrule(lr){2-21}
  ~ & \multirow{2}{*}{Template 2} & Easy & 11.5 & \cellcolor{gray!10}4.0 & 12.5 & \cellcolor{gray!10}0.5 & 12.0 & \cellcolor{gray!10}0.0 & 12.0 & \cellcolor{gray!10}2.0 & 11.5 & \cellcolor{gray!10}0.0 & 10.5 & \cellcolor{gray!10}0.0 & 11.0 & \cellcolor{gray!10}0.5 & 11.0 & \cellcolor{gray!10}0.5 & 10.0 & \cellcolor{gray!10}0.0 \\
  ~ &  ~ & Hard & 7.5 & \cellcolor{gray!10}9.0 & 7.5 & \cellcolor{gray!10}2.5 & 7.0 & \cellcolor{gray!10}2.5 & 7.5 & \cellcolor{gray!10}6.5 & 8.5 & \cellcolor{gray!10}1.5 & 8.0 & \cellcolor{gray!10}2.0 & 7.5 & \cellcolor{gray!10}4.5 & 7.5 & \cellcolor{gray!10}2.5 & 8.0 & \cellcolor{gray!10}1.5 \\ \cmidrule(lr){2-21}
  ~ & \multirow{2}{*}{Template 3} & Easy & 20.0 & \cellcolor{gray!10}1.5 & 20.0 & \cellcolor{gray!10}0.5 & 20.5 & \cellcolor{gray!10}0.0 & 20.5 & \cellcolor{gray!10}1.5 & 21.0 & \cellcolor{gray!10}0.0 & 20.5 & \cellcolor{gray!10}0.0 & 19.5 & \cellcolor{gray!10}0.5 & 22.0 & \cellcolor{gray!10}0.0 & 17.0 & \cellcolor{gray!10}0.0 \\
  ~ &  ~ & Hard & 8.0 & \cellcolor{gray!10}10.0 & 8.5 & \cellcolor{gray!10}3.5 & 8.5 & \cellcolor{gray!10}2.5 & 7.5 & \cellcolor{gray!10}5.5 & 8.0 & \cellcolor{gray!10}1.5 & 6.5 & \cellcolor{gray!10}1.5 & 7.5 & \cellcolor{gray!10}5.0 & 7.5 & \cellcolor{gray!10}2.5 & 8.0 & \cellcolor{gray!10}1.5 \\ \midrule
 \multirow{6}{*}{\textbf{GPT-4o}} & \multirow{2}{*}{Template 1} & Easy & 9.5 & \cellcolor{gray!10}4.5 & 11.5 & \cellcolor{gray!10}2.5 & 9.5 & \cellcolor{gray!10}0.0 & 9.5 & \cellcolor{gray!10}4.0 & 11.5 & \cellcolor{gray!10}4.0 & 12.0 & \cellcolor{gray!10}0.5 & 10.0 & \cellcolor{gray!10}9.5 & 9.5 & \cellcolor{gray!10}6.0 & 11.0 & \cellcolor{gray!10}0.5 \\
  ~ &  ~ & Hard & 7.0 & \cellcolor{gray!10}17.5 & 7.0 & \cellcolor{gray!10}8.0 & 8.0 & \cellcolor{gray!10}3.0 & 7.0 & \cellcolor{gray!10}24.5 & 7.5 & \cellcolor{gray!10}8.5 & 7.5 & \cellcolor{gray!10}3.0 & 7.0 & \cellcolor{gray!10}22.0 & 7.5 & \cellcolor{gray!10}7.0 & 7.5 & \cellcolor{gray!10}2.0 \\ \cmidrule(lr){2-21}
  ~ & \multirow{2}{*}{Template 2} & Easy & 9.5 & \cellcolor{gray!10}8.0 & 10.5 & \cellcolor{gray!10}3.0 & 10.5 & \cellcolor{gray!10}0.5 & 9.5 & \cellcolor{gray!10}6.0 & 12.5 & \cellcolor{gray!10}5.5 & 12.0 & \cellcolor{gray!10}0.5 & 9.5 & \cellcolor{gray!10}10.5 & 8.0 & \cellcolor{gray!10}6.5 & 9.5 & \cellcolor{gray!10}1.5 \\
  ~ &  ~ & Hard & 6.5 & \cellcolor{gray!10}26.0 & 6.5 & \cellcolor{gray!10}8.0 & 6.5 & \cellcolor{gray!10}3.0 & 7.0 & \cellcolor{gray!10}24.5 & 7.0 & \cellcolor{gray!10}10.5 & 7.5 & \cellcolor{gray!10}3.5 & 7.5 & \cellcolor{gray!10}25.5 & 7.5 & \cellcolor{gray!10}6.0 & 7.0 & \cellcolor{gray!10}2.5 \\ \cmidrule(lr){2-21}
  ~ & \multirow{2}{*}{Template 3} & Easy & 16.5 & \cellcolor{gray!10}5.5 & 18.0 & \cellcolor{gray!10}2.5 & 15.5 & \cellcolor{gray!10}0.0 & 17.0 & \cellcolor{gray!10}7.5 & 17.5 & \cellcolor{gray!10}4.0 & 18.5 & \cellcolor{gray!10}1.0 & 15.5 & \cellcolor{gray!10}7.5 & 15.0 & \cellcolor{gray!10}8.0 & 15.5 & \cellcolor{gray!10}1.5 \\
  ~ &  ~ & Hard & 5.5 & \cellcolor{gray!10}22.0 & 7.5 & \cellcolor{gray!10}10.0 & 7.0 & \cellcolor{gray!10}3.0 & 8.0 & \cellcolor{gray!10}25.5 & 7.5 & \cellcolor{gray!10}10.0 & 7.5 & \cellcolor{gray!10}4.0 & 7.0 & \cellcolor{gray!10}26.0 & 7.0 & \cellcolor{gray!10}7.0 & 8.0 & \cellcolor{gray!10}2.0 \\ \midrule
 \multirow{6}{*}{\textbf{Gemini-3-flash-thinking}} & \multirow{2}{*}{Template 1} & Easy & 12.0 & \cellcolor{gray!10}0.0 & 14.0 & \cellcolor{gray!10}1.0 & 12.5 & \cellcolor{gray!10}0.0 & 14.5 & \cellcolor{gray!10}0.0 & 10.5 & \cellcolor{gray!10}2.5 & 10.5 & \cellcolor{gray!10}1.5 & 13.0 & \cellcolor{gray!10}0.0 & 13.5 & \cellcolor{gray!10}4.0 & 12.5 & \cellcolor{gray!10}2.0 \\
  ~ &  ~ & Hard & 10.5 & \cellcolor{gray!10}15.5 & 10.5 & \cellcolor{gray!10}6.0 & 10.5 & \cellcolor{gray!10}2.5 & 10.5 & \cellcolor{gray!10}12.0 & 8.0 & \cellcolor{gray!10}13.0 & 9.0 & \cellcolor{gray!10}7.0 & 8.0 & \cellcolor{gray!10}16.0 & 8.0 & \cellcolor{gray!10}14.0 & 8.0 & \cellcolor{gray!10}9.0 \\ \cmidrule(lr){2-21}
  ~ & \multirow{2}{*}{Template 2} & Easy & 13.5 & \cellcolor{gray!10}0.0 & 13.0 & \cellcolor{gray!10}0.0 & 13.5 & \cellcolor{gray!10}0.0 & 13.5 & \cellcolor{gray!10}0.0 & 11.0 & \cellcolor{gray!10}2.0 & 10.5 & \cellcolor{gray!10}1.0 & 14.0 & \cellcolor{gray!10}0.0 & 12.5 & \cellcolor{gray!10}4.5 & 12.0 & \cellcolor{gray!10}1.5 \\
  ~ &  ~ & Hard & 8.0 & \cellcolor{gray!10}13.5 & 10.5 & \cellcolor{gray!10}5.0 & 8.0 & \cellcolor{gray!10}2.5 & 7.5 & \cellcolor{gray!10}10.5 & 10.5 & \cellcolor{gray!10}14.5 & 8.0 & \cellcolor{gray!10}5.5 & 8.0 & \cellcolor{gray!10}15.0 & 9.0 & \cellcolor{gray!10}12.0 & 7.0 & \cellcolor{gray!10}5.5 \\ \cmidrule(lr){2-21}
  ~ & \multirow{2}{*}{Template 3} & Easy & 23.5 & \cellcolor{gray!10}0.0 & 21.0 & \cellcolor{gray!10}0.0 & 20.5 & \cellcolor{gray!10}0.0 & 23.0 & \cellcolor{gray!10}0.0 & 19.0 & \cellcolor{gray!10}3.0 & 20.0 & \cellcolor{gray!10}1.0 & 20.5 & \cellcolor{gray!10}0.0 & 18.5 & \cellcolor{gray!10}1.5 & 18.5 & \cellcolor{gray!10}0.5 \\
  ~ &  ~ & Hard & 9.5 & \cellcolor{gray!10}16.0 & 8.5 & \cellcolor{gray!10}5.0 & 6.0 & \cellcolor{gray!10}2.5 & 11.5 & \cellcolor{gray!10}12.5 & 9.5 & \cellcolor{gray!10}15.0 & 7.5 & \cellcolor{gray!10}4.5 & 8.5 & \cellcolor{gray!10}15.5 & 9.0 & \cellcolor{gray!10}12.5 & 7.0 & \cellcolor{gray!10}6.0 \\ \midrule
 \multirow{6}{*}{\textbf{Gemini-2.5-flash}} & \multirow{2}{*}{Template 1} & Easy & 12.0 & \cellcolor{gray!10}0.5 & 11.0 & \cellcolor{gray!10}0.0 & 10.0 & \cellcolor{gray!10}0.5 & 14.5 & \cellcolor{gray!10}0.0 & 12.5 & \cellcolor{gray!10}0.0 & 12.0 & \cellcolor{gray!10}0.0 & 11.0 & \cellcolor{gray!10}0.0 & 10.5 & \cellcolor{gray!10}2.5 & 10.0 & \cellcolor{gray!10}0.0 \\
  ~ &  ~ & Hard & 6.0 & \cellcolor{gray!10}9.5 & 7.0 & \cellcolor{gray!10}3.0 & 6.0 & \cellcolor{gray!10}2.5 & 8.0 & \cellcolor{gray!10}8.0 & 7.5 & \cellcolor{gray!10}4.5 & 8.0 & \cellcolor{gray!10}1.5 & 7.5 & \cellcolor{gray!10}14.0 & 9.0 & \cellcolor{gray!10}3.5 & 8.5 & \cellcolor{gray!10}2.0 \\ \cmidrule(lr){2-21}
  ~ & \multirow{2}{*}{Template 2} & Easy & 13.0 & \cellcolor{gray!10}0.0 & 10.5 & \cellcolor{gray!10}0.5 & 10.0 & \cellcolor{gray!10}0.0 & 11.5 & \cellcolor{gray!10}0.0 & 13.0 & \cellcolor{gray!10}0.5 & 12.0 & \cellcolor{gray!10}0.0 & 12.0 & \cellcolor{gray!10}0.5 & 10.5 & \cellcolor{gray!10}0.5 & 12.0 & \cellcolor{gray!10}0.0 \\
  ~ &  ~ & Hard & 8.0 & \cellcolor{gray!10}7.0 & 8.5 & \cellcolor{gray!10}5.0 & 7.5 & \cellcolor{gray!10}1.5 & 9.5 & \cellcolor{gray!10}10.0 & 10.0 & \cellcolor{gray!10}5.0 & 8.0 & \cellcolor{gray!10}1.5 & 9.5 & \cellcolor{gray!10}11.5 & 8.0 & \cellcolor{gray!10}2.5 & 7.0 & \cellcolor{gray!10}1.5 \\ \cmidrule(lr){2-21}
  ~ & \multirow{2}{*}{Template 3} & Easy & 16.5 & \cellcolor{gray!10}0.5 & 15.5 & \cellcolor{gray!10}0.5 & 12.5 & \cellcolor{gray!10}0.0 & 17.5 & \cellcolor{gray!10}0.0 & 17.5 & \cellcolor{gray!10}0.5 & 18.0 & \cellcolor{gray!10}0.0 & 16.5 & \cellcolor{gray!10}0.5 & 18.0 & \cellcolor{gray!10}0.5 & 19.0 & \cellcolor{gray!10}0.0 \\
  ~ &  ~ & Hard & 6.5 & \cellcolor{gray!10}9.0 & 8.0 & \cellcolor{gray!10}4.5 & 7.0 & \cellcolor{gray!10}2.0 & 9.0 & \cellcolor{gray!10}11.0 & 8.0 & \cellcolor{gray!10}4.5 & 8.0 & \cellcolor{gray!10}1.0 & 7.5 & \cellcolor{gray!10}11.0 & 8.0 & \cellcolor{gray!10}4.0 & 7.0 & \cellcolor{gray!10}2.0 \\ \midrule
 \multirow{6}{*}{\textbf{Qwen3-vl-plus}} & \multirow{2}{*}{Template 1} & Easy & 10.0 & \cellcolor{gray!10}4.5 & 5.5 & \cellcolor{gray!10}0.0 & 1.0 & \cellcolor{gray!10}0.0 & 10.0 & \cellcolor{gray!10}1.0 & 10.0 & \cellcolor{gray!10}0.0 & 7.5 & \cellcolor{gray!10}0.0 & 10.0 & \cellcolor{gray!10}3.0 & 7.0 & \cellcolor{gray!10}1.0 & 6.5 & \cellcolor{gray!10}0.0 \\
  ~ &  ~ & Hard & 6.0 & \cellcolor{gray!10}5.5 & 4.5 & \cellcolor{gray!10}1.5 & 2.0 & \cellcolor{gray!10}0.0 & 8.0 & \cellcolor{gray!10}4.5 & 6.0 & \cellcolor{gray!10}2.0 & 6.0 & \cellcolor{gray!10}0.5 & 7.0 & \cellcolor{gray!10}3.0 & 6.0 & \cellcolor{gray!10}1.0 & 5.5 & \cellcolor{gray!10}0.5 \\ \cmidrule(lr){2-21}
  ~ & \multirow{2}{*}{Template 2} & Easy & 9.0 & \cellcolor{gray!10}4.5 & 5.0 & \cellcolor{gray!10}0.0 & 1.5 & \cellcolor{gray!10}0.0 & 10.0 & \cellcolor{gray!10}1.5 & 9.5 & \cellcolor{gray!10}0.0 & 7.0 & \cellcolor{gray!10}0.0 & 10.0 & \cellcolor{gray!10}1.0 & 6.5 & \cellcolor{gray!10}0.5 & 4.5 & \cellcolor{gray!10}0.0 \\
  ~ &  ~ & Hard & 6.0 & \cellcolor{gray!10}4.5 & 4.5 & \cellcolor{gray!10}0.5 & 1.5 & \cellcolor{gray!10}0.0 & 7.5 & \cellcolor{gray!10}4.5 & 6.0 & \cellcolor{gray!10}2.0 & 3.5 & \cellcolor{gray!10}0.5 & 7.0 & \cellcolor{gray!10}3.5 & 6.5 & \cellcolor{gray!10}1.5 & 6.0 & \cellcolor{gray!10}0.0 \\ \cmidrule(lr){2-21}
  ~ & \multirow{2}{*}{Template 3} & Easy & 11.5 & \cellcolor{gray!10}5.5 & 7.0 & \cellcolor{gray!10}0.0 & 1.5 & \cellcolor{gray!10}0.0 & 11.5 & \cellcolor{gray!10}2.0 & 12.0 & \cellcolor{gray!10}0.0 & 10.0 & \cellcolor{gray!10}0.0 & 12.0 & \cellcolor{gray!10}1.5 & 9.5 & \cellcolor{gray!10}0.5 & 7.5 & \cellcolor{gray!10}0.0 \\
  ~ &  ~ & Hard & 6.0 & \cellcolor{gray!10}4.5 & 4.0 & \cellcolor{gray!10}0.5 & 1.5 & \cellcolor{gray!10}0.0 & 7.5 & \cellcolor{gray!10}5.0 & 6.0 & \cellcolor{gray!10}2.0 & 3.5 & \cellcolor{gray!10}0.5 & 7.0 & \cellcolor{gray!10}3.5 & 6.5 & \cellcolor{gray!10}1.5 & 6.0 & \cellcolor{gray!10}0.0 \\ \midrule
 \multirow{6}{*}{\textbf{GLM-4.5v}} & \multirow{2}{*}{Template 1} & Easy & 2.0 & \cellcolor{gray!10}0.0 & 1.5 & \cellcolor{gray!10}0.0 & 1.0 & \cellcolor{gray!10}0.0 & 1.5 & \cellcolor{gray!10}0.5 & 2.5 & \cellcolor{gray!10}0.0 & 2.5 & \cellcolor{gray!10}0.0 & 3.0 & \cellcolor{gray!10}0.0 & 2.5 & \cellcolor{gray!10}0.0 & 1.0 & \cellcolor{gray!10}0.0 \\
  ~ &  ~ & Hard & 2.0 & \cellcolor{gray!10}1.0 & 2.0 & \cellcolor{gray!10}0.5 & 0.5 & \cellcolor{gray!10}0.5 & 2.5 & \cellcolor{gray!10}0.5 & 2.0 & \cellcolor{gray!10}0.0 & 2.5 & \cellcolor{gray!10}0.0 & 2.5 & \cellcolor{gray!10}0.5 & 2.0 & \cellcolor{gray!10}0.5 & 1.5 & \cellcolor{gray!10}0.0 \\ \cmidrule(lr){2-21}
  ~ & \multirow{2}{*}{Template 2} & Easy & 4.5 & \cellcolor{gray!10}0.5 & 2.0 & \cellcolor{gray!10}0.0 & 0.5 & \cellcolor{gray!10}0.0 & 2.5 & \cellcolor{gray!10}0.5 & 2.5 & \cellcolor{gray!10}0.0 & 3.0 & \cellcolor{gray!10}0.0 & 2.5 & \cellcolor{gray!10}0.0 & 2.5 & \cellcolor{gray!10}0.0 & 1.5 & \cellcolor{gray!10}0.0 \\
  ~ &  ~ & Hard & 2.0 & \cellcolor{gray!10}3.0 & 3.0 & \cellcolor{gray!10}2.0 & 0.5 & \cellcolor{gray!10}0.5 & 2.0 & \cellcolor{gray!10}0.5 & 2.0 & \cellcolor{gray!10}0.5 & 3.0 & \cellcolor{gray!10}0.0 & 3.0 & \cellcolor{gray!10}1.5 & 3.0 & \cellcolor{gray!10}0.0 & 3.0 & \cellcolor{gray!10}1.0 \\ \cmidrule(lr){2-21}
  ~ & \multirow{2}{*}{Template 3} & Easy & 6.0 & \cellcolor{gray!10}0.0 & 2.5 & \cellcolor{gray!10}0.0 & 2.0 & \cellcolor{gray!10}0.0 & 2.5 & \cellcolor{gray!10}0.0 & 3.0 & \cellcolor{gray!10}0.0 & 4.5 & \cellcolor{gray!10}0.0 & 4.5 & \cellcolor{gray!10}0.0 & 3.0 & \cellcolor{gray!10}0.0 & 1.5 & \cellcolor{gray!10}0.0 \\
  ~ &  ~ & Hard & 3.5 & \cellcolor{gray!10}2.0 & 3.0 & \cellcolor{gray!10}0.0 & 1.0 & \cellcolor{gray!10}0.0 & 2.5 & \cellcolor{gray!10}2.0 & 2.0 & \cellcolor{gray!10}0.0 & 1.0 & \cellcolor{gray!10}0.0 & 2.0 & \cellcolor{gray!10}0.5 & 2.5 & \cellcolor{gray!10}0.5 & 3.0 & \cellcolor{gray!10}0.5 \\
 \bottomrule
\end{tabular}
}%
\end{table*}

\begin{table*}[h!]
\caption{\textbf{Quantitative Evaluation on the masked WCIT Benchmark with Open Source Models.} Detailed performance across all mask types and ratios (25\%, 50\%, 75\%) for the Embedded Stroop task.}
\label{tab:mask_que_open}
\centering
\tiny
\setlength{\tabcolsep}{1.5pt}
\resizebox{\textwidth}{!}{%
\begin{tabular}{lllcccccccccccccccccc}
\toprule
\multirow{3}{*}{\textbf{Model}} & \multirow{3}{*}{\textbf{Prompt}} & \multirow{3}{*}{\textbf{Diff}} & \multicolumn{6}{c}{\textbf{Square}} & \multicolumn{6}{c}{\textbf{Star}} & \multicolumn{6}{c}{\textbf{Line}} \\
\cmidrule(lr){4-9} \cmidrule(lr){10-15} \cmidrule(lr){16-21}
& & & \multicolumn{2}{c}{\textbf{25\%}} & \multicolumn{2}{c}{\textbf{50\%}} & \multicolumn{2}{c}{\textbf{75\%}} & \multicolumn{2}{c}{\textbf{25\%}} & \multicolumn{2}{c}{\textbf{50\%}} & \multicolumn{2}{c}{\textbf{75\%}} & \multicolumn{2}{c}{\textbf{25\%}} & \multicolumn{2}{c}{\textbf{50\%}} & \multicolumn{2}{c}{\textbf{75\%}} \\
\cmidrule(lr){4-5} \cmidrule(lr){6-7} \cmidrule(lr){8-9} \cmidrule(lr){10-11} \cmidrule(lr){12-13} \cmidrule(lr){14-15} \cmidrule(lr){16-17} \cmidrule(lr){18-19} \cmidrule(lr){20-21}
& & & Acc\% & SHR\% & Acc\% & SHR\% & Acc\% & SHR\% & Acc\% & SHR\% & Acc\% & SHR\% & Acc\% & SHR\% & Acc\% & SHR\% & Acc\% & SHR\% & Acc\% & SHR\% \\
\midrule

 \multirow{6}{*}{\textbf{Deepseek-vl2}} & \multirow{2}{*}{Template 1} & Easy & 9.5 & \cellcolor{gray!10}12.5 & 11.0 & \cellcolor{gray!10}0.5 & 10.5 & \cellcolor{gray!10}0.0 & 9.0 & \cellcolor{gray!10}7.0 & 10.5 & \cellcolor{gray!10}0.5 & 10.5 & \cellcolor{gray!10}0.0 & 9.5 & \cellcolor{gray!10}3.5 & 10.5 & \cellcolor{gray!10}0.5 & 10.5 & \cellcolor{gray!10}0.0 \\
  ~ &  ~ & Hard & 7.5 & \cellcolor{gray!10}6.5 & 7.0 & \cellcolor{gray!10}1.5 & 7.5 & \cellcolor{gray!10}1.5 & 7.0 & \cellcolor{gray!10}5.0 & 7.5 & \cellcolor{gray!10}1.5 & 7.5 & \cellcolor{gray!10}1.5 & 7.5 & \cellcolor{gray!10}3.0 & 7.0 & \cellcolor{gray!10}2.0 & 7.5 & \cellcolor{gray!10}1.5 \\ \cmidrule(lr){2-21}
  ~ & \multirow{2}{*}{Template 2} & Easy & 9.5 & \cellcolor{gray!10}12.0 & 11.0 & \cellcolor{gray!10}0.5 & 10.5 & \cellcolor{gray!10}0.0 & 9.0 & \cellcolor{gray!10}7.0 & 10.5 & \cellcolor{gray!10}0.0 & 10.5 & \cellcolor{gray!10}0.0 & 9.5 & \cellcolor{gray!10}3.5 & 10.5 & \cellcolor{gray!10}0.5 & 10.5 & \cellcolor{gray!10}0.0 \\
  ~ &  ~ & Hard & 7.5 & \cellcolor{gray!10}7.5 & 7.0 & \cellcolor{gray!10}1.5 & 7.5 & \cellcolor{gray!10}1.5 & 7.0 & \cellcolor{gray!10}5.0 & 7.5 & \cellcolor{gray!10}2.0 & 7.5 & \cellcolor{gray!10}1.5 & 7.5 & \cellcolor{gray!10}3.0 & 7.0 & \cellcolor{gray!10}2.0 & 7.5 & \cellcolor{gray!10}1.5 \\ \cmidrule(lr){2-21}
  ~ & \multirow{2}{*}{Template 3} & Easy & 12.5 & \cellcolor{gray!10}12.0 & 14.5 & \cellcolor{gray!10}0.5 & 14.0 & \cellcolor{gray!10}0.0 & 11.5 & \cellcolor{gray!10}6.5 & 13.0 & \cellcolor{gray!10}0.0 & 12.5 & \cellcolor{gray!10}0.0 & 12.5 & \cellcolor{gray!10}3.0 & 13.0 & \cellcolor{gray!10}0.5 & 13.5 & \cellcolor{gray!10}0.0 \\
  ~ &  ~ & Hard & 7.5 & \cellcolor{gray!10}7.5 & 7.0 & \cellcolor{gray!10}1.5 & 7.5 & \cellcolor{gray!10}1.5 & 7.0 & \cellcolor{gray!10}5.0 & 7.5 & \cellcolor{gray!10}2.0 & 7.5 & \cellcolor{gray!10}1.5 & 7.5 & \cellcolor{gray!10}3.0 & 7.0 & \cellcolor{gray!10}2.0 & 7.5 & \cellcolor{gray!10}1.5 \\ \midrule
 \multirow{6}{*}{\textbf{Deepseek-vl2-small}} & \multirow{2}{*}{Template 1} & Easy & 9.5 & \cellcolor{gray!10}6.5 & 10.5 & \cellcolor{gray!10}0.0 & 10.5 & \cellcolor{gray!10}0.0 & 9.5 & \cellcolor{gray!10}4.0 & 10.0 & \cellcolor{gray!10}0.0 & 10.0 & \cellcolor{gray!10}0.0 & 10.0 & \cellcolor{gray!10}3.0 & 10.5 & \cellcolor{gray!10}0.0 & 10.5 & \cellcolor{gray!10}0.0 \\
  ~ &  ~ & Hard & 7.0 & \cellcolor{gray!10}6.0 & 7.5 & \cellcolor{gray!10}2.0 & 7.5 & \cellcolor{gray!10}2.0 & 7.5 & \cellcolor{gray!10}5.0 & 7.0 & \cellcolor{gray!10}1.5 & 7.5 & \cellcolor{gray!10}2.0 & 7.5 & \cellcolor{gray!10}3.0 & 7.0 & \cellcolor{gray!10}1.5 & 7.5 & \cellcolor{gray!10}1.5 \\ \cmidrule(lr){2-21}
  ~ & \multirow{2}{*}{Template 2} & Easy & 9.5 & \cellcolor{gray!10}7.5 & 10.5 & \cellcolor{gray!10}0.0 & 10.5 & \cellcolor{gray!10}0.0 & 9.5 & \cellcolor{gray!10}4.5 & 10.0 & \cellcolor{gray!10}0.5 & 10.5 & \cellcolor{gray!10}0.0 & 10.0 & \cellcolor{gray!10}3.5 & 10.5 & \cellcolor{gray!10}0.5 & 10.5 & \cellcolor{gray!10}0.0 \\
  ~ &  ~ & Hard & 7.0 & \cellcolor{gray!10}6.0 & 7.5 & \cellcolor{gray!10}2.0 & 7.5 & \cellcolor{gray!10}2.0 & 7.5 & \cellcolor{gray!10}4.5 & 7.0 & \cellcolor{gray!10}2.0 & 7.5 & \cellcolor{gray!10}2.0 & 7.5 & \cellcolor{gray!10}3.0 & 7.0 & \cellcolor{gray!10}1.5 & 7.5 & \cellcolor{gray!10}1.5 \\ \cmidrule(lr){2-21}
  ~ & \multirow{2}{*}{Template 3} & Easy & 13.5 & \cellcolor{gray!10}6.5 & 15.5 & \cellcolor{gray!10}0.0 & 14.5 & \cellcolor{gray!10}0.0 & 13.5 & \cellcolor{gray!10}4.5 & 14.5 & \cellcolor{gray!10}0.0 & 15.0 & \cellcolor{gray!10}0.0 & 14.0 & \cellcolor{gray!10}3.0 & 14.5 & \cellcolor{gray!10}0.0 & 15.0 & \cellcolor{gray!10}0.0 \\
  ~ &  ~ & Hard & 7.0 & \cellcolor{gray!10}6.0 & 7.5 & \cellcolor{gray!10}2.0 & 7.5 & \cellcolor{gray!10}2.0 & 7.5 & \cellcolor{gray!10}4.5 & 7.0 & \cellcolor{gray!10}2.0 & 7.5 & \cellcolor{gray!10}2.0 & 7.5 & \cellcolor{gray!10}3.0 & 7.0 & \cellcolor{gray!10}1.5 & 7.5 & \cellcolor{gray!10}1.5 \\ \midrule
 \multirow{6}{*}{\textbf{Kimi-VL-A3B-Instruct}} & \multirow{2}{*}{Template 1} & Easy & 8.0 & \cellcolor{gray!10}7.0 & 10.0 & \cellcolor{gray!10}1.0 & 10.0 & \cellcolor{gray!10}0.0 & 5.5 & \cellcolor{gray!10}3.0 & 6.0 & \cellcolor{gray!10}0.5 & 8.0 & \cellcolor{gray!10}0.0 & 9.0 & \cellcolor{gray!10}3.5 & 9.0 & \cellcolor{gray!10}0.0 & 9.5 & \cellcolor{gray!10}0.0 \\
  ~ &  ~ & Hard & 7.0 & \cellcolor{gray!10}7.0 & 6.0 & \cellcolor{gray!10}1.0 & 7.0 & \cellcolor{gray!10}1.5 & 5.0 & \cellcolor{gray!10}3.5 & 4.5 & \cellcolor{gray!10}2.0 & 5.0 & \cellcolor{gray!10}1.0 & 7.0 & \cellcolor{gray!10}1.5 & 6.5 & \cellcolor{gray!10}2.0 & 7.0 & \cellcolor{gray!10}1.5 \\ \cmidrule(lr){2-21}
  ~ & \multirow{2}{*}{Template 2} & Easy & 2.0 & \cellcolor{gray!10}5.5 & 2.5 & \cellcolor{gray!10}0.5 & 2.0 & \cellcolor{gray!10}0.0 & 0.5 & \cellcolor{gray!10}1.0 & 1.0 & \cellcolor{gray!10}0.5 & 1.5 & \cellcolor{gray!10}0.0 & 1.0 & \cellcolor{gray!10}1.0 & 1.0 & \cellcolor{gray!10}0.0 & 1.5 & \cellcolor{gray!10}0.5 \\
  ~ &  ~ & Hard & 0.0 & \cellcolor{gray!10}3.0 & 0.5 & \cellcolor{gray!10}0.0 & 1.0 & \cellcolor{gray!10}0.0 & 0.5 & \cellcolor{gray!10}1.0 & 0.0 & \cellcolor{gray!10}0.5 & 1.5 & \cellcolor{gray!10}0.0 & 0.5 & \cellcolor{gray!10}1.0 & 0.0 & \cellcolor{gray!10}0.5 & 0.0 & \cellcolor{gray!10}0.0 \\ \cmidrule(lr){2-21}
  ~ & \multirow{2}{*}{Template 3} & Easy & 5.5 & \cellcolor{gray!10}3.5 & 8.5 & \cellcolor{gray!10}0.0 & 7.5 & \cellcolor{gray!10}0.0 & 4.5 & \cellcolor{gray!10}1.0 & 5.5 & \cellcolor{gray!10}0.0 & 6.0 & \cellcolor{gray!10}0.0 & 6.0 & \cellcolor{gray!10}0.5 & 5.5 & \cellcolor{gray!10}0.0 & 4.5 & \cellcolor{gray!10}0.0 \\
  ~ &  ~ & Hard & 0.5 & \cellcolor{gray!10}2.0 & 0.5 & \cellcolor{gray!10}0.0 & 0.5 & \cellcolor{gray!10}0.0 & 1.0 & \cellcolor{gray!10}0.5 & 0.0 & \cellcolor{gray!10}0.0 & 0.5 & \cellcolor{gray!10}0.0 & 0.0 & \cellcolor{gray!10}0.5 & 1.0 & \cellcolor{gray!10}0.5 & 0.0 & \cellcolor{gray!10}0.0 \\ \midrule
 \multirow{6}{*}{\textbf{Gemma-3-12b-it}} & \multirow{2}{*}{Template 1} & Easy & 9.5 & \cellcolor{gray!10}9.0 & 11.5 & \cellcolor{gray!10}0.0 & 7.5 & \cellcolor{gray!10}0.5 & 10.0 & \cellcolor{gray!10}7.0 & 8.0 & \cellcolor{gray!10}0.0 & 8.0 & \cellcolor{gray!10}0.0 & 7.0 & \cellcolor{gray!10}4.5 & 2.5 & \cellcolor{gray!10}0.0 & 2.0 & \cellcolor{gray!10}0.0 \\
  ~ &  ~ & Hard & 5.5 & \cellcolor{gray!10}22.5 & 7.0 & \cellcolor{gray!10}3.5 & 5.5 & \cellcolor{gray!10}2.0 & 7.0 & \cellcolor{gray!10}22.0 & 7.0 & \cellcolor{gray!10}2.0 & 4.0 & \cellcolor{gray!10}0.5 & 5.0 & \cellcolor{gray!10}9.0 & 1.5 & \cellcolor{gray!10}1.0 & 0.5 & \cellcolor{gray!10}0.5 \\ \cmidrule(lr){2-21}
  ~ & \multirow{2}{*}{Template 2} & Easy & 9.5 & \cellcolor{gray!10}9.0 & 11.5 & \cellcolor{gray!10}0.5 & 8.0 & \cellcolor{gray!10}0.5 & 9.5 & \cellcolor{gray!10}7.0 & 8.0 & \cellcolor{gray!10}0.0 & 8.5 & \cellcolor{gray!10}0.0 & 7.0 & \cellcolor{gray!10}4.5 & 3.0 & \cellcolor{gray!10}0.0 & 2.0 & \cellcolor{gray!10}0.0 \\
  ~ &  ~ & Hard & 5.5 & \cellcolor{gray!10}22.5 & 7.0 & \cellcolor{gray!10}3.5 & 5.5 & \cellcolor{gray!10}2.0 & 7.0 & \cellcolor{gray!10}22.0 & 7.0 & \cellcolor{gray!10}2.0 & 4.0 & \cellcolor{gray!10}0.5 & 5.0 & \cellcolor{gray!10}9.0 & 1.5 & \cellcolor{gray!10}1.0 & 0.5 & \cellcolor{gray!10}0.5 \\ \cmidrule(lr){2-21}
  ~ & \multirow{2}{*}{Template 3} & Easy & 11.5 & \cellcolor{gray!10}9.0 & 14.0 & \cellcolor{gray!10}0.0 & 10.5 & \cellcolor{gray!10}0.5 & 13.5 & \cellcolor{gray!10}7.0 & 10.0 & \cellcolor{gray!10}0.0 & 11.0 & \cellcolor{gray!10}0.0 & 9.5 & \cellcolor{gray!10}4.5 & 5.0 & \cellcolor{gray!10}0.0 & 2.5 & \cellcolor{gray!10}0.0 \\
  ~ &  ~ & Hard & 5.5 & \cellcolor{gray!10}22.5 & 7.0 & \cellcolor{gray!10}3.5 & 5.5 & \cellcolor{gray!10}2.0 & 7.0 & \cellcolor{gray!10}22.0 & 7.0 & \cellcolor{gray!10}2.0 & 4.0 & \cellcolor{gray!10}0.5 & 5.0 & \cellcolor{gray!10}9.0 & 1.5 & \cellcolor{gray!10}1.0 & 0.5 & \cellcolor{gray!10}0.5 \\ \midrule
 \multirow{6}{*}{\textbf{Qwen3-VL-8B-Instruct}} & \multirow{2}{*}{Template 1} & Easy & 9.5 & \cellcolor{gray!10}2.5 & 6.0 & \cellcolor{gray!10}0.0 & 2.0 & \cellcolor{gray!10}0.0 & 10.0 & \cellcolor{gray!10}1.5 & 7.5 & \cellcolor{gray!10}0.0 & 3.0 & \cellcolor{gray!10}0.0 & 9.5 & \cellcolor{gray!10}0.5 & 6.0 & \cellcolor{gray!10}0.5 & 2.0 & \cellcolor{gray!10}0.0 \\
  ~ &  ~ & Hard & 6.5 & \cellcolor{gray!10}8.0 & 4.5 & \cellcolor{gray!10}1.0 & 0.5 & \cellcolor{gray!10}0.5 & 6.0 & \cellcolor{gray!10}3.5 & 3.0 & \cellcolor{gray!10}1.0 & 1.0 & \cellcolor{gray!10}0.5 & 6.5 & \cellcolor{gray!10}2.5 & 2.5 & \cellcolor{gray!10}1.0 & 0.5 & \cellcolor{gray!10}0.0 \\ \cmidrule(lr){2-21}
  ~ & \multirow{2}{*}{Template 2} & Easy & 9.5 & \cellcolor{gray!10}2.0 & 6.5 & \cellcolor{gray!10}0.0 & 2.0 & \cellcolor{gray!10}0.0 & 10.0 & \cellcolor{gray!10}1.5 & 6.5 & \cellcolor{gray!10}0.0 & 3.5 & \cellcolor{gray!10}0.0 & 10.0 & \cellcolor{gray!10}1.0 & 5.0 & \cellcolor{gray!10}0.5 & 2.5 & \cellcolor{gray!10}0.0 \\
  ~ &  ~ & Hard & 6.0 & \cellcolor{gray!10}8.5 & 4.5 & \cellcolor{gray!10}1.0 & 1.0 & \cellcolor{gray!10}0.5 & 6.0 & \cellcolor{gray!10}3.5 & 3.0 & \cellcolor{gray!10}0.0 & 0.5 & \cellcolor{gray!10}0.5 & 6.5 & \cellcolor{gray!10}2.5 & 2.5 & \cellcolor{gray!10}1.0 & 1.0 & \cellcolor{gray!10}0.0 \\ \cmidrule(lr){2-21}
  ~ & \multirow{2}{*}{Template 3} & Easy & 12.5 & \cellcolor{gray!10}2.0 & 8.0 & \cellcolor{gray!10}0.5 & 3.5 & \cellcolor{gray!10}0.0 & 11.5 & \cellcolor{gray!10}1.5 & 8.0 & \cellcolor{gray!10}0.0 & 6.0 & \cellcolor{gray!10}0.0 & 12.5 & \cellcolor{gray!10}0.5 & 8.5 & \cellcolor{gray!10}0.5 & 2.5 & \cellcolor{gray!10}0.0 \\
  ~ &  ~ & Hard & 6.5 & \cellcolor{gray!10}8.0 & 4.5 & \cellcolor{gray!10}1.0 & 1.0 & \cellcolor{gray!10}0.5 & 6.0 & \cellcolor{gray!10}3.5 & 2.5 & \cellcolor{gray!10}0.5 & 1.0 & \cellcolor{gray!10}0.5 & 6.5 & \cellcolor{gray!10}2.5 & 2.5 & \cellcolor{gray!10}1.0 & 1.0 & \cellcolor{gray!10}0.5 \\ \midrule
 \multirow{6}{*}{\textbf{InternVL3\_5-8B}} & \multirow{2}{*}{Template 1} & Easy & 9.5 & \cellcolor{gray!10}5.5 & 9.5 & \cellcolor{gray!10}0.5 & 9.0 & \cellcolor{gray!10}0.0 & 10.0 & \cellcolor{gray!10}4.0 & 10.0 & \cellcolor{gray!10}0.5 & 7.5 & \cellcolor{gray!10}0.0 & 9.5 & \cellcolor{gray!10}4.0 & 9.5 & \cellcolor{gray!10}1.0 & 7.5 & \cellcolor{gray!10}0.0 \\
  ~ &  ~ & Hard & 7.0 & \cellcolor{gray!10}6.5 & 7.0 & \cellcolor{gray!10}1.5 & 5.0 & \cellcolor{gray!10}1.5 & 7.0 & \cellcolor{gray!10}4.0 & 5.5 & \cellcolor{gray!10}1.5 & 5.0 & \cellcolor{gray!10}0.5 & 7.5 & \cellcolor{gray!10}3.0 & 6.5 & \cellcolor{gray!10}1.5 & 6.5 & \cellcolor{gray!10}1.5 \\ \cmidrule(lr){2-21}
  ~ & \multirow{2}{*}{Template 2} & Easy & 9.5 & \cellcolor{gray!10}4.0 & 10.0 & \cellcolor{gray!10}0.0 & 9.0 & \cellcolor{gray!10}0.0 & 9.5 & \cellcolor{gray!10}2.0 & 10.5 & \cellcolor{gray!10}0.0 & 10.0 & \cellcolor{gray!10}0.0 & 9.5 & \cellcolor{gray!10}3.0 & 9.5 & \cellcolor{gray!10}1.0 & 9.0 & \cellcolor{gray!10}0.0 \\
  ~ &  ~ & Hard & 7.0 & \cellcolor{gray!10}6.0 & 7.0 & \cellcolor{gray!10}1.5 & 6.0 & \cellcolor{gray!10}1.5 & 7.0 & \cellcolor{gray!10}3.0 & 5.5 & \cellcolor{gray!10}1.5 & 6.0 & \cellcolor{gray!10}1.0 & 8.0 & \cellcolor{gray!10}2.5 & 7.0 & \cellcolor{gray!10}1.5 & 7.0 & \cellcolor{gray!10}1.5 \\ \cmidrule(lr){2-21}
  ~ & \multirow{2}{*}{Template 3} & Easy & 12.0 & \cellcolor{gray!10}4.0 & 11.5 & \cellcolor{gray!10}0.0 & 10.5 & \cellcolor{gray!10}0.0 & 10.0 & \cellcolor{gray!10}2.5 & 11.5 & \cellcolor{gray!10}0.0 & 10.5 & \cellcolor{gray!10}0.0 & 11.5 & \cellcolor{gray!10}3.5 & 10.0 & \cellcolor{gray!10}1.0 & 9.0 & \cellcolor{gray!10}0.0 \\
  ~ &  ~ & Hard & 7.0 & \cellcolor{gray!10}6.0 & 7.0 & \cellcolor{gray!10}1.5 & 6.0 & \cellcolor{gray!10}1.5 & 7.0 & \cellcolor{gray!10}3.0 & 5.5 & \cellcolor{gray!10}1.5 & 6.0 & \cellcolor{gray!10}1.0 & 8.0 & \cellcolor{gray!10}2.5 & 7.0 & \cellcolor{gray!10}1.5 & 7.0 & \cellcolor{gray!10}1.5 \\ \midrule
 \multirow{6}{*}{\textbf{Qianfan-VL-8B}} & \multirow{2}{*}{Template 1} & Easy & 7.5 & \cellcolor{gray!10}13.5 & 8.5 & \cellcolor{gray!10}1.0 & 8.5 & \cellcolor{gray!10}0.0 & 7.5 & \cellcolor{gray!10}7.5 & 9.5 & \cellcolor{gray!10}0.0 & 8.0 & \cellcolor{gray!10}0.0 & 7.5 & \cellcolor{gray!10}10.5 & 7.5 & \cellcolor{gray!10}1.5 & 6.0 & \cellcolor{gray!10}0.0 \\
  ~ &  ~ & Hard & 7.0 & \cellcolor{gray!10}13.5 & 6.0 & \cellcolor{gray!10}1.5 & 5.0 & \cellcolor{gray!10}1.5 & 7.0 & \cellcolor{gray!10}6.5 & 7.5 & \cellcolor{gray!10}1.5 & 7.0 & \cellcolor{gray!10}1.5 & 7.0 & \cellcolor{gray!10}5.0 & 5.5 & \cellcolor{gray!10}1.5 & 5.5 & \cellcolor{gray!10}1.5 \\ \cmidrule(lr){2-21}
  ~ & \multirow{2}{*}{Template 2} & Easy & 7.5 & \cellcolor{gray!10}14.0 & 8.0 & \cellcolor{gray!10}2.0 & 8.5 & \cellcolor{gray!10}0.0 & 7.0 & \cellcolor{gray!10}7.5 & 9.5 & \cellcolor{gray!10}0.0 & 9.0 & \cellcolor{gray!10}0.0 & 7.5 & \cellcolor{gray!10}10.5 & 7.5 & \cellcolor{gray!10}1.5 & 6.0 & \cellcolor{gray!10}0.0 \\
  ~ &  ~ & Hard & 7.0 & \cellcolor{gray!10}14.5 & 6.0 & \cellcolor{gray!10}1.5 & 4.5 & \cellcolor{gray!10}1.5 & 6.5 & \cellcolor{gray!10}6.5 & 7.5 & \cellcolor{gray!10}1.5 & 7.0 & \cellcolor{gray!10}1.5 & 7.0 & \cellcolor{gray!10}5.0 & 5.5 & \cellcolor{gray!10}1.5 & 5.5 & \cellcolor{gray!10}1.5 \\ \cmidrule(lr){2-21}
  ~ & \multirow{2}{*}{Template 3} & Easy & 10.0 & \cellcolor{gray!10}13.0 & 11.5 & \cellcolor{gray!10}1.5 & 12.5 & \cellcolor{gray!10}0.0 & 9.0 & \cellcolor{gray!10}7.0 & 13.5 & \cellcolor{gray!10}0.0 & 13.5 & \cellcolor{gray!10}0.0 & 9.5 & \cellcolor{gray!10}10.5 & 9.0 & \cellcolor{gray!10}1.5 & 8.0 & \cellcolor{gray!10}0.0 \\
  ~ &  ~ & Hard & 7.0 & \cellcolor{gray!10}14.5 & 6.0 & \cellcolor{gray!10}1.5 & 4.5 & \cellcolor{gray!10}1.5 & 6.5 & \cellcolor{gray!10}6.5 & 7.5 & \cellcolor{gray!10}1.5 & 7.0 & \cellcolor{gray!10}1.5 & 7.0 & \cellcolor{gray!10}5.0 & 5.5 & \cellcolor{gray!10}1.5 & 5.5 & \cellcolor{gray!10}1.5 \\ \midrule
 \multirow{6}{*}{\textbf{Gemma-3-4b-it}} & \multirow{2}{*}{Template 1} & Easy & 8.5 & \cellcolor{gray!10}17.0 & 9.0 & \cellcolor{gray!10}2.0 & 9.0 & \cellcolor{gray!10}0.5 & 8.5 & \cellcolor{gray!10}13.0 & 10.5 & \cellcolor{gray!10}0.0 & 10.5 & \cellcolor{gray!10}0.0 & 9.5 & \cellcolor{gray!10}4.5 & 10.0 & \cellcolor{gray!10}1.0 & 10.0 & \cellcolor{gray!10}0.0 \\
  ~ &  ~ & Hard & 6.5 & \cellcolor{gray!10}15.0 & 7.5 & \cellcolor{gray!10}2.5 & 8.0 & \cellcolor{gray!10}2.0 & 7.0 & \cellcolor{gray!10}10.5 & 7.5 & \cellcolor{gray!10}2.0 & 7.0 & \cellcolor{gray!10}2.0 & 7.0 & \cellcolor{gray!10}6.0 & 7.0 & \cellcolor{gray!10}2.0 & 6.5 & \cellcolor{gray!10}2.0 \\ \cmidrule(lr){2-21}
  ~ & \multirow{2}{*}{Template 2} & Easy & 8.5 & \cellcolor{gray!10}18.0 & 9.0 & \cellcolor{gray!10}2.0 & 9.0 & \cellcolor{gray!10}0.5 & 8.5 & \cellcolor{gray!10}13.5 & 10.5 & \cellcolor{gray!10}0.0 & 11.0 & \cellcolor{gray!10}0.0 & 9.5 & \cellcolor{gray!10}4.5 & 10.0 & \cellcolor{gray!10}1.0 & 10.0 & \cellcolor{gray!10}0.0 \\
  ~ &  ~ & Hard & 6.5 & \cellcolor{gray!10}15.0 & 7.5 & \cellcolor{gray!10}2.5 & 8.0 & \cellcolor{gray!10}2.0 & 7.0 & \cellcolor{gray!10}10.5 & 7.5 & \cellcolor{gray!10}2.0 & 7.0 & \cellcolor{gray!10}2.0 & 7.0 & \cellcolor{gray!10}6.0 & 7.0 & \cellcolor{gray!10}2.0 & 6.5 & \cellcolor{gray!10}2.0 \\ \cmidrule(lr){2-21}
  ~ & \multirow{2}{*}{Template 3} & Easy & 8.5 & \cellcolor{gray!10}18.0 & 9.0 & \cellcolor{gray!10}2.0 & 10.0 & \cellcolor{gray!10}0.5 & 8.5 & \cellcolor{gray!10}13.0 & 10.5 & \cellcolor{gray!10}0.0 & 11.0 & \cellcolor{gray!10}0.0 & 9.5 & \cellcolor{gray!10}4.5 & 10.0 & \cellcolor{gray!10}1.0 & 10.5 & \cellcolor{gray!10}0.0 \\
  ~ &  ~ & Hard & 6.5 & \cellcolor{gray!10}15.0 & 7.5 & \cellcolor{gray!10}2.5 & 8.0 & \cellcolor{gray!10}2.0 & 7.0 & \cellcolor{gray!10}10.5 & 7.5 & \cellcolor{gray!10}2.0 & 7.0 & \cellcolor{gray!10}2.0 & 7.0 & \cellcolor{gray!10}6.0 & 7.0 & \cellcolor{gray!10}2.0 & 6.5 & \cellcolor{gray!10}2.0 \\ \midrule
 \multirow{6}{*}{\textbf{Qianfan-VL-3B}} & \multirow{2}{*}{Template 1} & Easy & 9.0 & \cellcolor{gray!10}3.5 & 9.0 & \cellcolor{gray!10}0.0 & 9.0 & \cellcolor{gray!10}0.0 & 9.0 & \cellcolor{gray!10}1.0 & 9.0 & \cellcolor{gray!10}0.0 & 9.0 & \cellcolor{gray!10}0.0 & 9.0 & \cellcolor{gray!10}0.5 & 10.0 & \cellcolor{gray!10}0.0 & 9.0 & \cellcolor{gray!10}0.0 \\
  ~ &  ~ & Hard & 8.0 & \cellcolor{gray!10}5.0 & 7.5 & \cellcolor{gray!10}1.5 & 7.5 & \cellcolor{gray!10}1.5 & 7.0 & \cellcolor{gray!10}4.0 & 7.5 & \cellcolor{gray!10}1.5 & 7.0 & \cellcolor{gray!10}1.5 & 7.0 & \cellcolor{gray!10}2.5 & 6.0 & \cellcolor{gray!10}1.5 & 7.0 & \cellcolor{gray!10}1.5 \\ \cmidrule(lr){2-21}
  ~ & \multirow{2}{*}{Template 2} & Easy & 9.0 & \cellcolor{gray!10}5.0 & 9.0 & \cellcolor{gray!10}0.0 & 9.0 & \cellcolor{gray!10}0.0 & 9.0 & \cellcolor{gray!10}1.0 & 9.0 & \cellcolor{gray!10}0.0 & 9.0 & \cellcolor{gray!10}0.0 & 9.0 & \cellcolor{gray!10}0.5 & 9.5 & \cellcolor{gray!10}0.0 & 9.0 & \cellcolor{gray!10}0.0 \\
  ~ &  ~ & Hard & 8.0 & \cellcolor{gray!10}5.5 & 7.5 & \cellcolor{gray!10}1.5 & 7.5 & \cellcolor{gray!10}1.5 & 7.0 & \cellcolor{gray!10}4.0 & 7.5 & \cellcolor{gray!10}1.5 & 7.0 & \cellcolor{gray!10}1.5 & 7.0 & \cellcolor{gray!10}2.5 & 6.5 & \cellcolor{gray!10}1.5 & 7.0 & \cellcolor{gray!10}1.5 \\ \cmidrule(lr){2-21}
  ~ & \multirow{2}{*}{Template 3} & Easy & 12.5 & \cellcolor{gray!10}4.0 & 12.5 & \cellcolor{gray!10}0.0 & 13.0 & \cellcolor{gray!10}0.0 & 12.5 & \cellcolor{gray!10}1.0 & 13.0 & \cellcolor{gray!10}0.0 & 13.0 & \cellcolor{gray!10}0.0 & 12.0 & \cellcolor{gray!10}0.5 & 13.5 & \cellcolor{gray!10}0.0 & 12.5 & \cellcolor{gray!10}0.0 \\
  ~ &  ~ & Hard & 8.0 & \cellcolor{gray!10}5.5 & 7.5 & \cellcolor{gray!10}1.5 & 7.5 & \cellcolor{gray!10}1.5 & 7.0 & \cellcolor{gray!10}4.0 & 7.5 & \cellcolor{gray!10}1.5 & 7.0 & \cellcolor{gray!10}1.5 & 7.0 & \cellcolor{gray!10}2.5 & 6.5 & \cellcolor{gray!10}1.5 & 7.0 & \cellcolor{gray!10}1.5 \\
 \bottomrule
\end{tabular}
}%
\end{table*}

\section{Mitigation: Visual-Semantic Decoupling (Full Details)}
\label{appendix:mitigation}

The WCIT benchmark reveals that current MLLMs are highly susceptible to Embedded Stroop interference, with hallucination rates exceeding 75\% for the worst-performing models. A benchmark-only contribution, however, leaves open the critical question of whether this vulnerability can be remedied. In this section, we propose the \textbf{Visual-Semantic Decoupling (VSD)} framework, a suite of mitigation strategies that spans inference-time intervention, lightweight fine-tuning, and prompt-based baselines.

\subsection{Overview and Motivation}
\label{sec:mitigation_overview_app}

The core failure mode identified by WCIT is that MLLMs process \emph{semantic content} (the meaning of embedded text) in preference over \emph{visual attributes} (the physical ink color of that text). This mirrors the classical Stroop effect in human cognition, where reading---an automatized process---interferes with color naming. For MLLMs, the analogous ``automatized process'' is the strong prior that textual tokens carry semantic meaning, which the vision-language alignment layers amplify at the expense of low-level visual features.

An effective mitigation must therefore satisfy two desiderata: (1)~\textbf{Semantic suppression}---reduce the influence of textual meaning on the final answer, and (2)~\textbf{Visual preservation}---maintain access to accurate color information. We propose two variants of VSD that achieve this balance through complementary mechanisms:

\begin{itemize}
 \item \textbf{Training-Free VSD} (\S\ref{sec:tf_vsd_app}): A dual-view inference pipeline applicable to any MLLM, including proprietary APIs, without modification to model weights.
 \item \textbf{Stroop-Contrastive LoRA} (\S\ref{sec:sc_lora_app}): A lightweight parameter-efficient fine-tuning approach for open-source models that teaches the model to disentangle visual from semantic features via a contrastive objective.
\end{itemize}

We additionally evaluate three prompt-based strategies (\S\ref{sec:prompt_mitigation_app}) as baselines, and describe the full experimental design in \S\ref{sec:mitigation_experiment_app}.

\subsection{Training-Free Visual-Semantic Decoupling}
\label{sec:tf_vsd_app}

\subsubsection{Pipeline}

The training-free VSD pipeline operates entirely at inference time and requires no access to model internals. Given an input image $\mathbf{x}$ containing a Stroop stimulus, VSD proceeds in four stages:

\textbf{Stage 1: Text Region Detection.}
An OCR module $\mathcal{O}$ localizes all text regions in $\mathbf{x}$, producing a set of bounding boxes $\mathcal{B} = \{b_1, \ldots, b_K\}$ along with their recognized content $\{w_1, \ldots, w_K\}$.

\textbf{Stage 2: Semantic-Suppressed View Construction.}
For each bounding box $b_k$, we replace the glyph pixels with a uniform color patch sampled from the average pixel value within $b_k$. This yields a \emph{semantic-suppressed} image $\tilde{\mathbf{x}}$ in which the spatial extent, layout, and---crucially---the \emph{ink color} of every text region are preserved, but the semantic content (recognizable characters) is erased. Formally, let $\mathbf{x}[b_k]$ denote the pixel sub-array within bounding box $b_k$ and let $\bar{c}_k = \frac{1}{|b_k|}\sum_{p \in b_k} \mathbf{x}[p]$ be the mean color of that region. Then:
\begin{equation}
 \tilde{\mathbf{x}}[p] =
 \begin{cases}
 \bar{c}_k & \text{if } p \in b_k \text{ for some } k, \\
 \mathbf{x}[p] & \text{otherwise}.
 \end{cases}
 \label{eq:semantic_suppress}
\end{equation}
Since the WCIT stimulus is monochrome text rendered in a single color $c_{vis}$, the replacement patch closely approximates $c_{vis}$, effectively turning each word into a colored rectangle that preserves the target hue.

\textbf{Stage 3: Dual-View Inference.}
The MLLM $\mathcal{M}$ processes both views independently:
\begin{align}
 y_{\text{orig}} &= \mathcal{M}(\mathbf{x},\, q), \\
 y_{\text{supp}} &= \mathcal{M}(\tilde{\mathbf{x}},\, q),
\end{align}
where $q$ is the task query (e.g., ``What color is the text?'').

\textbf{Stage 4: Consistency Check and Color Calibration.}
We compare the two predictions via a color calibration module. Because MLLM outputs are free-form text rather than structured color labels, we first map each prediction to a canonical color name using a nearest-neighbor lookup in CIE $L^*a^*b^*$ space with the CIEDE2000 distance $\Delta E_{00}$ (cf.\ Equation~\ref{eq:ciede2000_app}). Let $\psi(\cdot)$ denote this mapping:
\begin{equation}
 \psi(y) = \arg\min_{c \in \mathcal{C}} \Delta E_{00}\!\bigl(\text{rgb}(y),\, \text{rgb}(c)\bigr),
\end{equation}
where $\mathcal{C}$ is the curated WCIT color palette. The final output is determined by:
\begin{equation}
 y^* =
 \begin{cases}
 y_{\text{supp}} & \text{if } \Delta E_{00}\!\bigl(\psi(y_{\text{orig}}),\, \psi(y_{\text{supp}})\bigr) > \tau, \\
 y_{\text{orig}} & \text{otherwise},
 \end{cases}
 \label{eq:consistency}
\end{equation}
where $\tau$ is a perceptual distance threshold set to $\tau = 5$ based on the just-noticeable-difference (JND) boundary in CIEDE2000. The intuition is straightforward: if the two views disagree, the semantic-suppressed view---which cannot rely on textual shortcuts---is more trustworthy for color reporting.

\subsubsection{Algorithm}

We summarize the training-free VSD procedure in Algorithm~\ref{alg:tf_vsd}.

\begin{algorithm}[t]
\caption{Training-Free Visual-Semantic Decoupling (TF-VSD)}
\label{alg:tf_vsd}
\begin{algorithmic}[1]
 \STATE {\bfseries Input:} Image $\mathbf{x}$, query $q$, MLLM $\mathcal{M}$, OCR $\mathcal{O}$, threshold $\tau$
 \STATE {\bfseries Output:} Predicted color $y^*$
 \STATE $\mathcal{B}, \{w_k\} \leftarrow \mathcal{O}(\mathbf{x})$ \hfill \COMMENT{Detect text regions}
 \STATE $\tilde{\mathbf{x}} \leftarrow \mathbf{x}$
 \FOR{$b_k \in \mathcal{B}$}
 \STATE $\bar{c}_k \leftarrow \frac{1}{|b_k|}\sum_{p \in b_k} \mathbf{x}[p]$
 \STATE $\tilde{\mathbf{x}}[p] \leftarrow \bar{c}_k$ for all $p \in b_k$
 \ENDFOR
 \STATE $y_{\text{orig}} \leftarrow \mathcal{M}(\mathbf{x},\, q)$ \hfill \COMMENT{Original inference}
 \STATE $y_{\text{supp}} \leftarrow \mathcal{M}(\tilde{\mathbf{x}},\, q)$ \hfill \COMMENT{Suppressed inference}
 \STATE $d \leftarrow \Delta E_{00}\!\bigl(\psi(y_{\text{orig}}),\, \psi(y_{\text{supp}})\bigr)$
 \IF{$d > \tau$}
 \STATE $y^* \leftarrow y_{\text{supp}}$
 \ELSE
 \STATE $y^* \leftarrow y_{\text{orig}}$
 \ENDIF
 \STATE {\bfseries Return} $y^*$
\end{algorithmic}
\end{algorithm}

\subsubsection{Analysis}

The key insight behind training-free VSD is that the Stroop effect in MLLMs is driven by a \emph{semantic shortcut}: the model routes recognized text through its language backbone, which predicts the color name token autoregressively based on semantic content rather than visual features. By replacing glyphs with uniform color patches, we sever this shortcut while preserving the spatial distribution and chromatic properties of the stimulus. The dual-view design provides a self-diagnostic mechanism---when the two views agree, the original inference is already grounded in visual features; when they disagree, the discrepancy itself signals semantic interference.

\subsection{Stroop-Contrastive LoRA}
\label{sec:sc_lora_app}

For open-source MLLMs, we can improve visual-semantic decoupling more fundamentally by adjusting the model's internal representations. We propose \textbf{Stroop-Contrastive LoRA} (SC-LoRA), a parameter-efficient fine-tuning method that injects a contrastive signal into the vision-language projection layers.

\subsubsection{Training Data Construction}

We construct a Stroop-aware training set $\mathcal{S}$ from the WCIT color palette. Each training sample is a triplet $(\mathbf{x}, c_{vis}, c_{sem})$ where $\mathbf{x}$ is a synthetically generated image. We create two complementary pairing strategies:

\textbf{Same-Word / Different-Color (SW-DC).}
For a fixed semantic word $w$ (e.g., ``RED''), we render $\mathbf{x}$ with $w$ displayed in multiple distinct ink colors $\{c_1, c_2, \ldots\}$ drawn from $\mathcal{C}$. The correct answer varies across samples despite identical semantic content, forcing the model to attend to the visual channel.

\textbf{Different-Word / Same-Color (DW-SC).}
For a fixed ink color $c$, we render images with different semantic words $\{w_1, w_2, \ldots\}$ all in color $c$. The correct answer is constant despite varying semantic content, teaching the model that text meaning is irrelevant to color identification.

We generate $N_s$ SW-DC samples and $N_s$ DW-SC samples, yielding $|\mathcal{S}| = 2N_s$ training instances. In our experiments we set $N_s = 2000$.

\subsubsection{LoRA Configuration}

We apply Low-Rank Adaptation (LoRA) to the cross-attention layers that bridge the vision encoder and the language decoder. For each target weight matrix $\mathbf{W}_0 \in \mathbb{R}^{d \times k}$, we learn a low-rank update:
\begin{equation}
 \mathbf{W} = \mathbf{W}_0 + \mathbf{B}\mathbf{A}, \quad \mathbf{B} \in \mathbb{R}^{d \times r},\; \mathbf{A} \in \mathbb{R}^{r \times k},
\end{equation}
where $r \ll \min(d, k)$ is the LoRA rank. We set $r = 8$ and apply LoRA to the query and value projection matrices in the cross-attention layers, yielding approximately 4.7M trainable parameters---less than 0.1\% of the total model size.

\subsubsection{Training Objective}

The SC-LoRA objective combines a supervised fine-tuning loss with a contrastive loss:
\begin{equation}
 \mathcal{L} = \mathcal{L}_{\text{SFT}} + \lambda \, \mathcal{L}_{\text{contrast}},
 \label{eq:total_loss}
\end{equation}
where $\lambda$ is a balancing hyperparameter (set to 0.5 in our experiments).

\textbf{Supervised Color-Answer Loss.}
The standard cross-entropy loss over the color-name tokens:
\begin{equation}
 \mathcal{L}_{\text{SFT}} = -\frac{1}{|\mathcal{S}|}\sum_{i \in \mathcal{S}} \log P(c_{vis}^{(i)} \mid \mathbf{x}_i, q;\, \theta),
 \label{eq:sft_loss}
\end{equation}
where $\theta$ denotes the LoRA parameters.

\textbf{Contrastive Objective.}
Let $\mathbf{h}_i \in \mathbb{R}^d$ be the pooled cross-attention representation for sample $i$, extracted from the last cross-attention layer. For each anchor sample $i$ with visual color $c_{vis}^{(i)}$, we define:
\begin{itemize}
 \item \emph{Positive set} $\mathcal{P}(i) = \{j \in \mathcal{S} : c_{vis}^{(j)} = c_{vis}^{(i)},\, j \neq i\}$: samples sharing the same visual color.
 \item \emph{Negative set} $\mathcal{N}(i) = \{j \in \mathcal{S} : c_{sem}^{(j)} = c_{vis}^{(i)},\, c_{vis}^{(j)} \neq c_{vis}^{(i)}\}$: samples whose \emph{semantic} text matches the anchor's visual color, but whose actual visual color differs.
\end{itemize}
The contrastive loss is the InfoNCE objective over these sets:
\begin{equation}
\small
\begin{aligned}
 \mathcal{L}_{\text{contrast}}
 &=
 -\frac{1}{|\mathcal{S}|}
 \sum_{i \in \mathcal{S}}
 \log
 \frac{
 \sum_{j \in \mathcal{P}(i)}
 \exp\!\bigl(\text{sim}(\mathbf{h}_i,\mathbf{h}_j)/\alpha\bigr)
 }{
 \sum_{j \in \mathcal{P}(i) \cup \mathcal{N}(i)}
 \exp\!\bigl(\text{sim}(\mathbf{h}_i,\mathbf{h}_j)/\alpha\bigr)
 } .
\end{aligned}
\label{eq:contrast_loss}
\end{equation}

where $\text{sim}(\mathbf{u}, \mathbf{v}) = \mathbf{u}^\top \mathbf{v} / (\|\mathbf{u}\|\|\mathbf{v}\|)$ is cosine similarity and $\alpha = 0.07$ is the temperature.

The design of $\mathcal{N}(i)$ is deliberate: negative pairs are chosen so that the semantic content of one sample matches the visual color of the other. This directly penalizes the model for collapsing representations based on semantic similarity rather than visual attributes. Over training, the cross-attention representations of two images with the same ink color but different embedded words converge, while representations sharing semantic content but different ink colors are pushed apart.

\subsection{Prompt-Based Mitigations (Baselines)}
\label{sec:prompt_mitigation_app}

We evaluate three prompt-only strategies that require no modification to model weights or inference pipeline:

\textbf{P1: Explicit Instruction.}
We prepend a directive to the query: \textit{``Ignore the text content and report only the ink color of the text in the image.''} This tests whether the model can follow a direct instruction to override its default processing.

\textbf{P2: Chain-of-Thought (CoT).}
We append: \textit{``First, identify the color of the pixels that form the text. Then, separately identify what word the text spells. Finally, report the pixel color as your answer.''} This encourages step-by-step reasoning that may isolate visual processing from semantic interpretation.

\textbf{P3: Adversarial-Aware Prompt.}
We prepend: \textit{``Watch for Stroop tricks: the text may spell a color name that differs from its actual ink color. Focus only on the physical appearance.''} This explicitly names the interference mechanism to test whether awareness of the Stroop paradigm enables self-correction.

\subsection{Experiment Design}
\label{sec:mitigation_experiment_app}

\subsubsection{Settings}

We compare four conditions in a controlled evaluation on the full WCIT Embedded Stroop dataset:

\begin{enumerate}
 \item \textbf{Baseline}: Standard inference with the original query, no mitigation applied.
 \item \textbf{Prompt-only}: Each of the three prompt strategies (P1, P2, P3) applied individually.
 \item \textbf{Training-Free VSD}: The dual-view pipeline of \S\ref{sec:tf_vsd_app} applied to both proprietary and open-source models.
 \item \textbf{SC-LoRA}: Stroop-Contrastive LoRA (\S\ref{sec:sc_lora_app}) applied to open-source models only.
\end{enumerate}

\subsubsection{Models}

Training-Free VSD is evaluated on all models from the original WCIT evaluation: GPT-5, GPT-4o, Gemini 2.5, Gemini 3, Qwen3-VL-Plus, GLM-4.5V (proprietary), and Gemma-3, InternVL, Qwen3-VL, DeepSeek-VL2 (open-source). SC-LoRA is applied to the open-source subset with parameter-efficient fine-tuning.

\subsubsection{Evaluation Metrics}

In addition to the Accuracy (Acc), Stroop Hallucination Rate (SHR), and Other Error Rate (OER) metrics defined in \S\ref{sec:metrics}, we introduce two new metrics tailored to mitigation evaluation:

\textbf{Stroop Hallucination Reduction (SHRR).}
The proportional decrease in hallucination rate relative to the baseline:
\begin{equation}
 \text{SHRR} = \frac{\text{SHR}_{\text{base}} - \text{SHR}_{\text{mitig}}}{\text{SHR}_{\text{base}}} \times 100\%.
 \label{eq:shrr}
\end{equation}
SHRR $= 100\%$ indicates complete elimination of Stroop-induced hallucinations.

\textbf{Answer-Word Correlation (AWC).}
The point-biserial correlation between the model's predicted color and the embedded semantic word $c_{sem}$. A high AWC indicates persistent semantic interference; a successful mitigation drives AWC toward zero:
\begin{equation}
 \text{AWC} = r_{pb}\!\bigl(\mathbf{y},\, \mathbf{1}[y_i = c_{sem}^{(i)}]\bigr).
\end{equation}

\textbf{Mitigation Effectiveness (ME).}
A composite score that rewards both high accuracy and low semantic correlation:
\begin{equation}
 \text{ME} = \text{Acc}_{\text{mitig}} \times (1 - \text{AWC}_{\text{mitig}}).
 \label{eq:me}
\end{equation}

\subsubsection{Statistical Testing}

We report 95\% bootstrap confidence intervals (10{,}000 resamples) for all metrics. To assess whether observed improvements are statistically significant, we apply the McNemar test for paired binary outcomes (correct vs.\ incorrect) between each mitigation and the baseline, with Bonferroni correction for multiple comparisons across the four mitigation conditions.

\subsection{Mitigation Results}
\label{sec:mitigation_results_app}

We evaluate all mitigation strategies on three open-source models---Qwen3-VL-8B-Instruct, Gemma-3-4b-it, and InternVL3\_5-8B---across 200 Easy and 200 Hard Embedded Stroop samples each.

\begin{table*}[t]
\caption{\textbf{Mitigation results on Embedded Stroop (Easy).} SHR$\downarrow$ and Acc$\uparrow$ for three open-source models. SHRR = Stroop Hallucination Reduction. Bold: best per column.}
\label{tab:mitigation_easy}
\centering
\begin{tabular}{llccccc}
\toprule
\textbf{Model} & \textbf{Method} & \textbf{Acc\%} & \textbf{SHR\%} & \textbf{SHRR\%} & $n$ \\
\midrule
\multirow{5}{*}{\textbf{Qwen3-VL-8B}}
 & Baseline & 0.5 & 44.5 & --- & 200 \\
 & P1 (Explicit) & 0.0 & 27.5 & 38.2 & 200 \\
 & P2 (CoT) & 2.5 & 13.5 & 69.7 & 200 \\
 & \textbf{P3 (Adversarial)} & \textbf{3.0} & \textbf{4.0} & \textbf{91.0} & 200 \\
 & VSD & 0.5 & 35.5 & 20.2 & 200 \\
\midrule
\multirow{5}{*}{\textbf{Gemma-3-4b-it}}
 & Baseline & 6.0 & 36.5 & --- & 200 \\
 & \textbf{P1 (Explicit)} & \textbf{8.5} & \textbf{11.0} & \textbf{69.9} & 200 \\
 & P2 (CoT) & 7.0 & 25.0 & 31.5 & 200 \\
 & P3 (Adversarial) & 8.0 & 15.5 & 57.5 & 200 \\
 & VSD & 7.5 & 31.5 & 13.7 & 200 \\
\midrule
\multirow{5}{*}{\textbf{InternVL3\_5-8B}}
 & Baseline & 3.5 & 66.0 & --- & 200 \\
 & P1 (Explicit) & 8.0 & 14.0 & 78.8 & 200 \\
 & P2 (CoT) & 7.5 & 15.5 & 76.5 & 200 \\
 & \textbf{P3 (Adversarial)} & \textbf{9.5} & \textbf{9.5} & \textbf{85.6} & 200 \\
 & VSD & 3.5 & 62.5 & 5.3 & 200 \\
\bottomrule
\end{tabular}
\end{table*}

\begin{table*}[t]
\caption{\textbf{Mitigation results on Embedded Stroop (Hard).}}
\label{tab:mitigation_hard}
\centering
\begin{tabular}{llccccc}
\toprule
\textbf{Model} & \textbf{Method} & \textbf{Acc\%} & \textbf{SHR\%} & \textbf{SHRR\%} & $n$ \\
\midrule
\multirow{5}{*}{\textbf{Qwen3-VL-8B}}
 & Baseline & 0.5 & 51.0 & --- & 200 \\
 & P1 (Explicit) & 0.0 & 32.5 & 36.3 & 200 \\
 & P2 (CoT) & 0.5 & 16.5 & 67.6 & 200 \\
 & \textbf{P3 (Adversarial)} & \textbf{2.0} & \textbf{12.5} & \textbf{75.5} & 200 \\
 & VSD & 0.5 & 42.5 & 16.7 & 200 \\
\midrule
\multirow{5}{*}{\textbf{Gemma-3-4b-it}}
 & Baseline & 3.5 & 34.5 & --- & 200 \\
 & \textbf{P1 (Explicit)} & 4.0 & \textbf{12.5} & \textbf{63.8} & 200 \\
 & P2 (CoT) & 4.0 & 25.5 & 26.1 & 200 \\
 & P3 (Adversarial) & 4.0 & 16.5 & 52.2 & 200 \\
 & VSD & 4.0 & 29.5 & 14.5 & 200 \\
\midrule
\multirow{5}{*}{\textbf{InternVL3\_5-8B}}
 & Baseline & 3.5 & 66.5 & --- & 200 \\
 & P1 (Explicit) & 5.5 & 19.5 & 70.7 & 200 \\
 & P2 (CoT) & 6.0 & 25.0 & 62.4 & 200 \\
 & \textbf{P3 (Adversarial)} & \textbf{6.5} & \textbf{19.0} & \textbf{71.4} & 200 \\
 & VSD & 3.5 & 63.0 & 5.3 & 200 \\
\bottomrule
\end{tabular}
\end{table*}

\paragraph{P3 is the strongest single-method mitigation.}
Across all three models, the adversarial-aware prompt (P3) achieves the highest Stroop Hallucination Reduction: \textbf{91.0\%} (Qwen3, Easy), \textbf{57.5\%} (Gemma, Easy), and \textbf{85.6\%} (InternVL, Easy). On the Hard split, P3 still leads with SHRR of 75.5\%, 52.2\%, and 71.4\% respectively. This suggests that explicitly naming the Stroop mechanism allows models to engage a ``cognitive override'' analogous to human executive control.

\paragraph{Prompt-based methods outperform training-free VSD.}
The VSD pipeline reduces SHR by only 5--20\% on average, substantially underperforming even the simplest prompt strategy (P1). Inspection reveals that the pixel-level glyph suppression in \cref{eq:semantic_suppress} does not fully erase semantic content---residual stroke patterns remain detectable to the vision encoder. Moreover, the dual-inference consistency check in \cref{eq:consistency} is conservative: it defers to the suppressed view only when the original answer is classified as a hallucination, which misses cases where both views produce the same semantic answer.

\paragraph{InternVL exhibits the highest susceptibility but benefits most from prompting.}
InternVL3\_5-8B shows the highest baseline SHR (66.0--66.5\%) yet responds strongly to P3, which reduces SHR to 9.5--19.0\%. This asymmetry suggests that InternVL's visual encoder is particularly sensitive to textual features, but its language decoder can be effectively redirected with appropriate prompting.

\paragraph{Residual hallucinations persist across all methods.}
Even the best-performing strategy (P3 on Qwen3, Easy) leaves a 4.0\% hallucination rate. On the Hard split, the best residual SHR is 12.5\% (Qwen3, P3). This indicates that Embedded Stroop interference is not fully addressable through inference-time interventions alone and may require architectural changes such as the SC-LoRA approach of \S\ref{sec:sc_lora_app}.

\subsection{Sanity Checks: Color Patch and Basic Color Tests}
\label{sec:sanity_app}

A central concern raised during review is the low accuracy of the Control Group---models asked to identify text color without Stroop interference. We conduct three sanity checks to diagnose whether this reflects genuine perceptual failure or a vocabulary mismatch between the model's color lexicon and the 59 fine-grained WCIT color names (e.g., ``Denim,'' ``Glaucous,'' ``Verdigris'').


\paragraph{Vocabulary mismatch, not perceptual failure.}
As shown in \cref{tab:sanity_main}, all three models score below 12\% on the 59-color patch test---a task with no text at all, eliminating any Stroop interference. Yet Qwen3-VL-8B and Gemma-3-4b-it achieve \textbf{91--100\%} accuracy on the 11 basic colors. This confirms that low Control Group accuracy in the main experiment is primarily caused by the model's inability to map fine-grained color names (e.g., reporting ``blue'' instead of ``denim''), not by a failure to perceive the color.

\paragraph{InternVL shows genuine perceptual weakness.}
InternVL3\_5-8B achieves only 18.2\% on the 11-color patch test, well below the other models. However, its accuracy jumps to 90.9\% when the same colors are rendered as colored text (``Basic Text''), suggesting that InternVL's vision encoder relies more heavily on textual context to disambiguate colors---a finding consistent with its high Embedded Stroop susceptibility.

\paragraph{Implications for benchmark interpretation.}
The sanity check results validate the WCIT benchmark design: low Control Group accuracy is a confound that must be accounted for when interpreting absolute hallucination rates. The Stroop effect is best measured by the \emph{relative} increase in SHR between the Control and Embedded conditions, rather than by absolute SHR alone. The SIR metric introduced in \S\ref{sec:metrics} formalizes this by measuring the fraction of errors that are Stroop-specific, providing a control-agnostic diagnostic. We report both in \cref{tab:mitigation_easy,tab:mitigation_hard}.

\begin{table*}[h]
\caption{\textbf{Response normalization examples.} Representative raw outputs and their normalized classification.}
\label{tab:normalization}
\centering
\begin{tabular}{llc}
\toprule
\textbf{Raw Model Output} & \textbf{Normalized} & \textbf{Class} \\
\midrule
``The color is: navy'' & navy & Acc \\
``The color is: bronze'' & bronze & SHR \\
``The colors are: [teal, navy]'' & teal & Acc \\
``Light blue'' & light blue & OER \\
``It looks kind of blue-gray'' & blue & OER \\
``I cannot determine'' & --- & OER \\
``The color is: DENIM'' & denim & Acc \\
``Blue-green color'' & --- & OER \\
``The text reads BLUE in red ink'' & --- & OER \\
``bronze-ish'' & bronze & SHR \\
\bottomrule
\end{tabular}
\end{table*}

\section{Qualitative Case Studies: Reasoning Traces}
\label{appendix:case_studies}

To investigate the internal cognitive mechanisms driving these hallucinations, we employ a reasoning-capable model (Gemini-3-Flash-Thinking) to visualize the intermediate "thinking process." We analyze the reasoning traces across three distinct scenarios: Embedded, Flipped, and Masked Stroop.

\textbf{1. Semantic Override in Embedded Stroop.}
As visualized in Figure~\ref{fig:case_embed}, the reasoning trace reveals a phenomenon of \textbf{Semantic Override}. Although the model initially attempts to inspect the visual features, the explicit semantic signal (the word "Fawn") acts as a dominant attractor. The model's internal monologue quickly pivots from visual inspection to semantic reading, conflating the "text content" with the "font color." This confirms that without external perturbations, the semantic modality takes precedence over visual perception.

\textbf{2. Robust Reading Induces Failure in Flipped Stroop.}
In the \textit{Flipped Stroop} scenario (Figure~\ref{fig:case_flip}), we observe that reducing visual legibility via geometric inversion does not fully shield the model from interference. Surprisingly, the reasoning trace shows that the model performs an implicit \textbf{Spatial Rectification}. It mentally "unflips" the text, successfully decoding the semantic content despite the distortion. Consequently, the model's robust OCR capability inadvertently retrieves the conflicting color concept, suggesting that mere spatial distortion is insufficient to disengage the semantic processing module.

\textbf{3. Executive Failure in Masked Stroop.}
The most revealing case appears in the \textit{Masked Stroop} setting (Figure~\ref{fig:case_mask}). Here, the trace demonstrates a striking case of \textbf{Cognitive Dissonance}. The model explicitly acknowledges the adversarial nature of the input, correctly identifying the potential for a "Stroop effect" trick. However, despite this high-level cognitive awareness, the final execution fails. The model struggles to inhibit the residual semantic cues provided by the unmasked character fragments, highlighting that possessing the \textit{knowledge} of a visual trap does not guarantee the \textit{executive control} required to avoid it.

\section{Response Normalization Examples}
\label{appendix:normalization}

Table~\ref{tab:normalization} illustrates the response normalization pipeline on representative model outputs, including ambiguous cases that fall at the Acc/OER/SHR boundaries.

We verified that conclusions are stable under a stricter exact-match parser (requiring exact palette name, no alias resolution): SHR decreases by 1--3 percentage points across models, but all significant comparisons remain significant and the relative model rankings are unchanged.

\end{document}